\documentclass{article}
\usepackage{
preprint,times}

\usepackage{amsmath,amsfonts,bm}

\def\eqref#1{equation~\ref{#1}}
\def\1{\bm{1}}

\DeclareMathAlphabet{\mathsfit}{\encodingdefault}{\sfdefault}{m}{sl}
\SetMathAlphabet{\mathsfit}{bold}{\encodingdefault}{\sfdefault}{bx}{n}

\usepackage{hyperref}
\usepackage{url}

\usepackage[most]{tcolorbox}
\usepackage{booktabs}
\usepackage{multirow}
\usepackage{graphicx}
\usepackage{makecell}
\usepackage{rotating}
\usepackage{wrapfig}
\usepackage{lineno}
\usepackage{booktabs}
\usepackage{tabularx}
\usepackage{array}
\usepackage{caption}
\usepackage{placeins}
\usepackage{longtable}
\usepackage[table]{xcolor}
\usepackage{enumitem}
\usepackage{float} 
\usepackage{soul}
\usepackage{fontawesome5}
\usepackage{amssymb}
\usepackage{minitoc}

\mtcsetfont{parttoc}{section}{\normalsize\rmfamily\sc}
\mtcsetfont{parttoc}{subsection}{\normalsize\rmfamily\upshape\mdseries}
\mtcsettitlefont{parttoc}{\large\rmfamily\sc}

\definecolor{bodyconfigcolor}{HTML}{14B8A6}
\definecolor{motiondynamicscolor}{HTML}{A855F7}
\definecolor{temporalcolor}{HTML}{F59E0B} 

\colorlet{temporalbg}{temporalcolor!12}
\colorlet{motionbg}{motiondynamicscolor!12}
\colorlet{perceptionbg}{bodyconfigcolor!12}

\definecolor{visualgroundingcolor}{HTML}{F2B840}
\definecolor{motionstatecolor}{HTML}{5F9AE0}
\definecolor{temporallocalizationcolor}{HTML}{54C49B}
\definecolor{priorinferencecolor}{HTML}{776BBD}

\usepackage{xcolor}
\usepackage{colortbl}
\usepackage{pifont}

\definecolor{rowgray}{gray}{0.92}

\newcommand{\cmark}{\textcolor{green!60!black}{\ding{51}}}
\newcommand{\xmark}{\textcolor{red!70!black}{\ding{55}}}

\newcommand{\familyrule}{%
  \arrayrulecolor{black!35}%
  \specialrule{0.2pt}{0.7pt}{0.7pt}%
  \arrayrulecolor{black}%
}

\newcommand{\sectionrule}{%
  \specialrule{\lightrulewidth}{1.1pt}{1.1pt}%
}

\definecolor{sectiongray}{gray}{0.92}

\title{SOMA: A Benchmark for Fine-Grained Egocentric Behavior and Body-Motion Understanding}

\author{Matea Tashkovska, Naël Ouerghemi, Haozhe Qi, Alexander Mathis \\
\raisebox{-0.05em}{\includegraphics[height=0.75em]{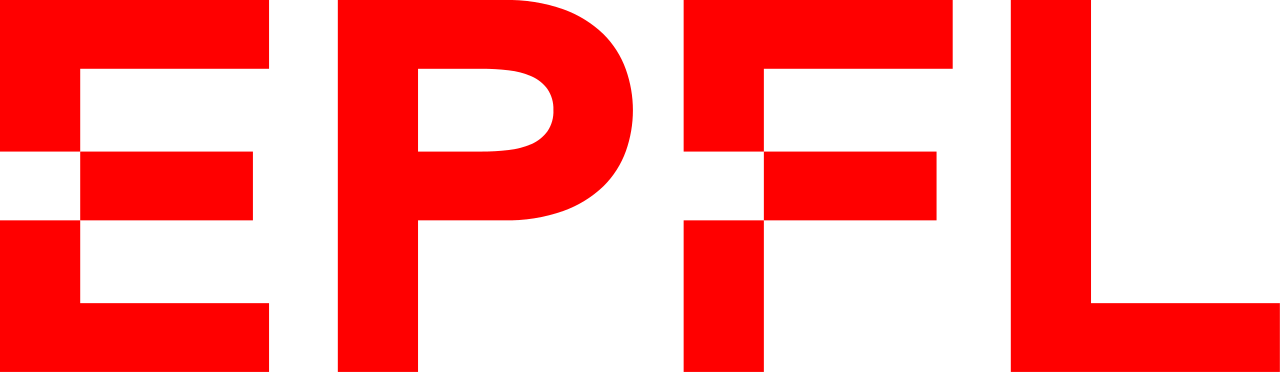}}\space\space École Polytechnique Fédérale de Lausanne (EPFL), Switzerland \\
\small\faEnvelope[regular] \space \texttt{alexander.mathis@epfl.ch} \\
}

\iclrfinalcopy 
\begin{document}
\doparttoc
\faketableofcontents

\maketitle
\lhead{Preprint}

\begin{figure}[H]
    \centering
    \includegraphics[width=\textwidth]{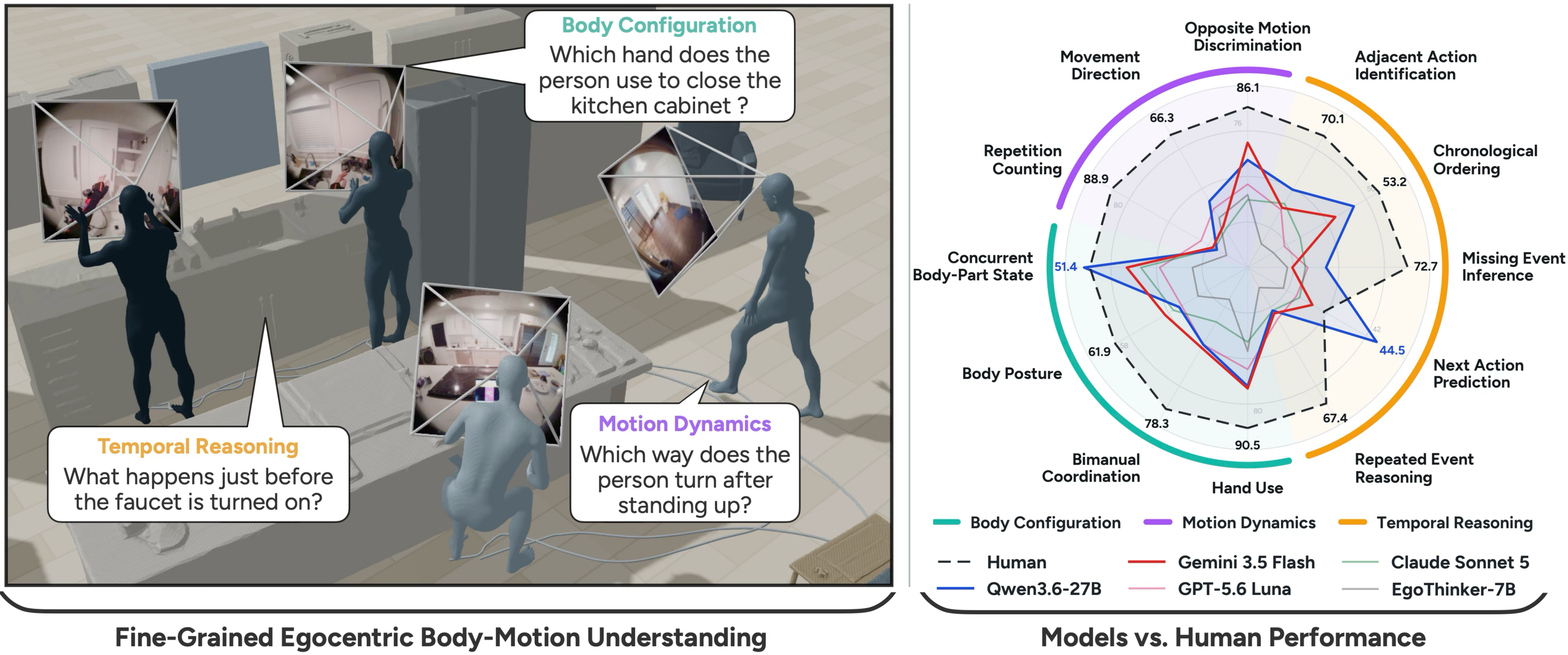}
    \caption{\textbf{Overview of SOMA.} Left: Representative questions from the three capability axes, \textbf{\textcolor{bodyconfigcolor}{Body Configuration}}, \textbf{\textcolor{motiondynamicscolor}{Motion Dynamics}}, and \textbf{\textcolor{temporalcolor}{Temporal Reasoning}}, grounded in egocentric video and body motion. Right: Performance of representative MLLMs across the twelve SOMA tasks compared with human performance (5 annotators), highlighting a human--model gap. Note: performance normalized for visualization purposes.}
    \label{fig:teaser}
\end{figure}

\vspace{-0.2cm}

\begin{abstract}
\vspace{-0.2cm}
Perceiving human body motion is critical for multimodal systems that understand and assist people. Yet current evaluations of Multimodal Large Language Models (MLLMs) largely focus on objects, scenes, and coarse actions, leaving their ability to reason about the body itself poorly assessed. This is especially challenging from the egocentric perspective, in which the body is only partially visible and motion must be inferred from occluded and temporally distributed cues. We introduce SOMA, a benchmark for fine-grained body-motion understanding from egocentric video, comprising 1,632 human-verified questions across three capability axes: body configuration, motion dynamics, and temporal reasoning. We evaluate 37 leading closed-source, open-source, and specialized egocentric MLLMs, complemented by an extensive human evaluation. We find that humans also find this task difficult from video alone (68.5\% accuracy), yet clearly outperform the strongest MLLM (Qwen3.6-27B), which reaches 50.0\% zero-shot accuracy. Our error analysis shows that MLLMs often misread the wearer's own body, misjudge events in time, and are prone to biases such as favoring the right hand or under-counting repetitions. To narrow the gap to human performance and support further progress, we introduce a training dataset of aligned video, body motion, and language supervision, and a multimodal baseline that jointly learns from these modalities. By identifying common failure modes and missing capabilities in current MLLMs, we hope SOMA will inspire future research on body-motion understanding and multimodal reasoning. Full benchmark, training data, and code will be released at: \href{https://cnai.epfl.ch/soma}{\texttt{cnai.epfl.ch/soma}}
\end{abstract}

\section{Introduction}
\vspace{-0.2cm}
Understanding human motion is fundamental for science, and for AI systems that interact with, assist, or learn from humans~\citep{yin2025visualmimic, li2025genmo,bonnetto2026behavior}. 
Recently, Multimodal Large Language Models (MLLMs) have demonstrated strong abilities to perceive and reason about human activity from video, yet largely from a third-person perspective~\citep{li2024llava, qi2025llavaction}. Egocentric video, captured from a first-person perspective using wearable devices, offers a fundamentally different viewpoint, central to augmented reality interfaces, wearable assistants, and embodied agents~\citep{yang_egolife_2025,li2025building,bonnetto2026behavior}.
However, this viewpoint introduces distinct challenges~\citep{grauman2022ego4d, he_egoexobench_2025}. 
First, the actor's body and actions are often only partially visible to the camera, despite being essential for understanding the wearer's intention and activity. Second, fine-grained motion dynamics, such as repetition counts or movement direction, are often hard to perceive from the egocentric view. Finally, motion blur and rapid viewpoint changes are common in egocentric video, requiring MLLMs to integrate ego-motion cues across frames. Fine-grained body-motion understanding from an egocentric view is therefore a challenging task.
Yet existing egocentric video benchmarks (e.g.,~\citep{mangalam_egoschema_2023, jia_egotaskqa_2022,bonnetto2026epfl}) target other dimensions of video understanding, such as long-form video comprehension or domain-specific, goal-oriented task reasoning (Table~\ref{tab:benchmark_comparison}).
None directly measures whether a model can infer the moving body itself: its configuration at a moment and the way its actions unfold over time. As a result, it remains unclear whether progress on egocentric video reflects understanding of human motion or recognition of the objects and scenes in which motion takes place.

To fill this gap, we introduce SOMA (Greek for \textit{body}), a multimodal benchmark for fine-grained understanding of human body motion and behavior from egocentric video. SOMA evaluates MLLMs on 12 tasks organized into three capability axes that mirror the aforementioned challenges: \textit{Body Configuration}, \textit{Motion Dynamics}, and \textit{Temporal Reasoning}, which probe how the body is configured, how it moves, and how its actions unfold in time. 
As a source for egocentric videos with annotations, we selected two comprehensive corpora, Nymeria~\citep{ma_nymeria_2024} and Ego-Exo4D~\citep{grauman_ego-exo4d_2024}. Based on a tailored pipeline including human verification, we derived a final benchmark of 1,632 multiple-choice questions. We additionally release SOMA-Train, a multimodal training dataset of aligned video, body motion, and language supervision, a combination absent from existing egocentric video benchmarks (Table~\ref{tab:benchmark_comparison}).

We evaluated 37 leading MLLMs on SOMA, spanning closed-source, open-source, and specialized egocentric systems. Furthermore, five human evaluators independently answered all 1,632 questions, requiring over 450 hours of manual labor. The best-performing model, Qwen3.6-27B, reaches only 50.0\% zero-shot accuracy, well below the 68.5\% mean accuracy achieved by humans. This gap persists even with denser temporal sampling and explicit test-time reasoning, which yield only limited, model-dependent gains. Error analysis shows that the most common failures stem from misreading the wearer's own body, along with difficulty localizing events in time. Motivated by this, we test whether an explicit body-motion signal helps: fine-tuning on SOMA-Train with aligned video and motion yields modest gains, suggesting that body motion can provide information complementary to video, paving the way toward developing MLLMs that better perceive human motion.

\begin{table*}[t]
\caption{
    \textbf{Benchmark Comparison.}
    \textit{\#Clips}: number of video or motion clips;
    \textit{\#Cat.}: number of categories;
    \textit{Input Modality}: \faCameraRetro\ RGB video, \faChild\ 3D body pose ($\land$: both modalities; $\veebar$: either modality);
    \textit{Ego}: egocentric perspective;
    \textit{Train}: provides a training split;
    \textit{Body}: targets the body, body parts, or posture;
    \textit{Mot.}: fine-grained motion beyond coarse activity recognition;
    \textit{Temp.}: temporal reasoning.
}
\label{tab:benchmark_comparison}
\vspace{0.5\baselineskip}
\vspace{-0.4cm}
\centering
\resizebox{\textwidth}{!}{
\renewcommand{\arraystretch}{1.05}
\setlength{\aboverulesep}{0.8pt}
\setlength{\belowrulesep}{0.8pt}
\begin{tabular}{llrrrcccccc}
\toprule
& \multirow{2}{*}{\textbf{Benchmark}} & \multirow{2}{*}{\makebox[1.8cm][r]{\textbf{\#Clips}}} & \multirow{2}{*}{\textbf{\#QA}} & \multirow{2}{*}{\textbf{\#Cat.}} & \multicolumn{3}{c}{\textbf{Attributes}} & \multicolumn{3}{c}{\textbf{QA Scope}} \\
\cmidrule(lr){6-8} \cmidrule(lr){9-11}
& & & & & \textbf{Modality} & \textbf{Ego} & \textbf{Train} & \textbf{Body} & \textbf{Mot.} & \textbf{Temp.} \\
\midrule

\multirow{4}{*}{\rotatebox[origin=c]{90}{\shortstack{\textit{Human-}\\\textit{centric}}}}
& BABEL-QA~\citep{endo2023motion} & 1,109 & 2,577 & 3 & \faChild & \xmark & \cmark & \cmark & \cmark & \cmark \\
& MoVid~\citep{chen2025motionllm} & 1,350 & 1,350 & 5 & \faCameraRetro \ $\veebar$ \faChild & \xmark & \cmark & \cmark & \cmark & \cmark \\
& MotionBench~\citep{hong_motionbench_2025} & 5,385 & 8,052 & 6 & \faCameraRetro & \xmark & \xmark & \cmark & \cmark & \cmark \\
& ExAct~\citep{yi_exact_2025} & 3,521 & 3,521 & 2 & \faCameraRetro & \xmark & \xmark & \cmark & \xmark & \cmark \\
\midrule

\multirow{6}{*}{\rotatebox[origin=c]{90}{\shortstack{\textit{Egocentric}\\\textit{Video QA}}}}
& EgoTaskQA~\citep{jia_egotaskqa_2022} & 2,000 & 40,000 & 4 & \faCameraRetro & \cmark & \cmark & \xmark & \xmark & \cmark \\
& EgoSchema~\citep{mangalam_egoschema_2023} & 5,063 & 5,063 & 1 & \faCameraRetro & \cmark & \xmark & \xmark & \xmark & \cmark \\
& EgoThink~\citep{cheng_egothink_2024} & 700 & 700 & 12 & \faCameraRetro & \cmark & \xmark & \xmark & \xmark & \xmark \\
& EgoGazeVQA~\citep{peng_eye_2025} & 913 & 1,757 & 11 & \faCameraRetro & \cmark & \xmark & \xmark & \xmark & \cmark \\
& EgoTempo~\citep{plizzari2025omnia} & 365 & 500 & 10 & \faCameraRetro & \cmark & \xmark & \xmark & \xmark & \cmark \\
& LEMONADE~\citep{bonnetto2026epfl} & 36,521 & 36,521 & 31 & \faCameraRetro & \cmark & \xmark & \xmark & \cmark & \cmark \\
\midrule

\rowcolor{rowgray}
& \textbf{SOMA} (Ours) & \textbf{1,611} & \textbf{1,632} & \textbf{12} & \faCameraRetro \ $\land$ \faChild & \cmark & \cmark & \cmark & \cmark & \cmark \\
\bottomrule
\vspace{-2.0em}
\end{tabular}
}
\end{table*}

\section{Related work}

\paragraph{Human Motion Understanding.}
Understanding human motion is a long-standing problem~\citep{aggarwal1999human}, studied through various approaches. One line of work focuses on perception: methods reconstruct 3D body pose from video~\citep{goel2023humans,qi2024hoisdf, yang_sam3dbody_2026} and operate on skeleton inputs to recognize and segment human actions~\citep{duan2022revisiting,jiang2023motiongpt,ren2024survey,stoffl2024elucidating}. A complementary line of work studies motion generation, modeling the distribution of plausible motions to synthesize new ones from text~\citep{guo2024momask,zeng2025light}. In contemporary work, MLLMs treat motion as a modality alongside text, unifying motion comprehension and generation within a single model~\citep{wu2025mg,chen2025motionllm,Li_2025_CVPR}. Yet, models that explicitly combine video and body motion remain uncommon, with few publicly available implementations and model weights. Furthermore, these MLLMs focus on the exocentric setting, where occlusions are rare and motion-capture data is more straightforward. To address this, SOMA evaluates how well current MLLMs perceive body motion from first-person video, complementing the body-motion benchmarks that operate on third-person or motion-capture input alone (Table~\ref{tab:benchmark_comparison}).

\vspace{-0.3cm}
\paragraph{MLLMs for Video Understanding.}
Building on the success of large language models~\citep{achiam2023gpt, yang2025qwen3, cai2024internlm2}, MLLMs extend text-only systems to additional modalities~\citep{han2024onellm, girdhar2022omnivore, comanici2025gemini}. Among these, video MLLMs~\citep{li2024llava, zhang2024video, wang2024internvideo2, li2024mvbench, qi2025llavaction} address the understanding and reasoning of video content. Motivated by applications in AR, robotics, and assistive systems, a separate line of work adapts MLLMs to the egocentric setting. For example, EgoThinker~\citep{pei2026egothinker} enhances egocentric reasoning through chain-of-thought rationales, and EgoGPT~\citep{yang_egolife_2025} fine-tunes a video MLLM for egocentric captioning and QA. However, existing egocentric MLLMs remain optimized for activity- and scene-level understanding, leaving the question of fine-grained body-motion perception largely unexplored. SOMA is designed to evaluate this capability explicitly, exposing the shortcomings of current models, while also providing a training split to further advance their development.

\vspace{-0.3cm}
\paragraph{Egocentric VQA Benchmarks.}
Egocentric video question-answering is expanding rapidly across several directions. Early benchmarks established the format for first-person video QA, with EgoTaskQA~\citep{jia_egotaskqa_2022} probing goal-oriented task understanding and EgoSchema~\citep{mangalam_egoschema_2023} targeting long-form video. More recent benchmarks explore specific aspects of egocentric understanding, including first-person perspective thinking~\citep{cheng_egothink_2024}, cross-domain and domain-specific understanding~\citep{bonnetto2026epfl,li_egocross_2026}, and temporal reasoning~\citep{plizzari2025omnia}. Despite this breadth, none of these benchmarks target the perception of the moving human body. Specifically, tasks such as activity recognition, task planning, and temporal grounding remain focused on objects and the surrounding environment, rather than the body itself. We fill this gap by introducing a body-centric question taxonomy grounded in fine-grained body-motion annotations, paired with video and 3D body pose, which probes competencies that existing benchmarks do not directly evaluate (Table~\ref{tab:benchmark_comparison}).

\begin{figure*}[t]
    \vspace{-2em}
    \centering
    \includegraphics[width=\textwidth]{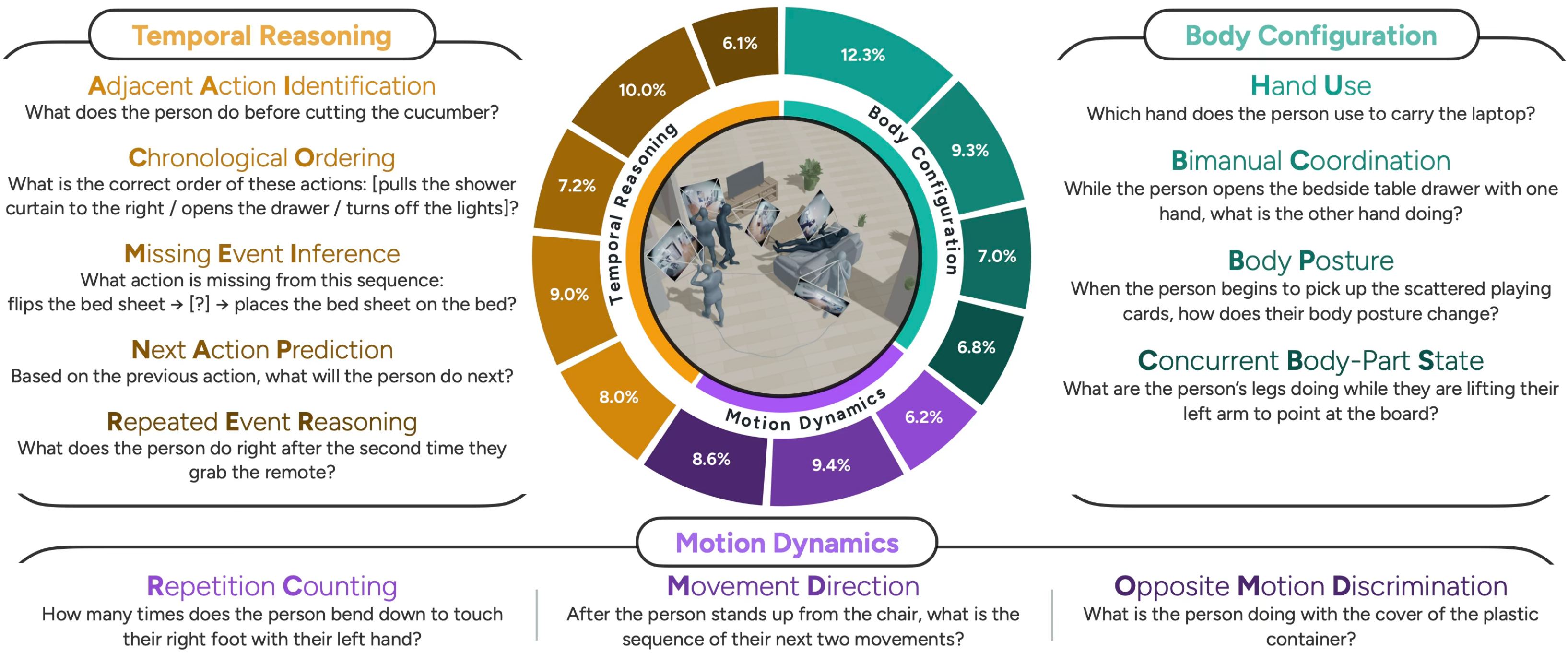}
    \vspace{-0.55cm}
    \caption{\textbf{SOMA Benchmark Taxonomy.} SOMA comprises 12 tasks organized into three capability axes: \textbf{\textcolor{bodyconfigcolor}{Body Configuration}}, \textbf{\textcolor{motiondynamicscolor}{Motion Dynamics}} and \textbf{\textcolor{temporalcolor}{Temporal Reasoning}}. We show the distribution of questions across tasks, together with a representative question for each task. See Appendix~\ref{app:examples} for more examples.}
    \vspace{-0.45cm}
    \label{fig:task_examples}
\end{figure*}

\vspace{-0.1cm}
\section{The SOMA Benchmark}
\label{sec:benchmark}

We introduce SOMA, a multimodal benchmark for fine-grained understanding of human body motion from egocentric videos. We describe the task suite (Sec.~\ref{sec:task_suite}), the data curation pipeline (Sec.~\ref{sec:pipeline}), dataset statistics (Sec.~\ref{sec:stats}), and our training dataset SOMA-Train (Sec.~\ref{sec:soma_train}).

\vspace{-0.1cm}
\subsection{Task Suite}
\label{sec:task_suite}

We organize the benchmark around three axes, each targeting one of the aforementioned challenges: \textit{Body Configuration}, \textit{Motion Dynamics}, and \textit{Temporal Reasoning}. Together, they comprise 12 tasks that span distinct competencies in understanding a moving body from the first-person view (Figure~\ref{fig:task_examples}). 

\vspace{-0.3cm}
\paragraph{Body Configuration.} These tasks probe perception of the body and its parts during an activity. \textit{Hand Use} identifies which hand performs a given action. \textit{Bimanual Coordination} characterizes how the two hands act together, including their role assignment and consistency across repeated instances. \textit{Body Posture} identifies the person's overall posture at a specified moment or window. \textit{Concurrent Body-Part State} identifies what one body part is doing given the activity of another.

\vspace{-0.3cm}
\paragraph{Motion Dynamics.} This axis tests perception of fine-grained motion. \textit{Repetition Counting} enumerates the occurrences of a specified repetitive motion. \textit{Movement Direction} identifies the direction of motion, turning, or a short navigation sequence. \textit{Opposite Motion Discrimination} distinguishes semantically opposed motions such as push and pull, open and close, that share visual context but differ in motion polarity.

\vspace{-0.3cm}
\paragraph{Temporal Reasoning.} This axis groups tasks that require reasoning about how motion-grounded events unfold over time, including their relationships and progression. \textit{Adjacent Action Identification} identifies the action immediately preceding or following a reference event. \textit{Chronological Ordering} places three or more actions in their correct temporal order. \textit{Missing Event Inference} infers the missing intermediate action between two anchoring events. \textit{Next Action Prediction} predicts the action following a trimmed clip, requiring inference of an unobserved action. \textit{Repeated Event Reasoning} identifies what occurs after a specified occurrence of a repeated event.

\subsection{Data Curation Pipeline}
\label{sec:pipeline}

\begin{figure}[t]
    \vspace{-1em}
    \centering
    \includegraphics[width=\textwidth]{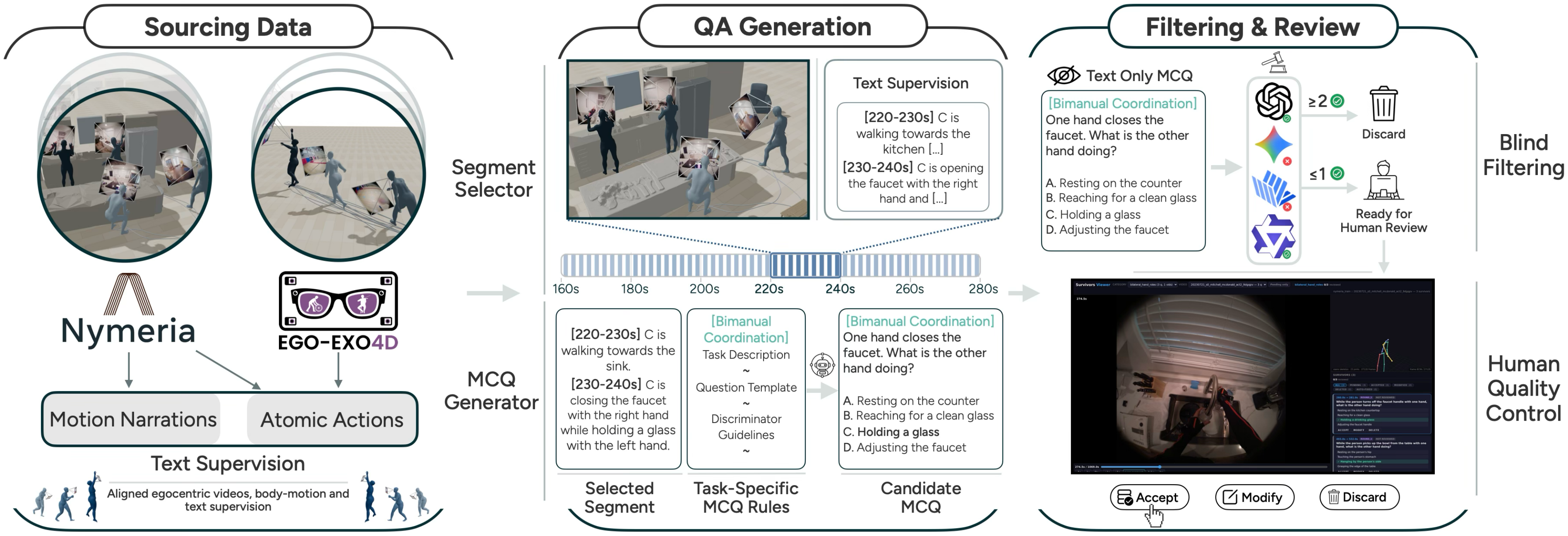}
    \caption{\textbf{SOMA Data Construction Pipeline.} We construct SOMA through task assignment, task-specific question generation, and quality control with blind-filtering and human validation.}
    \label{fig:pipeline}
    \vspace{-.4cm}
\end{figure}

We construct SOMA through a multi-stage pipeline: sourcing data and task assignment, question generation, filtering, and human review (Figure~\ref{fig:pipeline}). We describe each stage below.

\vspace{-0.2cm}

\paragraph{Sourcing Data and Task Assignment.}
We sourced videos from Nymeria~\citep{ma_nymeria_2024} and Ego-Exo4D~\citep{grauman_ego-exo4d_2024}, two egocentric corpora that span a wide range of real-world domains with rich action and body motion annotations. 
Both corpora provide atomic action annotations $\mathcal{A} = \{(a_j, \tau_j)\}$, where $a_j$ is a detailed natural-language action description and $\tau_j = [t_j^{s}, t_j^{e}]$ its temporal interval.
Nymeria additionally provides motion narrations $\mathcal{M} = \{(p_j, h_j, \ell_j, \tau_j)\}$, fine-grained descriptions of body posture $p_j$, hand motion $h_j$, and leg motion $\ell_j$ over the same interval. These narrations are central to SOMA, as they describe body state and movement in natural language, which supports our body-centric question generation.
Not every domain, however, supports every task equally; therefore, we discarded domain--task pairs that would produce a trivial question, such as asking about hand use during an activity where the hands are idle.

\vspace{-0.2cm}
\paragraph{Question and Answer Generation.}
Our question generation procedure consists of two steps. First, an LLM \textit{selector} receives the relevant textual annotation streams for a video (drawn from $\mathcal{A}$ and $\mathcal{M}$) and identifies a sub-sequence specifically relevant to the task. For example, for Bimanual Coordination, it identifies a segment in which both hands are simultaneously active.
The annotation streams from the selected segment are then passed as input to an LLM \textit{generator}, which produces a multiple-choice question with one correct answer and three task-specific distractors.
This two-step design narrows the context before question construction, so each example is grounded in task-relevant annotations. 
We implemented both modules with Gemini~2.5~Flash and manually designed detailed prompts with step-by-step instructions and in-context examples for each task (Appendix~\ref{app:prompts}), producing 3,928 candidate questions.
The only exception to this step is Repetition Counting, as exact counts of fine-grained motions cannot be reliably extracted from the narrations, and we found it more effective to create the questions and counts manually (Appendix~\ref{app:rep_counting}). 

\paragraph{Blind-Language Filtering.}
Modern LLMs carry substantial prior knowledge and can often answer questions correctly without ``seeing'' the video. A central goal of SOMA is to test whether models can perceive body motion in egocentric video, so questions answerable from language alone undermine the benchmark. To identify such questions, we apply a blind-language filter: four models (Gemini 2.5 Flash~\citep{comanici2025gemini}, GPT-4o~\citep{hurst2024gpt}, Qwen2.5-72B~\citep{qwen2025qwen25technicalreport}, and InternVL3-78B~\citep{zhu2025internvl3}) answer each question without video, and we remove questions answered correctly by at least two solvers (Appendix~\ref{app:filtering_stats}). This step removed 45.9\% of the questions, leaving 2,127 for human review and concentrating the benchmark on questions that cannot be reliably solved from language priors and require visual understanding. 

\vspace{-0.2cm}
\paragraph{Human Review.}
Finally, to verify the correctness and quality of every remaining question, a human reviewer inspects each question through a custom interface that streams the source video segment alongside a synchronized 3D body-pose rendering, used as an additional reference for validation. For each question, the reviewer chooses to accept, modify (its text, options, or segment bounds), or delete it, to minimize question ambiguity, so that the final benchmark contains only high-quality, verified questions (Appendix~\ref{app:filtering_stats}). 

\subsection{Dataset Statistics}
\label{sec:stats}

SOMA consists of 1,632 multiple-choice questions that span 12 tasks across three axes (Figure~\ref{fig:task_examples}).
The questions are drawn from 397 unique videos, producing 1,611 clips that span diverse everyday scenarios. The average length is 35.6 seconds with most concentrated between 10 and 60 seconds. Body Configuration and Motion Dynamics clips fall mainly within this range, reflecting the short-term nature of body-motion cues, while Temporal Reasoning clips extend further into the long tail. 
Answer distributions are balanced across options to mitigate position bias. 

\vspace{-0.1cm}
\subsection{SOMA-Train}
\label{sec:soma_train}
To facilitate further investigation and motivate future research in the domain, we introduce a training split, \textbf{SOMA-Train}, which comprises aligned egocentric video, body-motion, and language supervision.
We adapted the benchmark generation pipeline (Sec.~\ref{sec:pipeline}), without human verification, and constructed 28,596 question-answer pairs over 110.2 hours of aligned video and body motion, while keeping the training data disjoint from the SOMA evaluation set at the participant and scene level. Since existing Ego-Exo4D~\citep{grauman_ego-exo4d_2024} body-motion fits are provided at 10~fps and cover only part of the videos, we follow \citet{patel_uniegomotion_2025} to reconstruct 3D body motion at 20~fps from the multi-view recordings. We detail the construction, debiasing method, and motion processing in Appendix~\ref{app:soma_train}.

\vspace{-0.1cm}
\section{Experimental Setup}
\label{sec:experimental-setup}

\vspace{-0.1cm}
\paragraph{Human Performance.} 
To contextualize model performance, we measure human accuracy without prior familiarity with the benchmark.  Five compensated evaluators, none of whom took part in data annotation, independently answered all 1,632 questions, totaling over 450 hours of evaluation. We report their mean accuracy as an estimate of human-level performance on this task. 

\vspace{-0.3cm}
\paragraph{Evaluation Setup.}
We evaluate 37 leading MLLMs spanning closed-source, open-source, and specialized egocentric systems (Table~\ref{tab:SOMA_results} and Appendix~\ref{app:evaluated-models}). We report per-task and overall accuracy~(\%), using \texttt{lmms-eval}~\citep{zhang2024lmmsevalrealitycheckevaluation}, a standard multimodal evaluation framework, to ensure consistent evaluation and facilitate future comparisons. Unless otherwise specified, models are evaluated zero-shot with temperature set to 0 and uniformly sampled 16 frames per segment to balance computational cost and comparability across models with different context lengths. We also ablate key evaluation settings (Sec.~\ref{sec:ablations}). Further implementation details, including prompt templates, are provided in Appendix~\ref{app:eval_details}.

\section{Experiments}
\label{sec:results}

We first compare MLLMs with human performance across the SOMA tasks (Sec.~\ref{sec:main_results}) and examine the effect of inference-time strategies (Sec.~\ref{sec:inference_analysis}). We then analyze common failure modes (Sec.~\ref{sec:error_analysis}) and study whether combining video and body-motion signals can improve performance (Sec.~\ref{sec:training}).

\vspace{-0.5em}
\subsection{MLLM and Human Performance on SOMA}
\label{sec:main_results}

\textbf{All evaluated models struggle on SOMA, and no model reaches human performance.} 
The best model, Qwen3.6-27B, reaches 50.0\% average accuracy, and the best closed-source model, Gemini 3.5 Flash, reaches 47.0\%, well below the human average of 68.5\% (Table~\ref{tab:SOMA_results}). The gap, however, is not uniform. Compared with the best model per task, it is largest on Repetition Counting (45.3 points), Repeated Event Reasoning (25.4), and Bimanual Coordination (25.3), whereas it closes or reverses on Concurrent Body-Part State ($-$1.1) and Next Action Prediction ($-$10.0). The difference in these tasks lies in whether the answer is present in the video or must be inferred. In Concurrent Body-Part State, the queried body part is often outside the field of view, and in Next Action Prediction, the action has not yet occurred. Since both humans and models must infer the answer, humans have little advantage on these tasks. In contrast, on the other tasks, the relevant cues are present in the video but are subtle or spread over time. While humans rely on these cues, we find that current models fail to use them reliably.

\textbf{Performance is uneven across tasks.} Models perform best on tasks where the evidence is concentrated in a few frames and can be directly observed, such as Hand Use and Opposite Motion Discrimination, where the best models reach 74.6\% and 71.4\%, respectively. In contrast, performance drops when the answer requires following the motion across the clip: on Chronological Ordering and Repeated Event Reasoning, most models score below 40\%. This indicates that current MLLMs are better at recognizing frame-level cues than at integrating body motion over time.

\textbf{Egocentric specialization alone is not sufficient.} Despite being adapted to egocentric video, the three specialized models do not outperform general-purpose open-source models of similar size. This suggests a gap in current egocentric models when it comes to body-motion perception, consistent with their training data, which focuses primarily on actions, objects, and intentions, with no dedicated supervision of whole-body posture or motion. Addressing this limitation will likely require egocentric models that are more robust and more sensitive to fine-grained body dynamics.

\textbf{Increasing model size improves performance, but larger models still fall short.} Qwen3-VL gains 13.1 points from 2B to 32B, and Qwen3.5 gains 17.8 points from 0.8B to 27B, while InternVL3.5 improves by 8.1 points from 1B to 38B and is not monotonic. Larger backbones therefore help on SOMA, but scaling alone does not close the gap to human performance, and the largest models tested remain far from saturating the benchmark.

\subsection{Impact of Temporal Sampling and Test-Time Reasoning}
\label{sec:inference_analysis}
\label{sec:ablations}

\paragraph{Effect of Temporal Sampling Density.} 
To study how temporal sampling affects performance, we evaluate nine models on 1 to 64 uniformly sampled input frames (Figure~\ref{fig:effect-analyis}A). 
Accuracy increases when moving from a single to multiple frames but saturates quickly. Averaged across models, it rises by 5.4 points from 1 to 16 frames, while adding frames beyond 16 contributes at most an additional 0.5. 
The magnitude of this gain is strongly model-dependent, ranging from 1.2 points (VideoLLaMA3-7B) to 10.3 points (Qwen3.5-9B) between 1 and 16 frames. 
Temporal sampling density is therefore not the principal limitation. Beyond access to visual evidence on the order of tens of frames, what models lack is the ability to integrate that evidence across time.

\setlength{\tabcolsep}{3.5pt}
\renewcommand{\arraystretch}{0.92}
\begin{table*}[!t]
\caption{
\textbf{Comparison Results on SOMA.}
Accuracy (\%) across the three capability axes and their twelve tasks.
HU: Hand Use; BC: Bimanual Coordination; BP: Body Posture; CBS: Concurrent Body-Part State; RC: Repetition Counting; MD: Movement Direction; OMD: Opposite Motion Discrimination; AAI: Adjacent Action Identification; CO: Chronological Ordering; MEI: Missing Event Inference; NAP: Next Action Prediction; RER: Repeated Event Reasoning. The best model results are marked in \textbf{bold}, and the second-best are \underline{underlined}.
Human performance is the mean over five annotators; standard deviations are reported in Table~\ref{tab:human_std}.}
\vspace{-0.4cm}
\label{tab:SOMA_results}
\vspace{0.5\baselineskip}
\centering
\fontsize{4.8}{5.2}\selectfont
\resizebox{\textwidth}{!}{
\begin{tabular}{l|cccc|ccc|ccccc|c}
\toprule
\multirow{2}{*}{\textbf{Models}} 
& \multicolumn{4}{c|}{\textbf{\textcolor{bodyconfigcolor}{Body Configuration}}} 
& \multicolumn{3}{c|}{\textbf{\textcolor{motiondynamicscolor}{Motion Dynamics}}} 
& \multicolumn{5}{c|}{\textbf{\textcolor{temporalcolor}{Temporal Reasoning}}} 
& \multirow{2}{*}{\textbf{Avg.}} \\
& HU & BC & BP & CBS
& RC & MD & OMD
& AAI & CO & MEI & NAP & RER
& \\
\midrule
\textit{Human Performance} 
& \textit{90.5} & \textit{78.3} & \textit{61.9} & \textit{50.3}
& \textit{88.9} & \textit{66.3} & \textit{86.1}
& \textit{70.1} & \textit{53.2} & \textit{72.7} & \textit{34.5} & \textit{67.4}
& \textit{68.5} \\
\textit{Random Chance} 
& \textit{25.0} & \textit{25.0} & \textit{25.0} & \textit{25.0}
& \textit{25.0} & \textit{25.0} & \textit{25.0}
& \textit{25.0} & \textit{25.0} & \textit{25.0} & \textit{25.0} & \textit{25.0}
& \textit{25.0} \\
\sectionrule
\multicolumn{14}{c}{\textit{\textbf{Closed-source MLLMs}}} \\
\sectionrule
Claude Sonnet 5 
& 52.7 & 42.4 & 43.5 & 40.5
& 36.6 & 33.3 & 47.9
& 44.3 & 33.3 & 39.0 & 29.9 & 35.0
& 40.3 \\
Claude Opus 4.8 
& 64.2 & 45.7 & 49.6 & \underline{49.5}
& 36.6 & 35.3 & 51.4
& 42.0 & 37.4 & 46.6 & 34.1 & 37.0
& 44.8 \\
\familyrule
GPT-5.6 Luna 
& 64.7 & \underline{51.7} & 40.9 & 36.9
& \textbf{43.6} & 40.5 & 54.3
& 42.0 & 29.3 & 39.8 & 29.3 & 38.0
& 43.4 \\
GPT-5.1 
& 59.7 & 39.7 & 37.4 & 42.3
& \underline{40.6} & 31.4 & 60.7
& 41.2 & 22.4 & 40.7 & 22.6 & \textbf{42.0}
& 40.3 \\
\familyrule
Gemini 3.5 Flash 
& 73.1 & \textbf{53.0} & 46.1 & 43.2
& 37.6 & 34.6 & \textbf{71.4}
& 42.7 & 42.2 & 34.7 & 32.3 & 36.0
& 47.0 \\
Gemini 2.5 Pro 
& 67.7 & 47.0 & 34.8 & 45.9
& 34.7 & \textbf{46.4} & 65.0
& \textbf{49.6} & 37.4 & \textbf{55.1} & 28.7 & 39.0
& 46.9 \\
Gemini 2.5 Flash 
& 43.3 & 41.1 & 38.3 & 36.9
& 36.6 & 36.6 & 39.3
& 38.2 & 27.2 & 26.3 & 20.1 & 32.0
& 34.8 \\
\sectionrule
\multicolumn{14}{c}{\textit{\textbf{Open-source MLLMs}}} \\
\sectionrule
Qwen3.8-27B 
& 72.1 & 47.7 & 46.1 & 44.1
& 36.6 & 40.5 & 58.6
& 44.3 & 34.7 & \underline{48.3} & 39.0 & 37.0
& 47.0 \\
Qwen3.6-27B 
& 72.1 & \underline{51.7} & 41.7 & \textbf{51.4}
& 35.6 & \underline{43.1} & 64.3
& \textbf{49.6} & 46.9 & 45.8 & \textbf{44.5} & 35.0
& \textbf{50.0} \\
\familyrule
Qwen3.5-0.8B 
& 40.3 & 39.1 & 25.2 & 35.1
& 24.8 & 28.1 & 33.6
& 29.8 & 32.7 & 26.3 & 31.7 & 24.0
& 31.7 \\
Qwen3.5-2B 
& 48.3 & 38.4 & 27.0 & 35.1
& 23.8 & 26.1 & 44.3
& 29.0 & 33.3 & 18.6 & 25.6 & 32.0
& 32.7 \\
Qwen3.5-4B 
& \textbf{74.6} & 42.4 & 34.8 & 40.5
& 39.6 & 32.0 & 51.4
& 45.0 & 38.8 & 46.6 & 25.0 & 32.0
& 43.1 \\
Qwen3.5-9B 
& 70.1 & 48.3 & 41.7 & 43.2
& 20.8 & 39.2 & 61.4
& 38.2 & 40.1 & 34.7 & 35.4 & 38.0
& 44.3 \\
Qwen3.5-27B 
& \underline{73.6} & 47.0 & 44.3 & 47.7
& 31.7 & 42.5 & \underline{67.9}
& \underline{45.8} & \textbf{49.7} & 45.8 & 39.0 & \textbf{42.0}
& \underline{49.5} \\
\familyrule
Qwen3-VL-2B 
& 33.8 & 33.8 & 33.0 & 34.2
& 34.7 & 33.3 & 32.1
& 32.8 & 26.5 & 27.1 & 19.5 & 36.0
& 31.1 \\
Qwen3-VL-4B 
& 46.8 & 29.1 & 33.9 & 32.4
& 28.7 & 32.0 & 52.9
& 34.4 & 32.0 & 22.9 & 22.6 & 36.0
& 34.1 \\
Qwen3-VL-8B 
& 56.2 & 33.1 & 32.2 & 28.8
& 32.7 & 32.7 & 57.9
& 38.2 & 26.5 & 39.0 & 19.5 & \textbf{42.0}
& 37.1 \\
Qwen3-VL-32B 
& 70.1 & 41.7 & 35.7 & 40.5
& 36.6 & 34.6 & 67.1
& 43.5 & 37.4 & 44.9 & 25.6 & \underline{40.0}
& 44.2 \\
\familyrule
InternVL3.5-1B 
& 33.3 & 25.8 & 36.5 & 32.4
& \underline{40.6} & 30.7 & 28.6
& 27.5 & 30.6 & 22.9 & 32.9 & 31.0
& 30.9 \\
InternVL3.5-2B 
& 33.3 & 31.8 & 45.2 & 34.2
& 36.6 & 33.3 & 35.0
& 35.9 & 29.9 & 25.4 & 28.0 & 36.0
& 33.4 \\
InternVL3.5-4B 
& 47.8 & 26.5 & 40.9 & 28.8
& 33.7 & 31.4 & 44.3
& 32.8 & 23.1 & 28.0 & 23.2 & 31.0
& 33.0 \\
InternVL3.5-8B 
& 35.8 & 31.1 & 41.7 & 37.8
& 31.7 & 32.0 & 46.4
& 34.4 & 17.7 & 37.3 & 29.3 & 27.0
& 33.4 \\
InternVL3.5-14B 
& 50.2 & 37.1 & 43.5 & 39.6
& 37.6 & 28.8 & 55.0
& 38.9 & 25.9 & 42.4 & 23.2 & 36.0
& 38.2 \\
InternVL3.5-38B 
& 52.2 & 44.4 & 37.4 & 32.4
& 36.6 & 30.1 & 59.3
& 42.0 & 29.9 & 45.8 & 17.7 & 38.0
& 39.0 \\
\familyrule
Gemma-4-12B 
& 45.3 & 41.7 & 36.5 & 46.8
& 32.7 & 22.9 & 30.7
& 28.2 & \underline{47.6} & 22.0 & 30.5 & 26.0
& 34.8 \\
Gemma-4-31B 
& 67.2 & 41.1 & 39.1 & 46.8
& 37.6 & 34.6 & 50.7
& 38.2 & 38.8 & 30.5 & \underline{40.2} & 38.0
& 43.1 \\
\familyrule
GLM-4.6V-Flash 
& 53.2 & 41.1 & 32.2 & 39.6
& 23.8 & 34.0 & 43.6
& 33.6 & 36.1 & 26.3 & 20.7 & 28.0
& 35.4 \\
GLM-4.6V 
& 54.7 & 36.4 & 36.5 & 36.9
& 35.6 & 34.6 & 57.1
& 42.0 & 40.8 & 26.3 & 20.7 & 37.0
& 38.8 \\
\familyrule
Kimi-VL-A3B 
& 38.8 & 30.5 & 33.9 & 34.2
& 32.7 & 34.6 & 45.0
& 35.1 & 25.2 & 24.6 & 23.2 & 30.0
& 32.5 \\
\familyrule
MiniCPM-o-4.5-9B 
& 55.2 & 35.8 & 40.9 & 46.8
& 32.7 & 29.4 & 42.9
& 35.1 & 30.6 & 26.3 & 23.2 & 30.0
& 36.3 \\
\familyrule
Molmo2-4B 
& 56.7 & 37.1 & \textbf{53.0} & 45.9
& 33.7 & 29.4 & 41.4
& 32.8 & 19.0 & 28.0 & 26.2 & 35.0
& 36.8 \\
Molmo2-8B 
& 54.2 & 42.4 & \underline{51.3} & 41.4
& 33.7 & 36.6 & 39.3
& 39.7 & 38.1 & 37.3 & 36.0 & 29.0
& 40.6 \\
\familyrule
LLaVA-OneVision-1.5-8B 
& 44.3 & 32.5 & 34.8 & 28.8
& 27.7 & 28.1 & 35.7
& 26.7 & 30.6 & 27.1 & 21.3 & 29.0
& 31.1 \\
\familyrule
VideoLLaMA3-7B 
& 36.8 & 35.1 & 40.9 & 35.1
& 38.6 & 34.6 & 32.1
& 30.5 & 23.1 & 24.6 & 25.0 & 33.0
& 32.3 \\
\sectionrule
\multicolumn{14}{c}{\textit{\textbf{Egocentric MLLMs}}} \\
\sectionrule
EgoGPT-7B
& 46.3 & 27.2 & 33.0 & 27.9
& 31.7 & 32.0 & 40.7
& 26.7 & 16.3 & 20.3 & 29.9 & 30.0
& 30.8 \\
EgoVLM-3B 
& 40.3 & 27.2 & 27.0 & 19.8
& 29.7 & 35.3 & 46.4
& 32.8 & 39.5 & 34.7 & 24.4 & 38.0
& 33.3 \\
EgoThinker-7B
& 56.7 & 33.1 & 35.7 & 30.6
& 30.7 & 37.3 & 50.0
& 29.0 & 26.5 & 33.1 & 26.8 & 27.0
& 35.8 \\
\bottomrule
\end{tabular}
}

\vspace{0.2cm}
\includegraphics[width=0.99\textwidth]{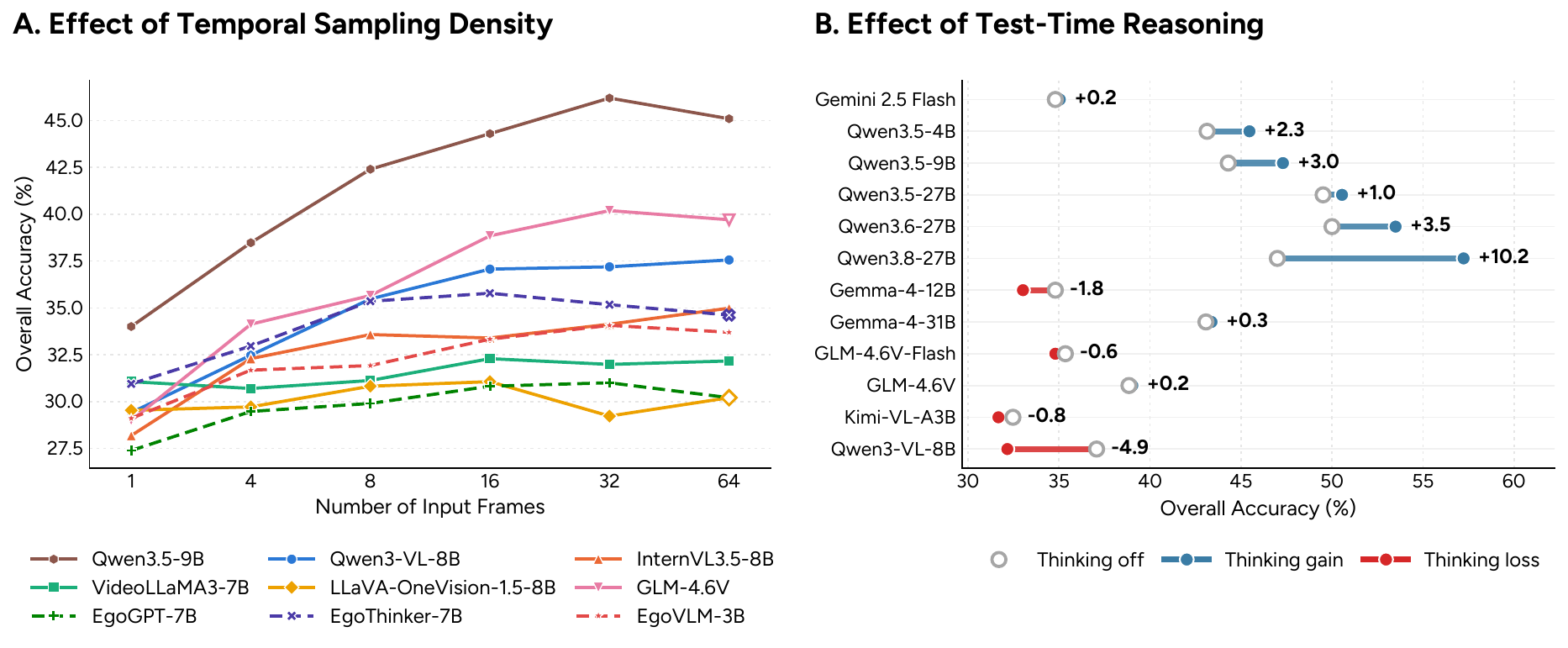}
\vspace{-0.2cm}
\captionof{figure}{\textbf{Effect of Inference Strategies on SOMA.}
\textbf{(A)} Increasing temporal sampling density.
\textbf{(B)} Enabling test-time reasoning (``thinking'').}
\label{fig:effect-analyis}
\end{table*}
\FloatBarrier

\vspace{-0.3cm}
\paragraph{Effect of Test-Time Reasoning.}
We next ask whether explicit test-time reasoning, effective on many language reasoning tasks, also helps with visual reasoning, specifically body-motion perception in video. We compare the performance change after enabling thinking for instruction-tuned models, or after switching to their thinking variant (Qwen3-VL-8B and Kimi-VL-A3B), under each model's recommended sampling parameters. The effect is strongly model-dependent (Figure~\ref{fig:effect-analyis}B). Five models improve by at least one point, all from the Qwen3.5–3.8 family, with gains of 1.0 to 10.2 points, while the remaining seven either show negligible change or decline.
Test-time reasoning is therefore not a sufficient solution for closing the human--model gap: simply encouraging ``thinking'' in current reasoning mechanisms does not address their limitations in perceiving body motion.

\subsection{Quantitative and Qualitative Error Analysis}
\label{sec:error_analysis}
The nature of some questions allows us to explicitly quantify model errors, which we complement with a qualitative analysis of reasoning traces.

\textbf{Models show a strong right-hand bias on handedness questions.} We split each handedness question by its correct answer and measure per-class accuracy. Averaged across the 37 models of Table~\ref{tab:SOMA_results}, accuracy reaches 78.8\% for \textit{right hand} but drops to 44.3\% for \textit{left hand} (Appendix~\ref{app:right-hand-prior}), on a question that is, in principle, visually unambiguous: the five human annotators are equally accurate on both classes (mean accuracy of 92.7\% left, 89.8\% right). This asymmetry is consistent with a prior inherited from pretraining data: since roughly 90\% of the human population is right-handed~\citep{papadatou2020human}, right-handed actions are over-represented in the video-text corpora used to train MLLMs, and models default to \textit{right} when the visual evidence is weak.

\textbf{Most models lean toward under-counting.} Repetition counting shows the largest human--model gap in SOMA (88.9\% for humans against 43.6\% for the best model on this task), so we look more closely at how models fail, classifying each incorrect prediction as an under-count (below the true count) or an over-count (above the true count). In total, 24 of the 37 evaluated models lean toward under-counting, consistent with a frame-sampling explanation: repetitions falling between sampled frames are simply missed, biasing the count downward. The asymmetry is nonetheless strongly model-dependent, indicating that beyond the shared sampling bottleneck, individual models apply their own priors when they cannot directly enumerate repetitions (Appendix~\ref{app:counting}).

\vspace{-0.3cm}
\paragraph{Errors in Reasoning Traces.}

\begin{wrapfigure}{r}{0.55\textwidth}
    \centering
    \vspace{-0.4cm}
    \includegraphics[width=\linewidth]{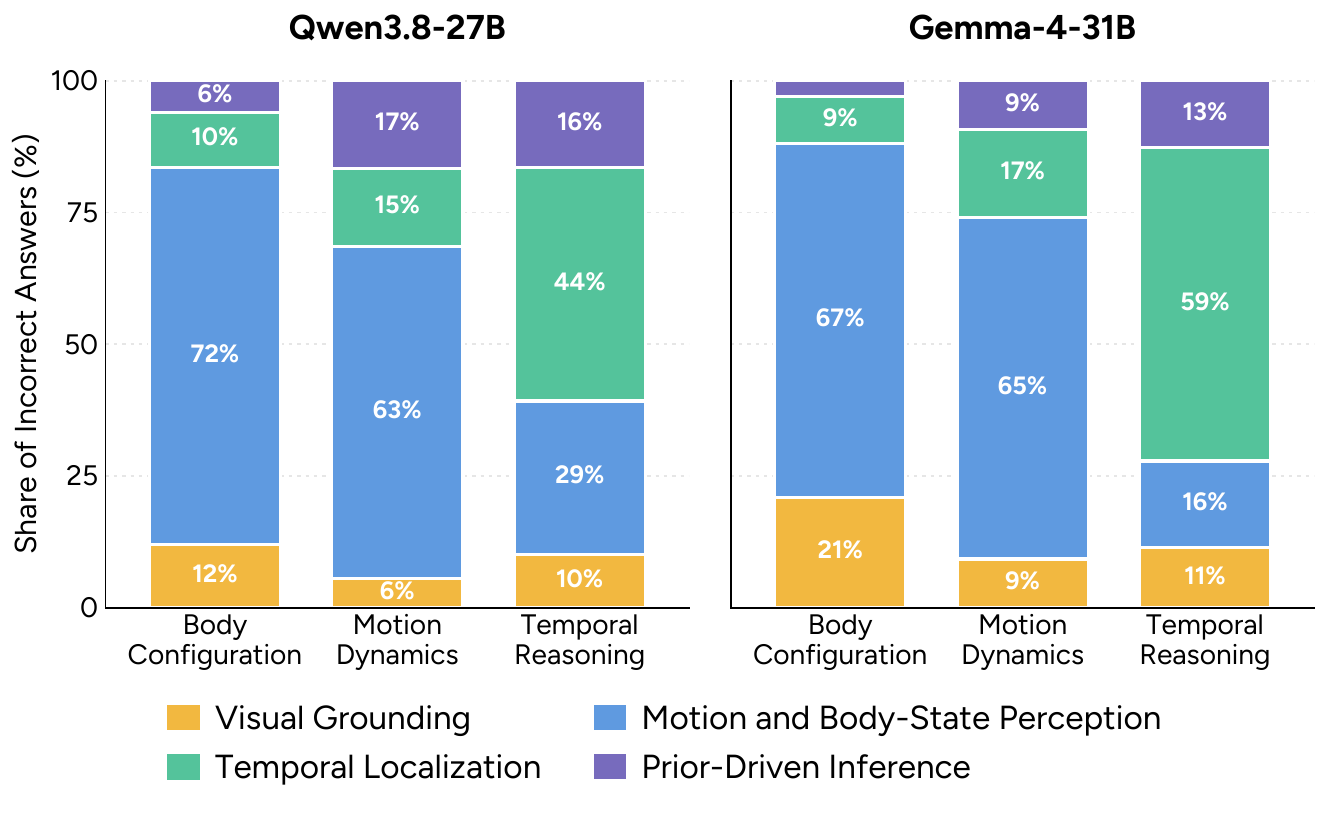}
    \vspace{-.6cm}
    \caption{\textbf{Distribution of Error Types.} Root causes of incorrect answers across two representative MLLMs.}
    \vspace{-.4cm}
    \label{fig:errors_analysis}
\end{wrapfigure}

To better understand the limitations of current MLLMs, we analyze the errors made in their reasoning traces. We focus on Qwen3.8-27B and Gemma-4-31B, the latest models from the two leading open-source families on SOMA, and select 200 questions, stratified across tasks, that both models answered incorrectly. By inspecting the wrong predictions, we identify four primary error types: \textbf{1) \textcolor{visualgroundingcolor}{Visual Grounding Errors}} involve misidentifying the objects, scenes, or people involved; \textbf{2) \textcolor{motionstatecolor}{Motion and Body-State Perception Errors}} involve incorrectly perceiving the wearer's own movement, posture, or limb use; \textbf{3) \textcolor{temporallocalizationcolor}{Temporal Localization Errors}} involve placing correctly identified events at the wrong point in time; and \textbf{4) \textcolor{priorinferencecolor}{Prior-Driven Inference Errors}} involve answering from expectation rather than visual evidence. Further details, including the annotation procedure and qualitative examples of each error type, are provided in Appendix~\ref{app:error-analysis-details}.

Motion and body-state perception is the dominant error for both models on Body Configuration and Motion Dynamics, indicating that most failures on these axes stem from misreading the wearer's own body, including misjudging their posture, the limb in use, or the direction of their movement (examples in Appendix~\ref{app:error-qualitative-examples}). 
In addition, models also struggle with temporal localization, showing difficulty in tracking the wearer's dynamics over time and in correct order. 
Visual grounding and prior-driven inference account for smaller shares of failures across all three axes, reflecting difficulties in identifying the relevant visual information and a tendency to favor plausible over observed answers. 
By examining these failures, our analysis points to specific perceptual and temporal weaknesses that the next generation of MLLMs should address.

\subsection{Training Body-Aware MLLMs} 
\label{sec:training}

We found that MLLMs fail most when they need to perceive the wearer's body, which is hard to infer from egocentric vision. Humans do not rely on vision only: other exteroceptive and proprioceptive signals complement vision (vestibular, muscle spindles and tactile) and provide a continuous estimate of one's own posture and movement~\citep{proske2012proprioceptive}. This raises a natural question: if a model is given a comparable signal, a stream of body-motion information alongside the video, does it perceive and reason better? 

\begin{wraptable}{r}{0.49\textwidth}
    \vspace{-.3cm}
    \caption{\textbf{Effect of Body-Motion Input.} Accuracy (\%) per capability axis after fine-tuning with different input modalities. Per-task results in Table~\ref{tab:motion_finetune_per_task}. $^\dagger$Zero-shot backbones are evaluated under the same video preprocessing as the fine-tuned models (16 frames, matched resolution).}
    \label{tab:SOMA_finetune}
    \vspace{-.35cm}
    \vspace{0.3\baselineskip}
    \centering
    \footnotesize
    \setlength{\tabcolsep}{3.5pt}
    \renewcommand{\arraystretch}{1.0}
    \resizebox{\linewidth}{!}{
    \begin{tabular}{llcccc}
    \toprule
    \textbf{Backbone} & \textbf{Setting} &
    \textcolor{bodyconfigcolor}{\textbf{Body}} & \textcolor{motiondynamicscolor}{\textbf{Mot.}} & \textcolor{temporalcolor}{\textbf{Temp.}} & \textbf{Avg.} \\
    \midrule
    \multirow{4}{*}{Qwen3-VL-8B}
     & Zero-Shot$^\dagger$   & 35.88 & 40.89 & 31.67 & 35.39 \\
     & Motion      & 27.32 & 28.12 & 29.02 & 28.21 \\
     & Video       & 44.99 & 43.75 & 37.13 & 41.49 \\
     & Video+Mot.  & \textbf{45.54} & \textbf{45.57} & \textbf{37.91} & \textbf{42.44} \\
    \midrule
    \multirow{4}{*}{InternVL3.5-8B}
     & Zero-Shot$^\dagger$   & 38.43 & 40.10 & 29.49 & 35.20 \\
     & Motion      & 30.97 & 28.39 & 29.02 & 29.54 \\
     & Video       & 45.17 & \textbf{44.53} & 36.35 & 41.42 \\
     & Video+Mot.  & \textbf{45.54} & 43.75 & \textbf{37.75} & \textbf{41.93} \\
    \bottomrule
    \end{tabular}
    }
    \vspace{-.2cm}
\end{wraptable}

\vspace{-.1cm}
We investigated this by extending two open-weight MLLM backbones, Qwen3-VL-8B~\citep{bai2025qwen3vl} and InternVL3.5-8B~\citep{wang2025internvl3_5}, with a body-motion pathway following the standard LLaVA/MotionLLM design~\citep{liu2023visual,chen2025motionllm}: a frozen VQ-VAE motion encoder~\citep{zhang2023t2mgpt,guo2022generating} was used to extract continuous latents, which were projected into the language-model embedding space and inserted as motion tokens in the input sequence (see Appendix~\ref{app:training_details} for architecture, training curriculum, and hyperparameters). We fine-tuned each backbone on SOMA-Train (Appendix~\ref{app:soma_train}) in three input settings, motion-only, video-only, and video+motion, and evaluated them on SOMA against the zero-shot video-only backbone (Table~\ref{tab:SOMA_finetune}).

\vspace{-0.2cm}
\paragraph{Results.}
Motion-only fine-tuned MLLMs already perform above chance, suggesting that body motion is informative on its own, though performance remains low: motion alone is often insufficient to infer visual content, such as surrounding scenes or objects. For comparison, existing motion-only models evaluated zero-shot remain close to random chance on SOMA (Appendix~\ref{app:motion-baselines}). Video fine-tuning improves both backbones well above zero-shot performance. Adding aligned motion provides a further but modest improvement for both backbones. However, the effect varies across capability axes: motion slightly improves Body Configuration and Temporal Reasoning for both backbones, while on Motion Dynamics it helps Qwen3-VL-8B but slightly hurts InternVL3.5-8B (see Appendix~\ref{app:training_extra_results} for task-level results). We also found results consistent with these findings when ablating video/motion modalities at test time (Table~\ref{tab:modality_ablation}). Overall, these results suggest that body motion can provide information complementary to the visual signal, but that its benefits depend on the task and remain limited for the provided, vanilla baselines. This motivates future research on multimodal sensory integration.

\section{Conclusion}
We introduced SOMA, a benchmark for fine-grained understanding of human body motion in egocentric video, spanning 12 tasks across three capability axes. Across closed-source, open-source, and specialized egocentric models, we found a large gap to human performance, showing that perceiving the moving body from a first-person view remains an open problem.

We see a natural parallel to the human brain, where visual input from the eyes is complemented by multiple sensors (tactile, vestibular, proprioceptive) to form a coherent representation of the subject and its environment~\citep{blanke2012multisensory,bonnetto2026behavior}. In egocentric video, the body is largely unseen, so a model must likewise infer posture, locomotion, and actions from indirect evidence such as hand pose and ego-motion. We found that current MLLMs underperform in this capacity. We hope that SOMA motivates the development of multimodal systems that better pair first-person video with motion signals to acquire the body awareness current models lack -- and that systems meant to assist, instruct, or learn from humans will ultimately require. We note that beyond MLLMs, emerging world models~\citep{matsuo2022deep} might be ideally suited for this challenge.

\subsection*{AI use statement}
In the preparation of this manuscript, we used generative AI tools for writing assistance. The primary purpose of their use was to aid in editing and polishing the writing of this paper to improve clarity and readability, such as improving grammar and refining sentence structure. All research ideas, methods, analyses, and conclusions in this paper were formulated and developed exclusively by the authors. We have manually reviewed and verified all AI-assisted edits for accuracy and take full responsibility for the final content of this work. In addition, LLMs were used as part of our benchmark construction pipeline to generate candidate question-answer pairs (Sec.~\ref{sec:pipeline}), and all generated questions were human-verified.

\subsection*{Ethics statement}
The ethical considerations and potential risks of this work are limited. SOMA is constructed from two existing public egocentric video datasets, Nymeria and Ego-Exo4D, collected under informed-consent protocols established by their original creators, which we enhance with human-verified novel question-answer annotations; this verification process is anonymized. We used these datasets for benchmark construction and evaluation, following their intended research use. We manually verified the generated QA to ensure quality. Our human evaluation study involved 5 paid participants, who gave informed consent and were compensated for their time. Our experiments evaluate publicly accessible MLLM systems, including released open-source checkpoints and public API models, and are limited to aggregate benchmark performance rather than deployment in real-world decision-making settings. Our analysis also revealed a handedness bias in MLLMs, which future research could address. More broadly, while direct risks are limited, advances in egocentric body-motion understanding could support beneficial applications such as assistive technologies and robotics, but may also raise concerns around surveillance or behavioral profiling.

\subsection*{Reproducibility statement}

The details needed to reproduce our work are provided throughout the main paper and appendix. The SOMA construction pipeline is described in Sec.~\ref{sec:pipeline}, with additional details, including the exact selector, generator, and blind-filtering prompts, given in Appendix~\ref{app:appendix_datacreation}. The human evaluation protocol is detailed in Appendix~\ref{app:human_eval}. The MLLM evaluation setup is explained in Sec.~\ref{sec:experimental-setup}, with specific implementation details given in Appendix~\ref{app:implementation_details}, together with the complete list of evaluated models in Appendix~\ref{app:evaluated-models} and the exact prompts used for evaluation in Appendices~\ref{app:eval_prompt} and~\ref{app:cot_prompt}. The construction of our training dataset is described in Appendix~\ref{app:soma_train}, and our fine-tuning architecture and hyperparameters are given in Appendix~\ref{app:training_details}. Code, benchmark data, and training data will be made publicly available upon paper acceptance.

\subsubsection*{Acknowledgments}

We thank members of the Mathis Group for Computational Neuroscience \& AI (EPFL) for their feedback throughout the project. This work was funded by the Swiss National Science Foundation (SNSF) through grants 320030-227871 (10000950) and 238136.

%\bibliography{bib/custom}
%\bibliographystyle{preprint}

\newpage
\appendix
\part{\normalfont\LARGE\sc\mdseries Appendix}
\vspace{-0.5em}
\noindent\rule{\textwidth}{0.6pt}
\vspace{0.3em}
\parttoc
\vspace{-0.3em}
\noindent\rule{\textwidth}{0.6pt}
\vspace{1em}

\newpage

\raggedbottom
\section{Additional SOMA Examples}
\label{app:examples}
We provide examples for each of the twelve tasks (Figures~\ref{fig:ex-hand-use-1} to \ref{fig:ex-repeated-event-reasoning-1}), each showing eight frames sampled uniformly from the clip, together with the question, its four options, and the correct answer (marked with \checkmark).

\newlength{\demogap}
\setlength{\demogap}{4pt}
\newlength{\demoframewidth}
\setlength{\demoframewidth}{\dimexpr(\textwidth-4\demogap-26pt)/4\relax}
\setlength{\demoframewidth}{0.730\demoframewidth}
\newlength{\demotimelinewidth}
\setlength{\demotimelinewidth}{\dimexpr\textwidth-70pt\relax}
\definecolor{demomuted}{HTML}{6B747D}
\definecolor{demorule}{HTML}{DFE3E7}

\newcommand{\demoul}[2]{\textcolor{#1}{#2}}

\newcommand{\demotask}[3]{%
  \par\vspace{12pt}\noindent
  \textcolor{#1}{\rule[-0.18em]{3pt}{1.05em}}\hspace{7pt}\textbf{#2.}\hspace{6pt}\textbf{#3}%
  \par\vspace{4pt}}

\definecolor{demoframe}{HTML}{1E3A40}
\newcommand{\demosource}[1]{%
  \tikz[baseline]{\node[rounded corners=2.4pt, draw=demoframe, line width=1.1pt,
    fill=white, inner xsep=5pt, inner ysep=2.5pt]
    {\includegraphics[height=13pt]{figures/logos/#1}};}}

\newcommand{\democlock}[1]{%
  \tikz[baseline]{%
    \node[circle, draw=demoframe, line width=1.0pt, fill=white, inner sep=0pt,
          minimum size=22pt, font=\scriptsize\sffamily\bfseries,
          text=demoframe] (c) {#1};
    \draw[demoframe, line width=1.0pt] (c.north) -- ++(0,-2.6pt);}}
\newtcolorbox{democard}[4]{enhanced, breakable=false, arc=3mm, boxrule=1.1pt,
  colframe=demoframe, colback=white, left=5pt, right=5pt, top=10pt, bottom=4pt,
  boxsep=0pt, before skip=4pt, after skip=6pt, fontupper=\sffamily,
  attach boxed title to top center={yshift=-\tcboxedtitleheight/2},
  boxed title style={boxrule=1.1pt, arc=2.4mm, colframe=demoframe,
                     colback=white, left=7pt, right=7pt, top=1.5pt, bottom=1.5pt},
  coltitle=#1, fonttitle=\small\sffamily\bfseries, title={#2},
  overlay unbroken={%
    \node[anchor=center, xshift=28pt] at (frame.north west) {\demosource{#3}};
    \node[anchor=center, xshift=-28pt] at (frame.north east) {\democlock{#4}};}}

\newcommand{\demoimg}[1]{%
  \begin{tikzpicture}
    \path[clip, rounded corners=9pt]
      (0,0) rectangle (\demoframewidth,\demoframewidth);
    \node[anchor=south west, inner sep=0pt] at (0,0)
      {\includegraphics[width=\demoframewidth]{#1}};
  \end{tikzpicture}}

\usetikzlibrary{calc}
\definecolor{demofilm}{HTML}{15191C}
\newcommand{\demostrip}[1]{%
  \begin{tikzpicture}
    \node[inner xsep=5pt, fill=demofilm,
          rounded corners=3pt, inner ysep=5pt] (g) {#1};
    \foreach \p in {0.018,0.058,...,0.99}{%
      \fill[white, rounded corners=0.9pt]
        ($(g.north west)!\p!(g.north east) + (-1.7pt,-4.6pt)$)
        rectangle ++(3.4pt,3.2pt);
      \fill[white, rounded corners=0.9pt]
        ($(g.south west)!\p!(g.south east) + (-1.7pt,1.4pt)$)
        rectangle ++(3.4pt,3.2pt);}
  \end{tikzpicture}}

\newcommand{\demotimeline}[2]{
  \par\vspace{2pt}
  \begin{tikzpicture}[x=\demotimelinewidth, y=1pt]
    \draw[demoframe!30, line width=0.7pt] (0,0) -- (1,0);
    \foreach \t in {#2} {
      \pgfmathsetmacro\xp{\t/#1}
      \draw[demoframe!45, line width=0.7pt] (\xp,-2.4) -- (\xp,2.4);
      \node[below=2.5pt, font=\tiny\sffamily, text=demomuted] at (\xp,-2.4)
        {\t\,s};}
    \node[left=3pt, font=\tiny\sffamily\bfseries, text=demoframe] at (0,0) {0};
    \node[right=3pt, font=\tiny\sffamily\bfseries, text=demoframe] at (1,0) {#1\,s};
  \end{tikzpicture}\par\vspace{1pt}}

\newcommand{\demoq}[1]{\par\vspace{4pt}{\centering\textbf{#1}\par}\vspace{2pt}}
\newcommand{\demopt}[3]{
  \ifx\relax#3\relax\textbf{#1.}~#2\else
  \textbf{#1.}~\textcolor{#3}{\textbf{#2}}~\textcolor{#3}{\checkmark}\fi}

\newenvironment{excard}
  {\par\noindent\begin{minipage}{\linewidth}\centering}
  {\end{minipage}\par}

\setlength{\abovecaptionskip}{5pt}
\setlength{\belowcaptionskip}{0pt}
\vspace*{14pt}

\begin{excard}
\begin{democard}{bodyconfigcolor}{Hand Use}{ee4d_mark.png}{36s}
{\centering\setlength{\tabcolsep}{0pt}\renewcommand{\arraystretch}{1}%
\demostrip{%
\begin{tabular}{@{}c@{\hspace{\demogap}}c@{\hspace{\demogap}}c@{\hspace{\demogap}}c@{}}
\demoimg{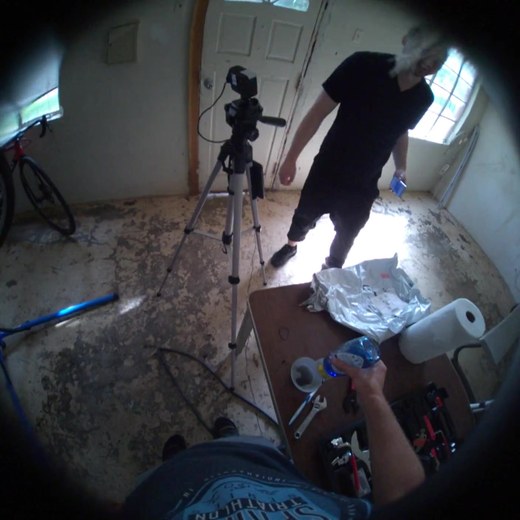} & \demoimg{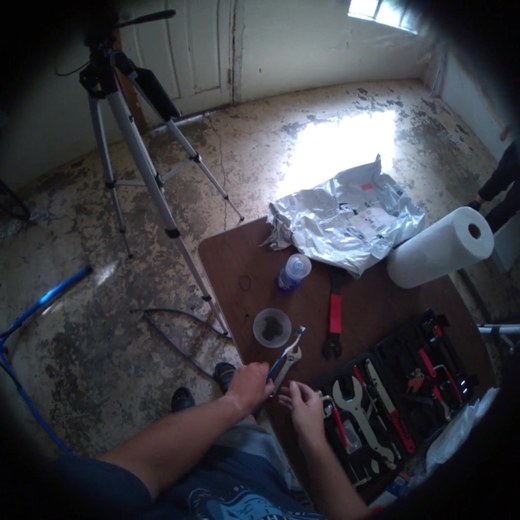} & \demoimg{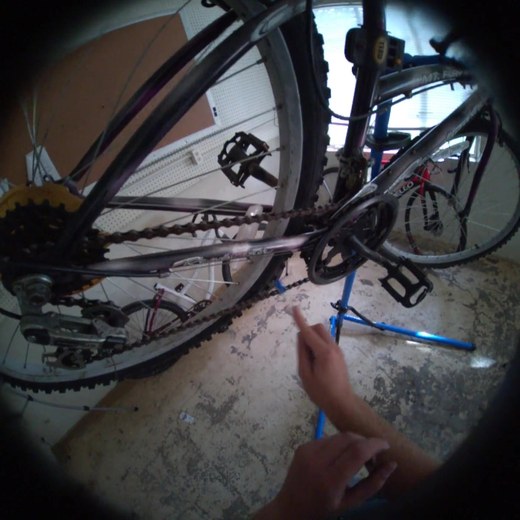} & \demoimg{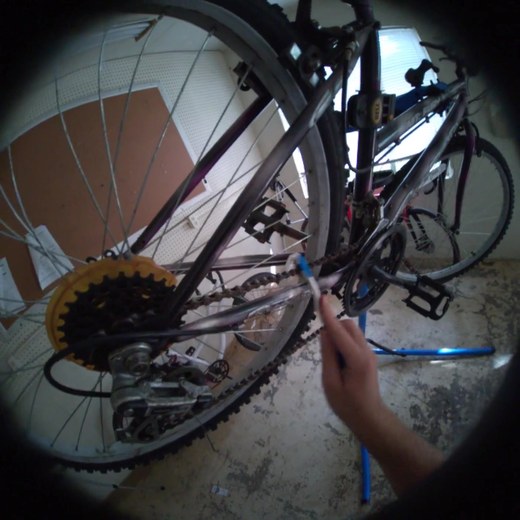}
\end{tabular}}\par\vspace{2pt}
\demostrip{%
\begin{tabular}{@{}c@{\hspace{\demogap}}c@{\hspace{\demogap}}c@{\hspace{\demogap}}c@{}}
\demoimg{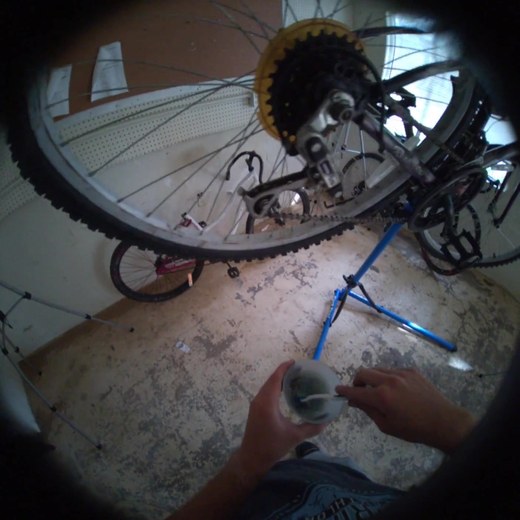} & \demoimg{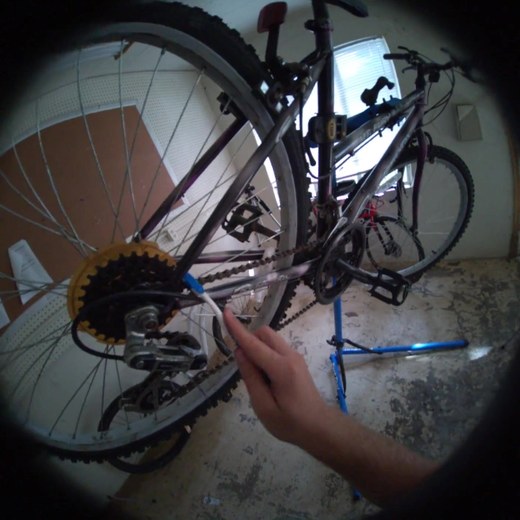} & \demoimg{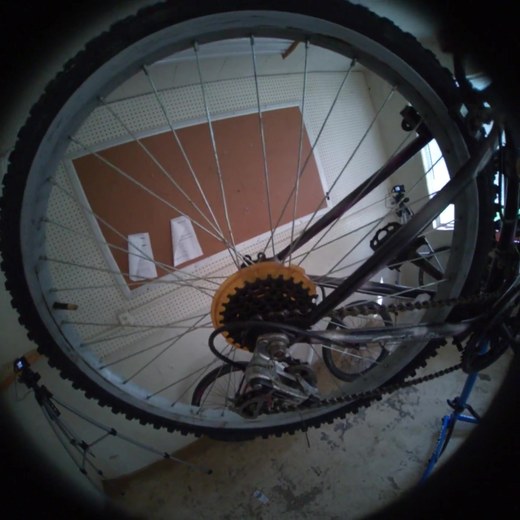} & \demoimg{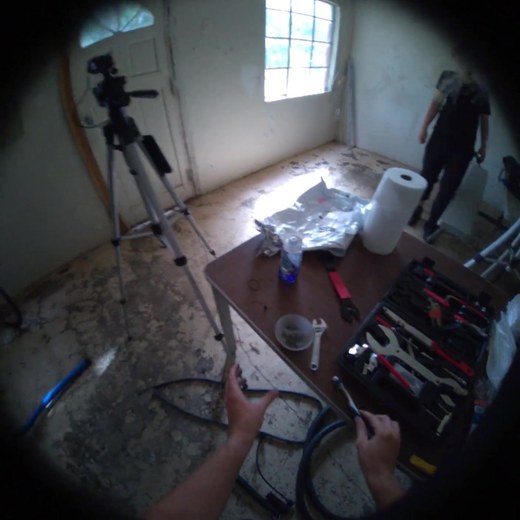}
\end{tabular}}\par}
\demotimeline{36}{0.0,5.2,10.2,15.5,20.8,26.0,31.0}
\demoq{Which hand does the person use to brush the chain?}
{\centering
\begin{tabular}{@{}l@{\hspace{30pt}}l@{}}
\demopt{A}{Right hand}{bodyconfigcolor} & \demopt{B}{Neither hand}{} \\[3pt]
\demopt{C}{Both hands}{} & \demopt{D}{Left hand}{}
\end{tabular}\par}
\end{democard}
\vspace{-2pt}\captionof{figure}{\small\textbf{Hand Use.} Ego-Exo4D clip, 36\,s.}
\label{fig:ex-hand-use-1}
\end{excard}

\begin{excard}
\begin{democard}{bodyconfigcolor}{Hand Use}{nymeria_mark.png}{30s}
{\centering\setlength{\tabcolsep}{0pt}\renewcommand{\arraystretch}{1}%
\demostrip{%
\begin{tabular}{@{}c@{\hspace{\demogap}}c@{\hspace{\demogap}}c@{\hspace{\demogap}}c@{}}
\demoimg{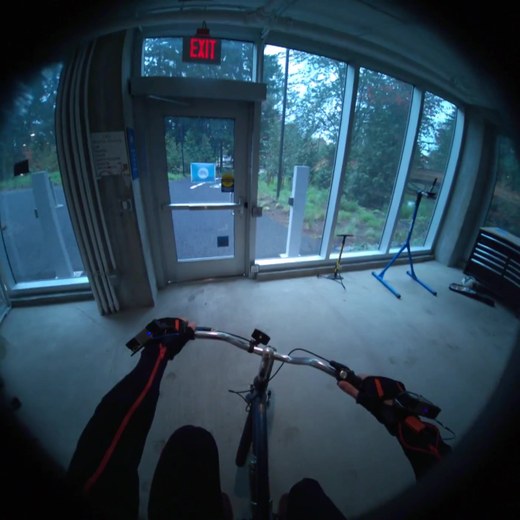} & \demoimg{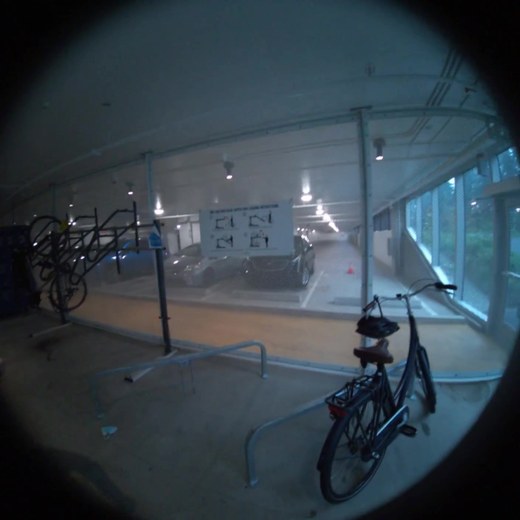} & \demoimg{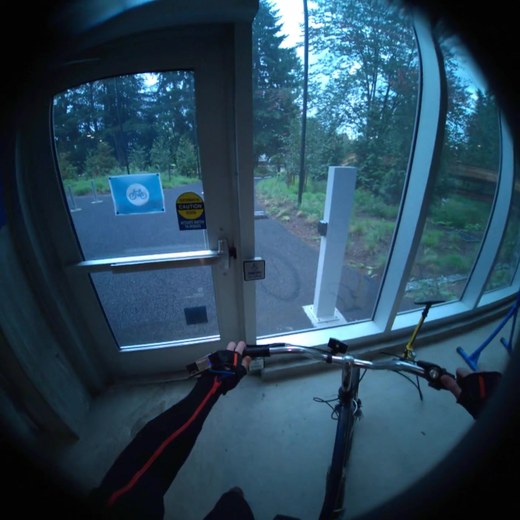} & \demoimg{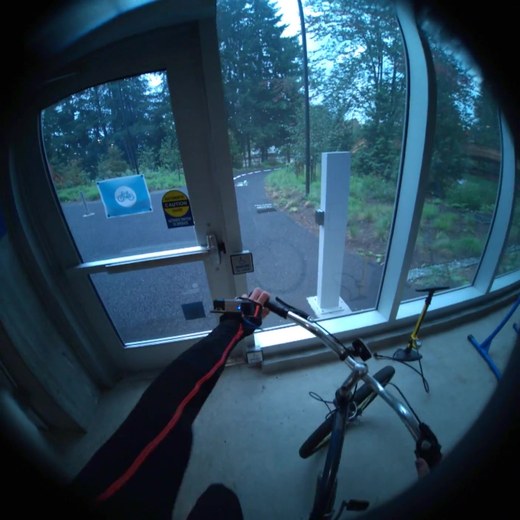}
\end{tabular}}\par\vspace{2pt}
\demostrip{%
\begin{tabular}{@{}c@{\hspace{\demogap}}c@{\hspace{\demogap}}c@{\hspace{\demogap}}c@{}}
\demoimg{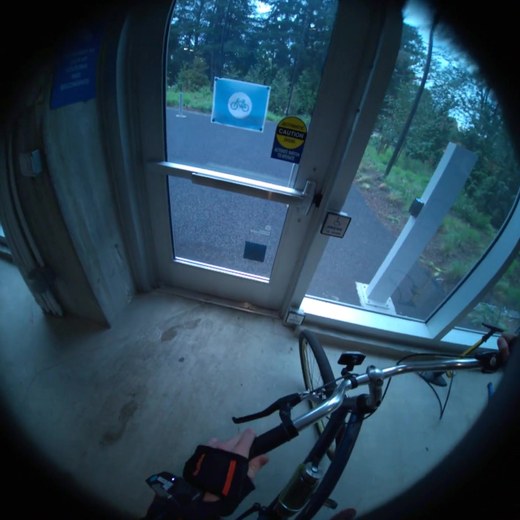} & \demoimg{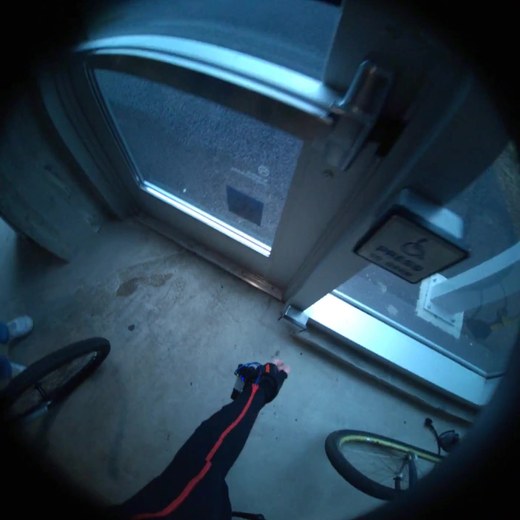} & \demoimg{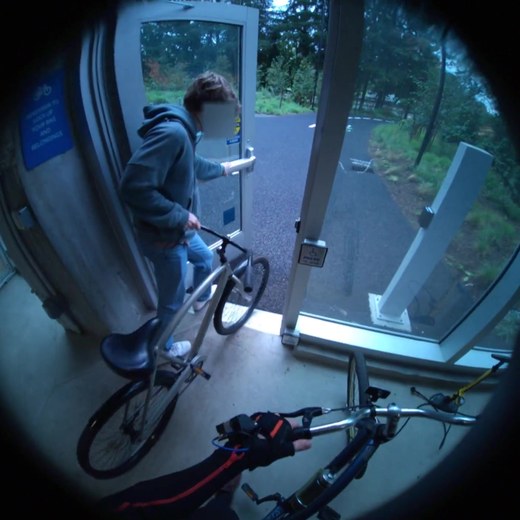} & \demoimg{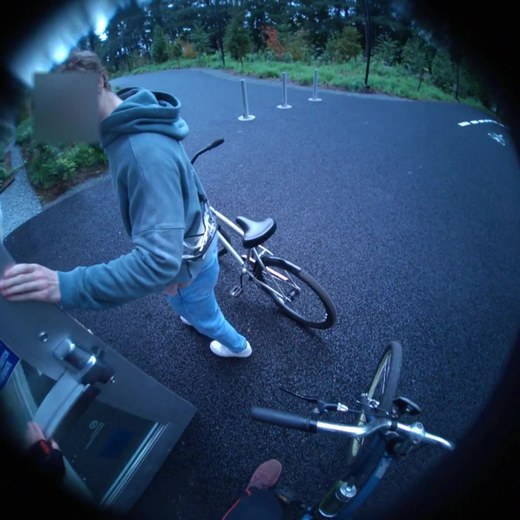}
\end{tabular}}\par}
\demotimeline{30}{0.0,4.2,8.5,12.8,17.0,21.2,25.5,29.8}
\demoq{Which hand does the person use to press the door lock?}
{\centering
\begin{tabular}{@{}l@{\hspace{30pt}}l@{}}
\demopt{A}{Neither hand}{} & \demopt{B}{Right hand}{} \\[3pt]
\demopt{C}{Left hand}{bodyconfigcolor} & \demopt{D}{Both hands}{}
\end{tabular}\par}
\end{democard}
\vspace{-2pt}\captionof{figure}{\small\textbf{Hand Use.} Nymeria clip, 30\,s.}
\label{fig:ex-hand-use-2}
\end{excard}

\begin{excard}
\begin{democard}{bodyconfigcolor}{Hand Use}{nymeria_mark.png}{10s}
{\centering\setlength{\tabcolsep}{0pt}\renewcommand{\arraystretch}{1}%
\demostrip{%
\begin{tabular}{@{}c@{\hspace{\demogap}}c@{\hspace{\demogap}}c@{\hspace{\demogap}}c@{}}
\demoimg{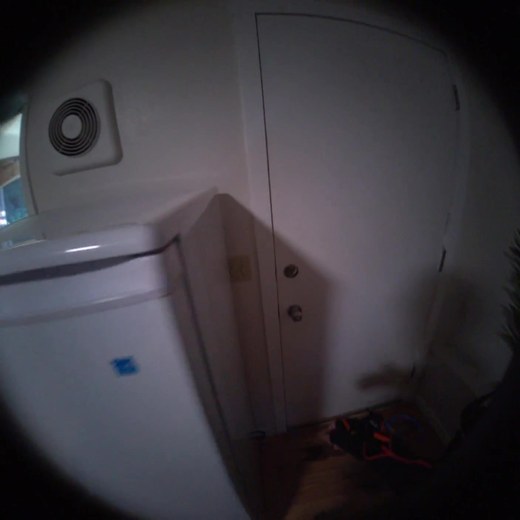} & \demoimg{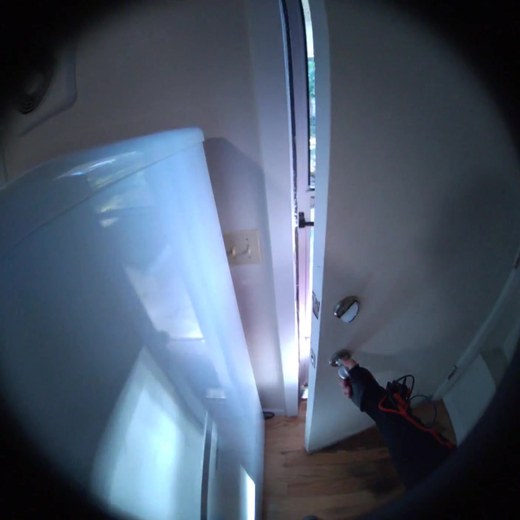} & \demoimg{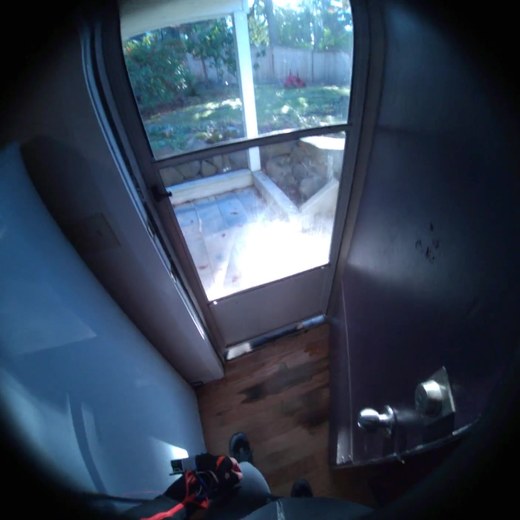} & \demoimg{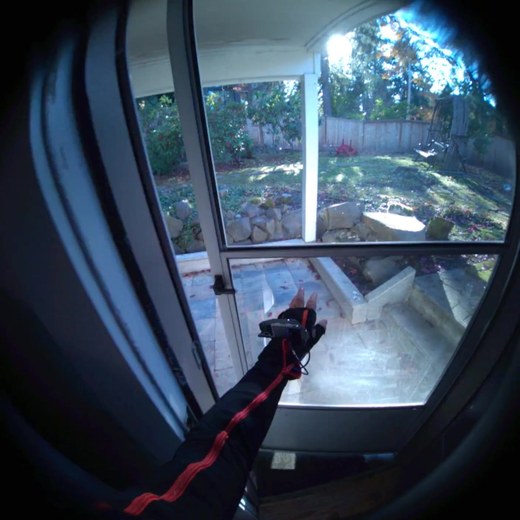}
\end{tabular}}\par\vspace{2pt}
\demostrip{%
\begin{tabular}{@{}c@{\hspace{\demogap}}c@{\hspace{\demogap}}c@{\hspace{\demogap}}c@{}}
\demoimg{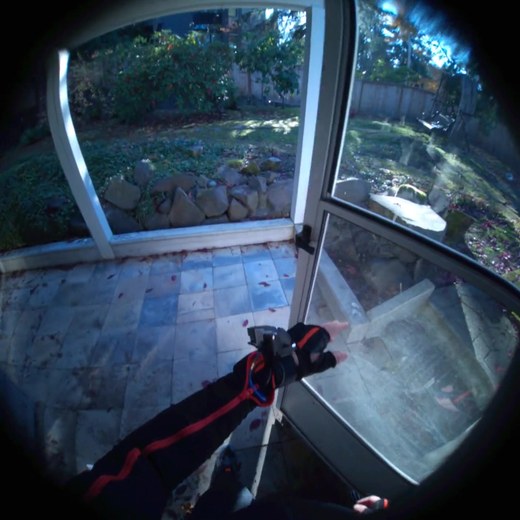} & \demoimg{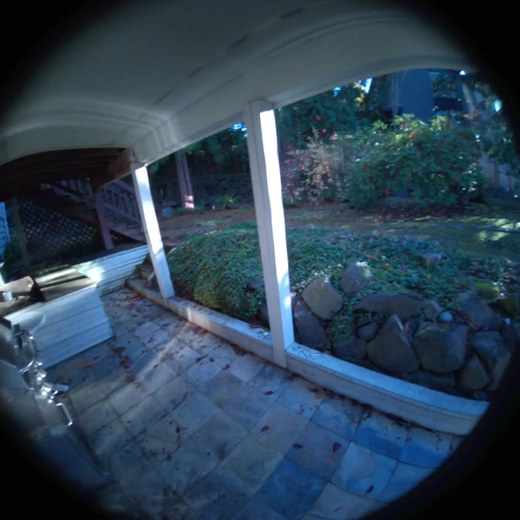} & \demoimg{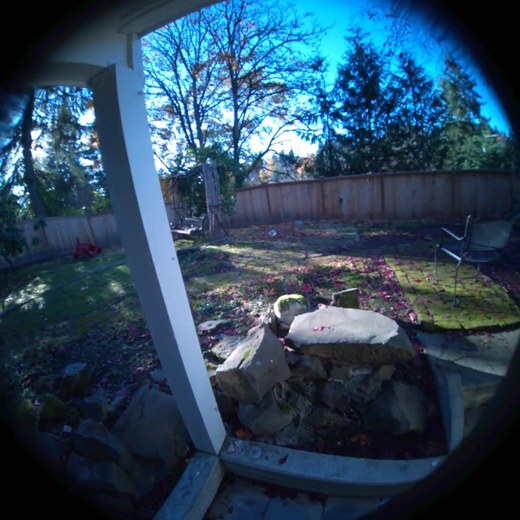} & \demoimg{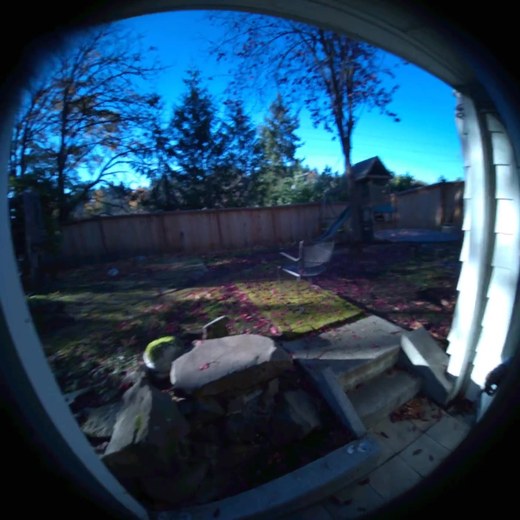}
\end{tabular}}\par}
\demotimeline{10}{0.0,1.5,2.8,4.2,5.5,7.0,8.2,9.8}
\demoq{Which hand does the person use to push the door?}
{\centering
\begin{tabular}{@{}l@{\hspace{30pt}}l@{}}
\demopt{A}{Neither hand}{} & \demopt{B}{Right hand}{} \\[3pt]
\demopt{C}{Left hand}{bodyconfigcolor} & \demopt{D}{Both hands}{}
\end{tabular}\par}
\end{democard}
\vspace{-2pt}\captionof{figure}{\small\textbf{Hand Use.} Nymeria clip, 10\,s.}
\label{fig:ex-hand-use-3}
\end{excard}

\begin{excard}
\begin{democard}{bodyconfigcolor}{Bimanual Coordination}{nymeria_mark.png}{18s}
{\centering\setlength{\tabcolsep}{0pt}\renewcommand{\arraystretch}{1}%
\demostrip{%
\begin{tabular}{@{}c@{\hspace{\demogap}}c@{\hspace{\demogap}}c@{\hspace{\demogap}}c@{}}
\demoimg{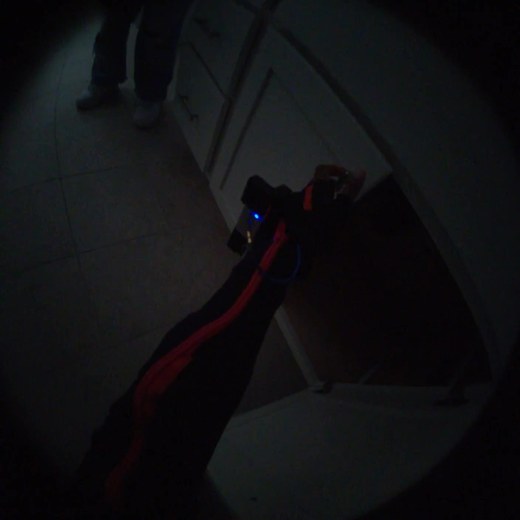} & \demoimg{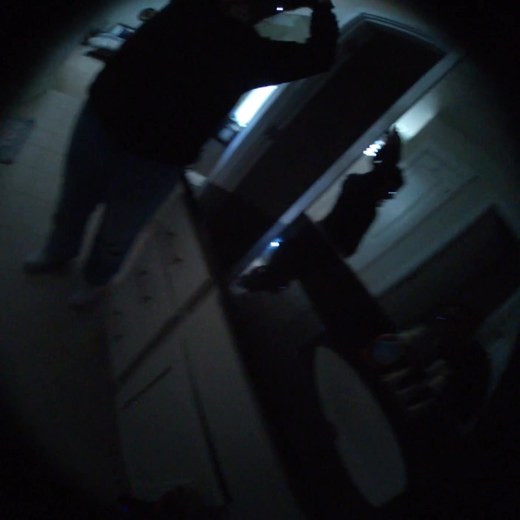} & \demoimg{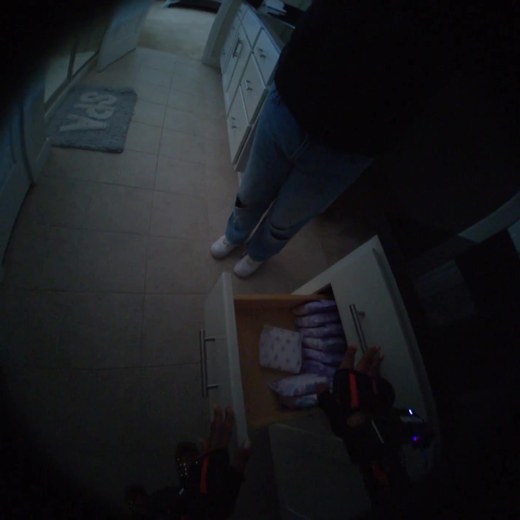} & \demoimg{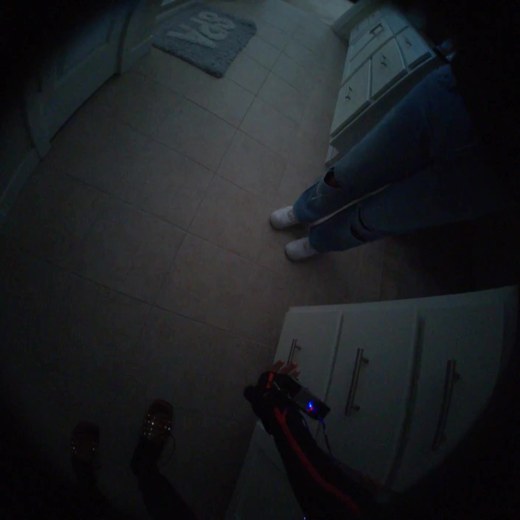}
\end{tabular}}\par\vspace{2pt}
\demostrip{%
\begin{tabular}{@{}c@{\hspace{\demogap}}c@{\hspace{\demogap}}c@{\hspace{\demogap}}c@{}}
\demoimg{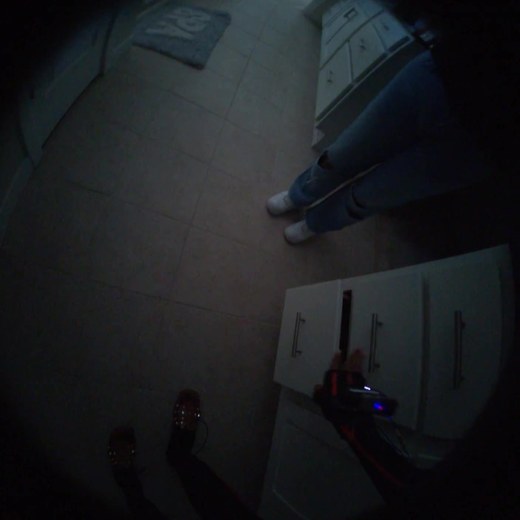} & \demoimg{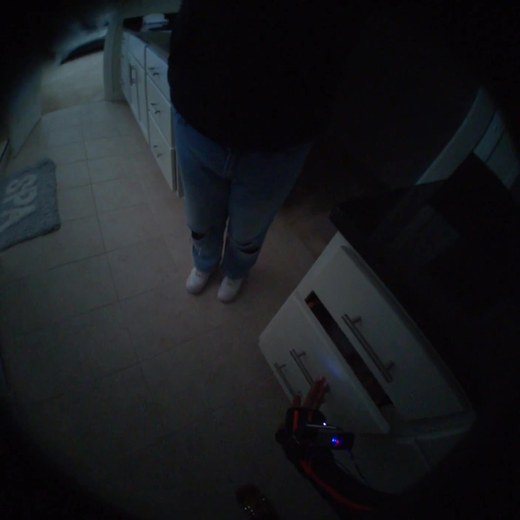} & \demoimg{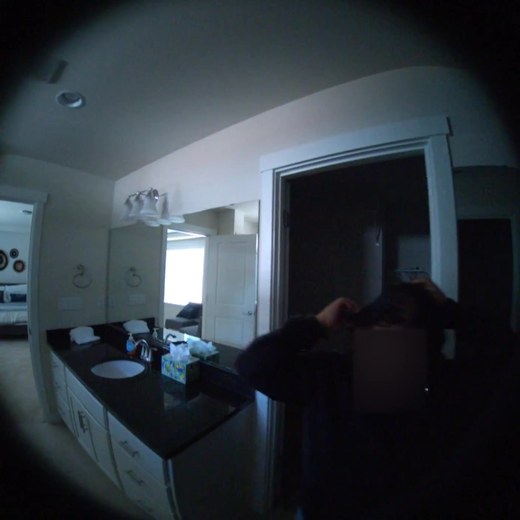} & \demoimg{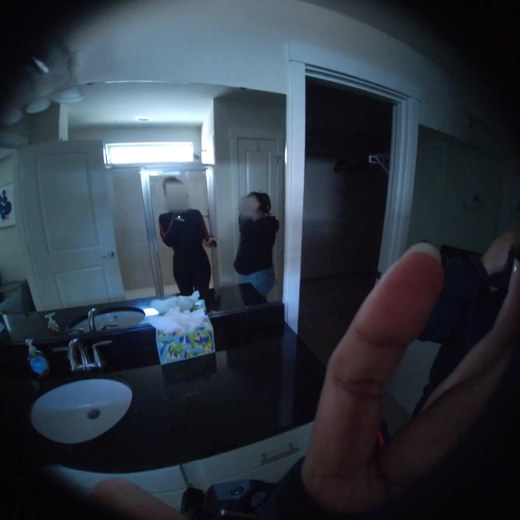}
\end{tabular}}\par}
\demotimeline{18}{0.0,2.5,5.2,7.8,10.2,12.8,15.5,18.0}
\demoq{While the person pulls the top drawer with one hand, what is the other hand doing at that moment?}
{\centering\small
\begin{tabular}{@{}l@{\hspace{30pt}}l@{}}
\demopt{A}{Holding the middle drawer}{bodyconfigcolor} & \demopt{B}{Holding the counter for stability}{} \\[3pt]
\demopt{C}{Holding a box of napkins}{} & \demopt{D}{Reaching for an item on the counter}{}
\end{tabular}\par}
\end{democard}
\vspace{-2pt}\captionof{figure}{\small\textbf{Bimanual Coordination.} Nymeria clip, 18\,s.}
\label{fig:ex-bimanual-coordination-1}
\end{excard}

\begin{excard}
\begin{democard}{bodyconfigcolor}{Bimanual Coordination}{nymeria_mark.png}{35s}
{\centering\setlength{\tabcolsep}{0pt}\renewcommand{\arraystretch}{1}%
\demostrip{%
\begin{tabular}{@{}c@{\hspace{\demogap}}c@{\hspace{\demogap}}c@{\hspace{\demogap}}c@{}}
\demoimg{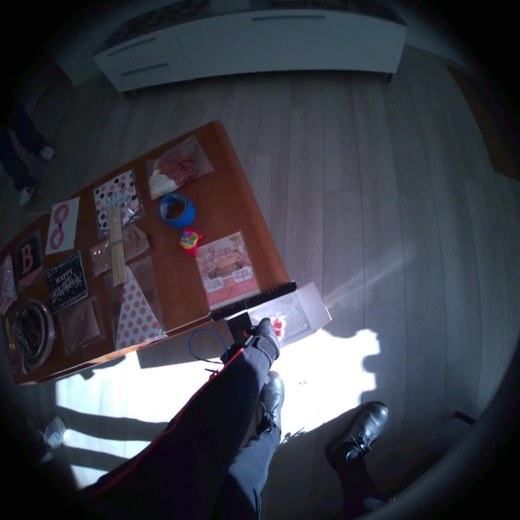} & \demoimg{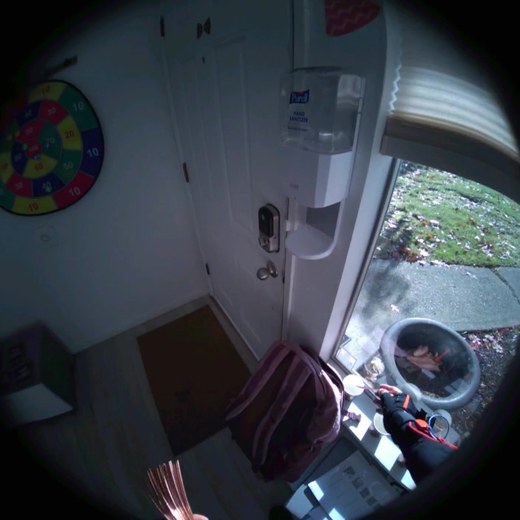} & \demoimg{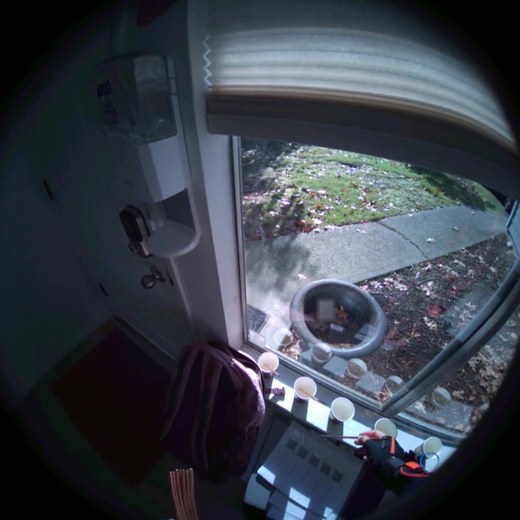} & \demoimg{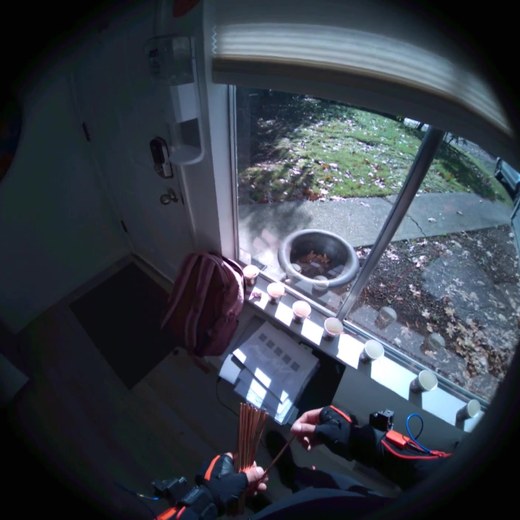}
\end{tabular}}\par\vspace{2pt}
\demostrip{%
\begin{tabular}{@{}c@{\hspace{\demogap}}c@{\hspace{\demogap}}c@{\hspace{\demogap}}c@{}}
\demoimg{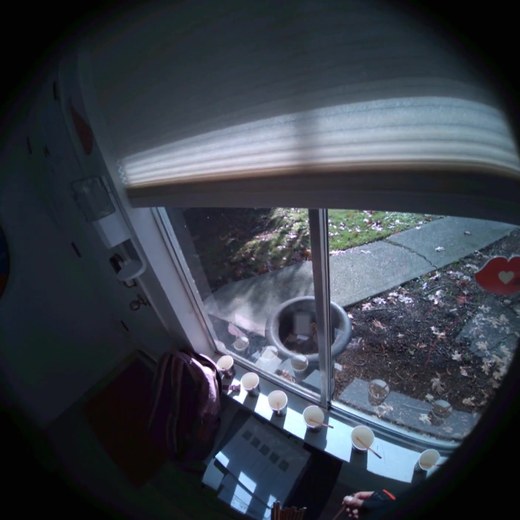} & \demoimg{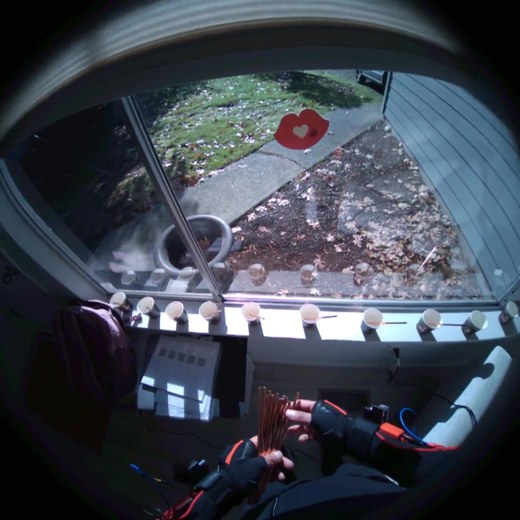} & \demoimg{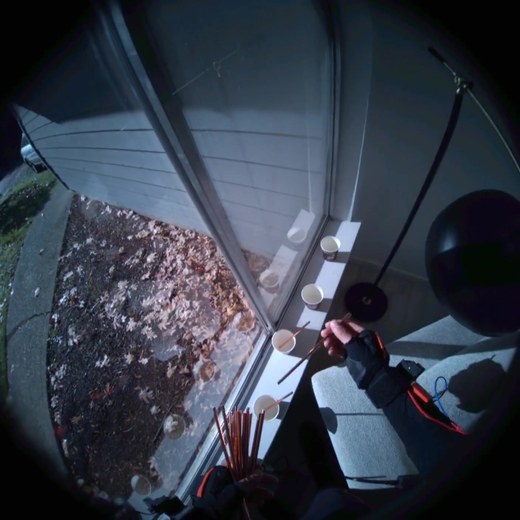} & \demoimg{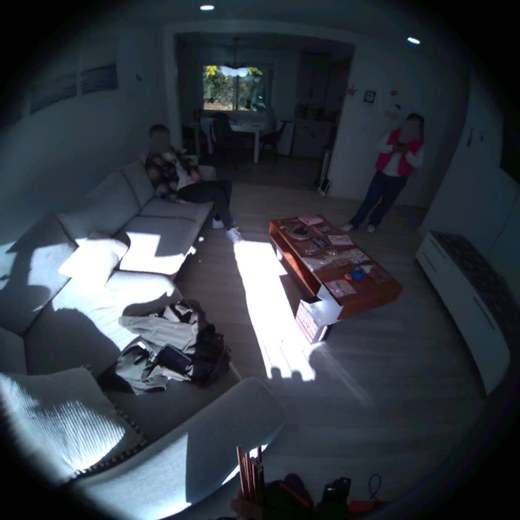}
\end{tabular}}\par}
\demotimeline{35}{0.0,5.0,10.0,15.0,19.8,24.8,29.8,34.8}
\demoq{While the person puts straws into a cup with one hand, what is the other hand doing at that moment?}
{\centering
\begin{tabular}{@{}l@{\hspace{30pt}}l@{}}
\demopt{A}{Reaching for another cup}{} & \demopt{B}{Holding the cup steady}{} \\[3pt]
\demopt{C}{Holding the remaining straws}{bodyconfigcolor} & \demopt{D}{Resting at their side}{}
\end{tabular}\par}
\end{democard}
\vspace{-2pt}\captionof{figure}{\small\textbf{Bimanual Coordination.} Nymeria clip, 35\,s.}
\label{fig:ex-bimanual-coordination-2}
\end{excard}

\begin{excard}
\begin{democard}{bodyconfigcolor}{Body Posture}{nymeria_mark.png}{82s}
{\centering\setlength{\tabcolsep}{0pt}\renewcommand{\arraystretch}{1}%
\demostrip{%
\begin{tabular}{@{}c@{\hspace{\demogap}}c@{\hspace{\demogap}}c@{\hspace{\demogap}}c@{}}
\demoimg{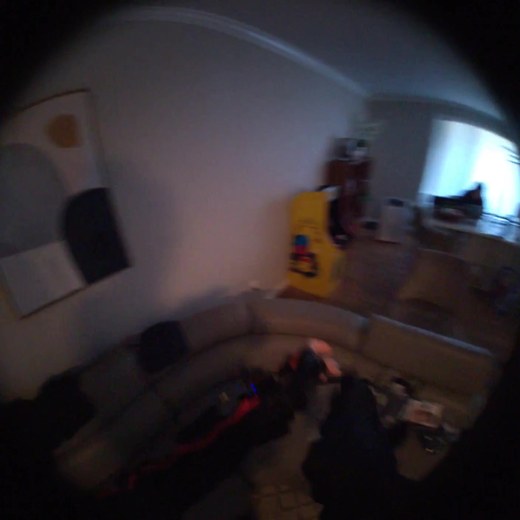} & \demoimg{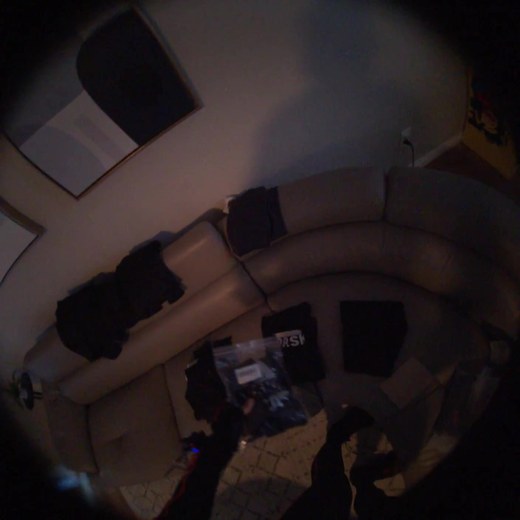} & \demoimg{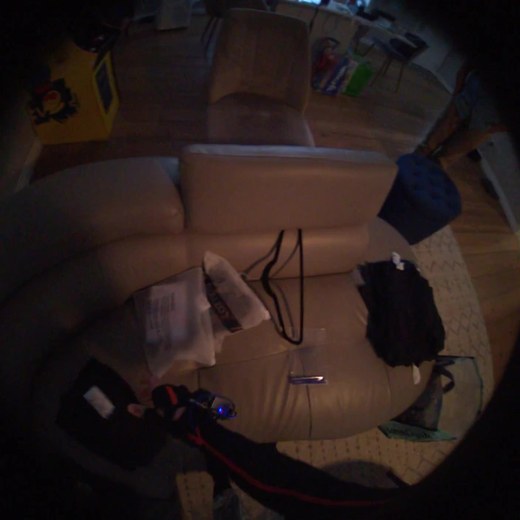} & \demoimg{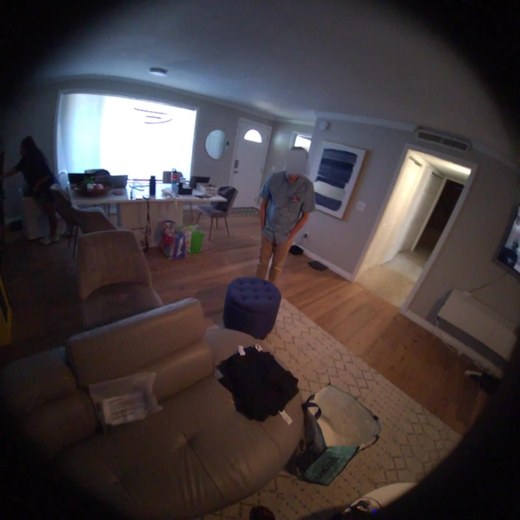}
\end{tabular}}\par\vspace{2pt}
\demostrip{%
\begin{tabular}{@{}c@{\hspace{\demogap}}c@{\hspace{\demogap}}c@{\hspace{\demogap}}c@{}}
\demoimg{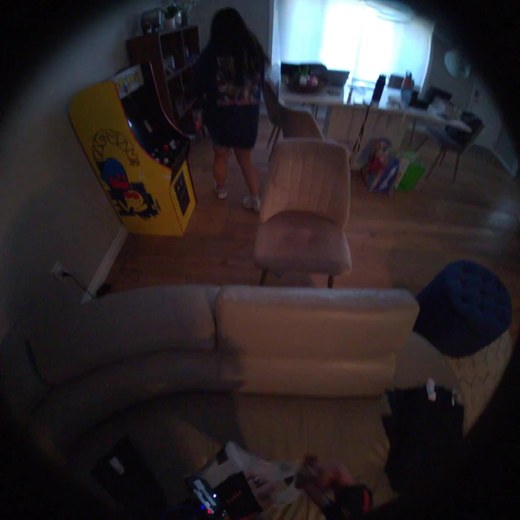} & \demoimg{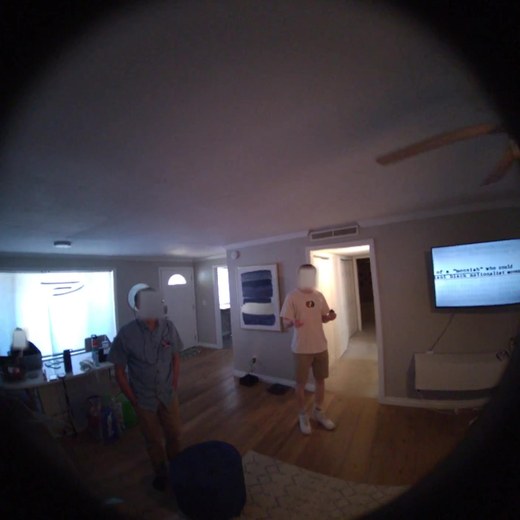} & \demoimg{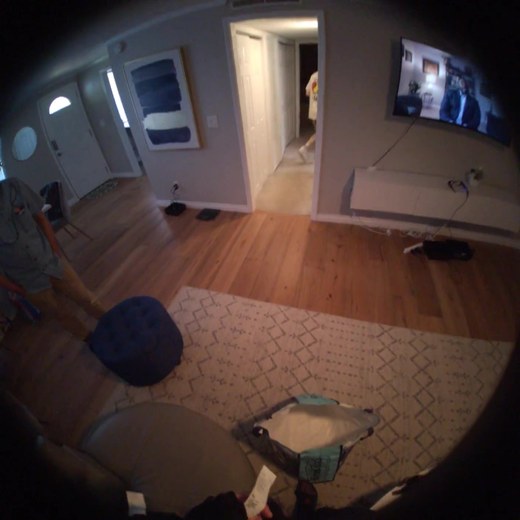} & \demoimg{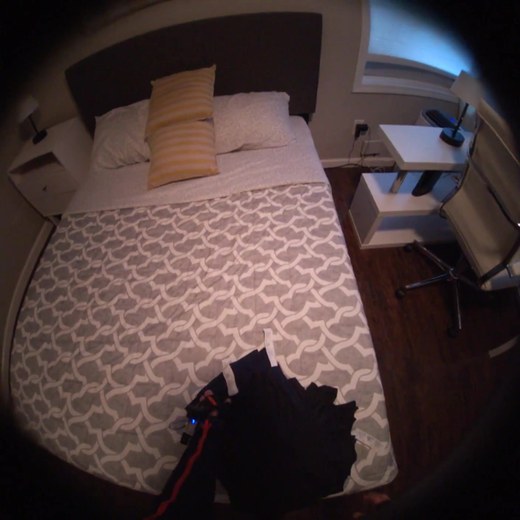}
\end{tabular}}\par}
\demotimeline{82}{0.0,11.5,23.2,34.8,46.5,58.0,69.8,81.2}
\demoq{What body posture is the person in at the end of the clip?}
{\centering
\begin{tabular}{@{}l@{\hspace{30pt}}l@{}}
\demopt{A}{Crouching down}{} & \demopt{B}{Standing straight}{} \\[3pt]
\demopt{C}{Walking}{} & \demopt{D}{Bending forward}{bodyconfigcolor}
\end{tabular}\par}
\end{democard}
\vspace{-2pt}\captionof{figure}{\small\textbf{Body Posture.} Nymeria clip, 82\,s.}
\label{fig:ex-body-posture-1}
\end{excard}

\begin{excard}
\begin{democard}{bodyconfigcolor}{Concurrent Body-Part State}{nymeria_mark.png}{41s}
{\centering\setlength{\tabcolsep}{0pt}\renewcommand{\arraystretch}{1}%
\demostrip{%
\begin{tabular}{@{}c@{\hspace{\demogap}}c@{\hspace{\demogap}}c@{\hspace{\demogap}}c@{}}
\demoimg{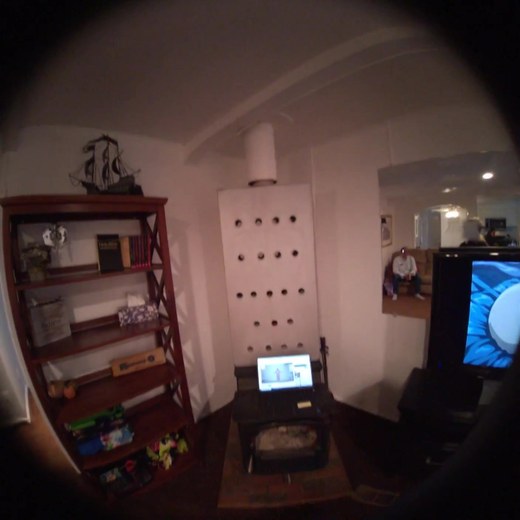} & \demoimg{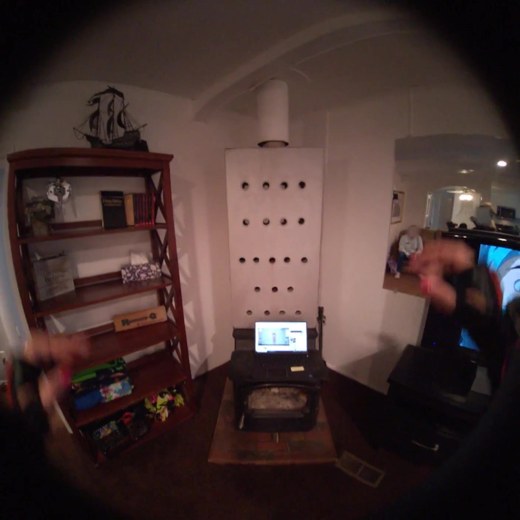} & \demoimg{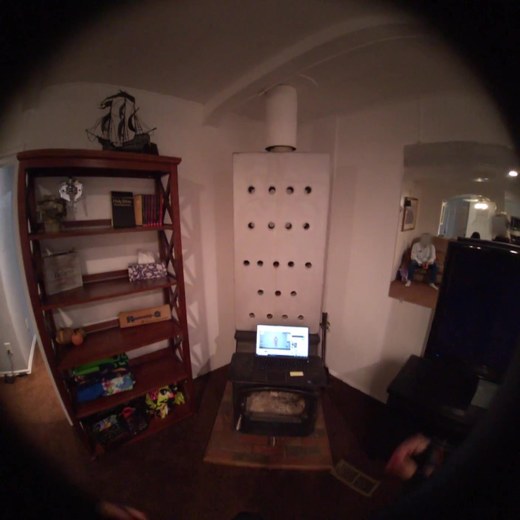} & \demoimg{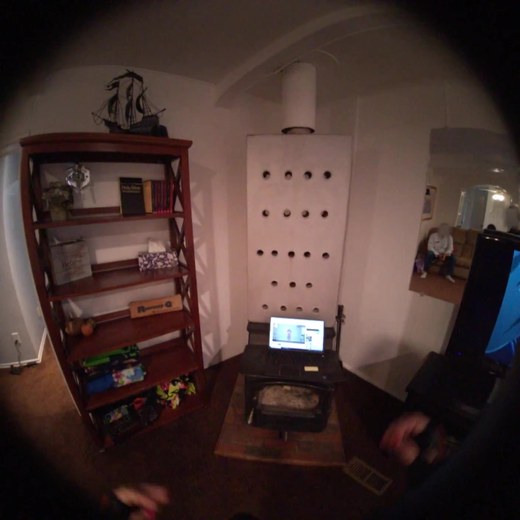}
\end{tabular}}\par\vspace{2pt}
\demostrip{%
\begin{tabular}{@{}c@{\hspace{\demogap}}c@{\hspace{\demogap}}c@{\hspace{\demogap}}c@{}}
\demoimg{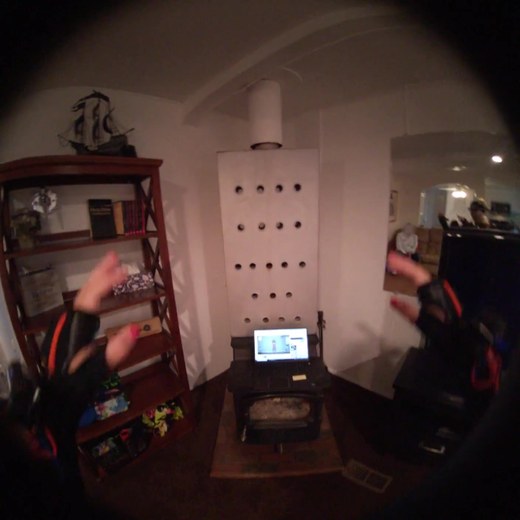} & \demoimg{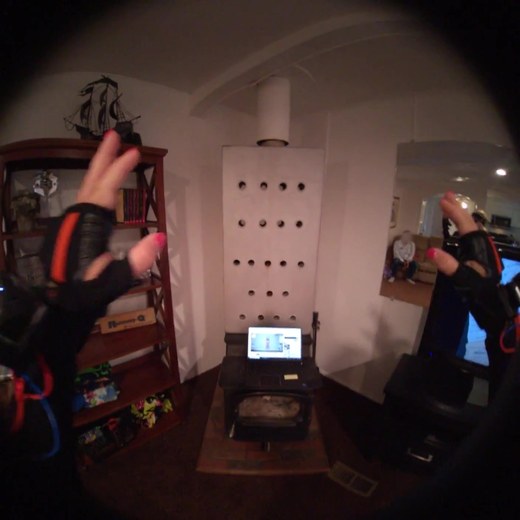} & \demoimg{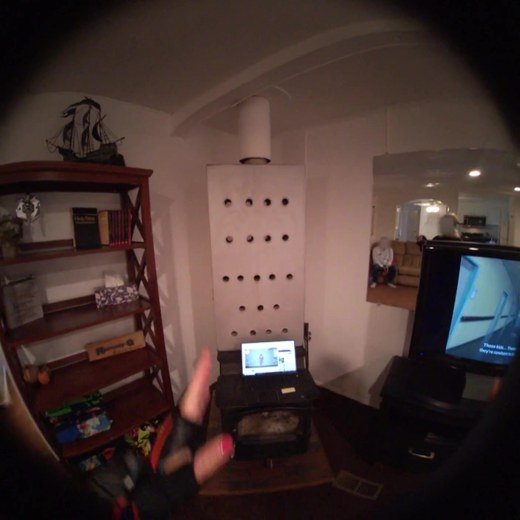} & \demoimg{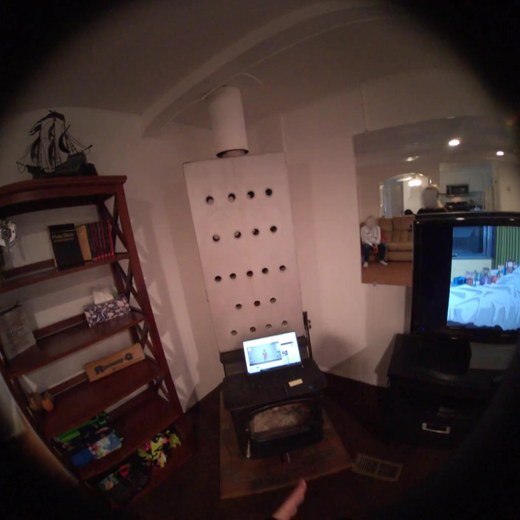}
\end{tabular}}\par}
\demotimeline{41}{0.0,5.8,11.8,17.5,23.2,29.0,35.0,40.8}
\demoq{What are the person's legs doing while the person alternately extends both of their arms upwards and lowers them down near their midsection?}
{\centering\small
\begin{tabular}{@{}l@{\hspace{30pt}}l@{}}
\demopt{A}{Alternately lifting both of their legs}{bodyconfigcolor} & \demopt{B}{Taking small steps in place}{} \\[3pt]
\demopt{C}{Shifting weight from side to side}{} & \demopt{D}{Standing still with feet planted}{}
\end{tabular}\par}
\end{democard}
\vspace{-2pt}\captionof{figure}{\small\textbf{Concurrent Body-Part State.} Nymeria clip, 41\,s.}
\label{fig:ex-concurrent-body-part-state-1}
\end{excard}

\begin{excard}
\begin{democard}{motiondynamicscolor}{Repetition Counting}{ee4d_mark.png}{4s}
{\centering\setlength{\tabcolsep}{0pt}\renewcommand{\arraystretch}{1}%
\demostrip{%
\begin{tabular}{@{}c@{\hspace{\demogap}}c@{\hspace{\demogap}}c@{\hspace{\demogap}}c@{}}
\demoimg{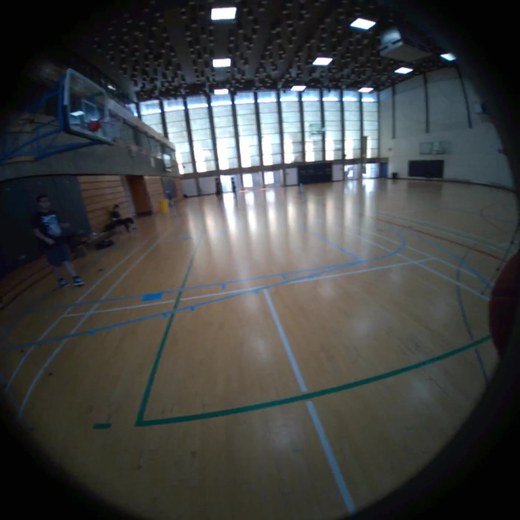} & \demoimg{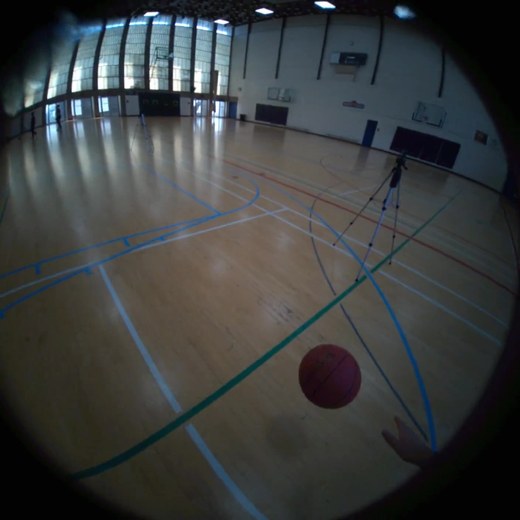} & \demoimg{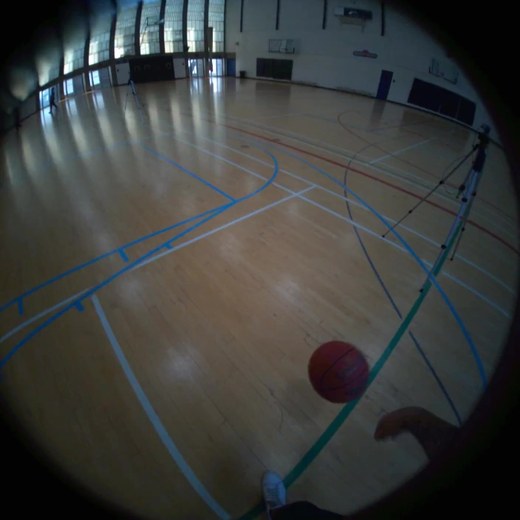} & \demoimg{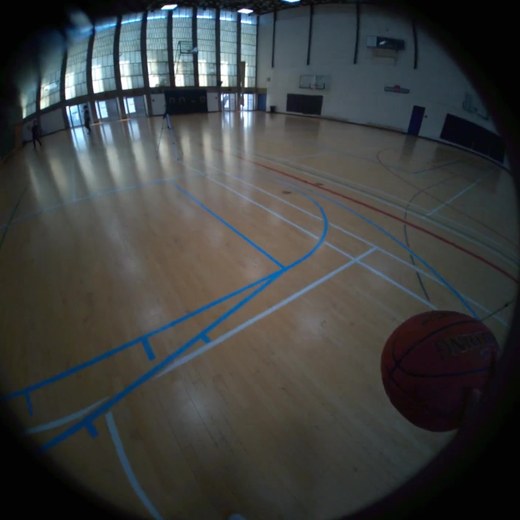}
\end{tabular}}\par\vspace{2pt}
\demostrip{%
\begin{tabular}{@{}c@{\hspace{\demogap}}c@{\hspace{\demogap}}c@{\hspace{\demogap}}c@{}}
\demoimg{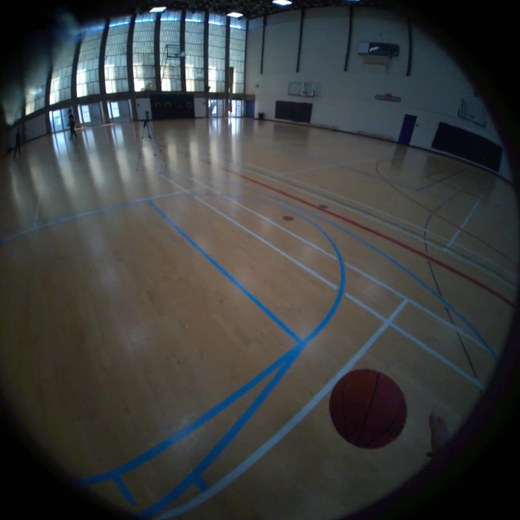} & \demoimg{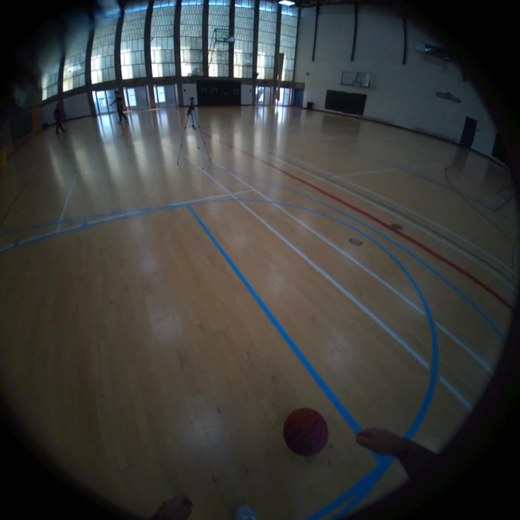} & \demoimg{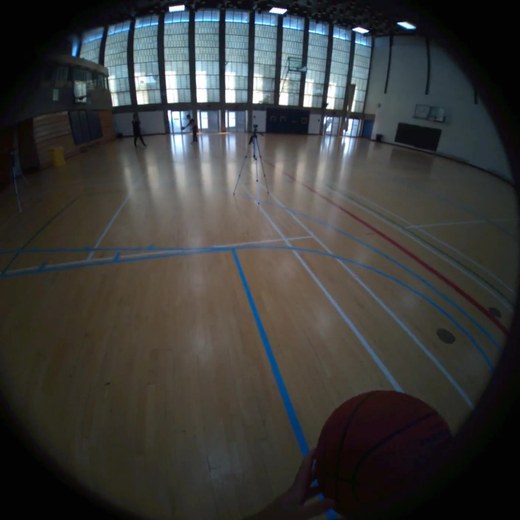} & \demoimg{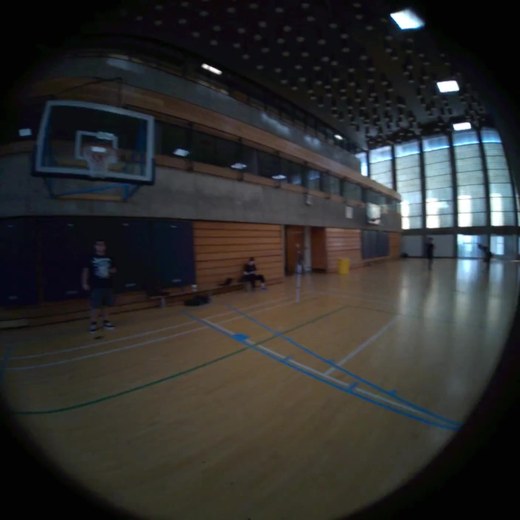}
\end{tabular}}\par}
\demotimeline{4}{0.0,0.5,1.0,1.5,2.0,2.5,3.0,3.5}
\demoq{How many times did the person bounce the ball on the floor with their right hand?}
{\centering
\begin{tabular}{@{}l@{\hspace{30pt}}l@{}}
\demopt{A}{5}{} & \demopt{B}{3}{} \\[3pt]
\demopt{C}{6}{} & \demopt{D}{4}{motiondynamicscolor}
\end{tabular}\par}
\end{democard}
\vspace{-2pt}\captionof{figure}{\small\textbf{Repetition Counting.} Ego-Exo4D clip, 4\,s.}
\label{fig:ex-repetition-counting-1}
\end{excard}

\begin{excard}
\begin{democard}{motiondynamicscolor}{Repetition Counting}{ee4d_mark.png}{21s}
{\centering\setlength{\tabcolsep}{0pt}\renewcommand{\arraystretch}{1}%
\demostrip{%
\begin{tabular}{@{}c@{\hspace{\demogap}}c@{\hspace{\demogap}}c@{\hspace{\demogap}}c@{}}
\demoimg{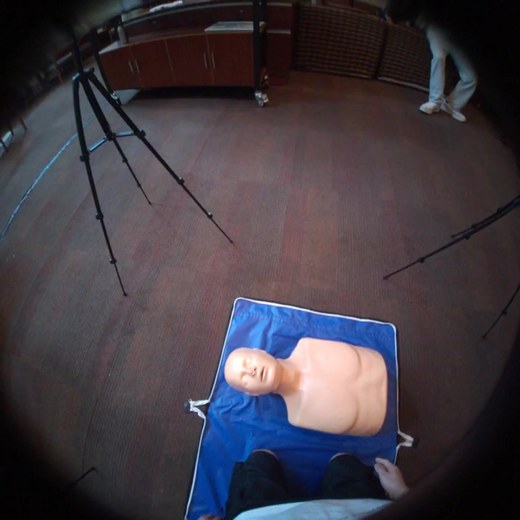} & \demoimg{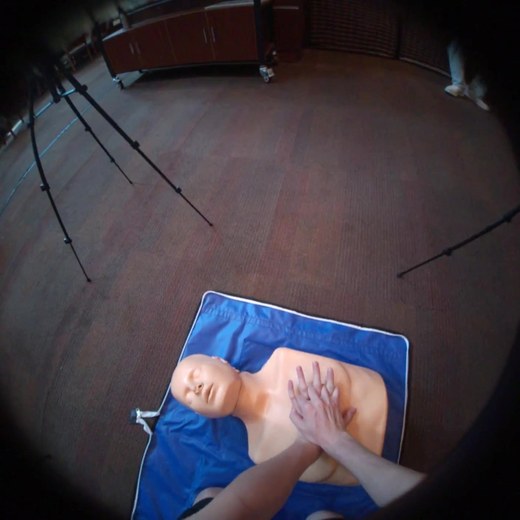} & \demoimg{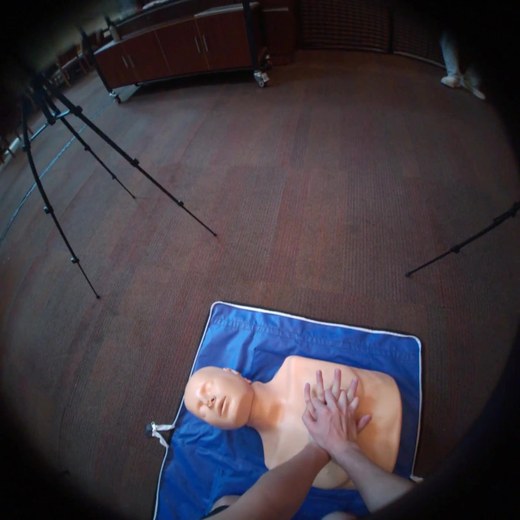} & \demoimg{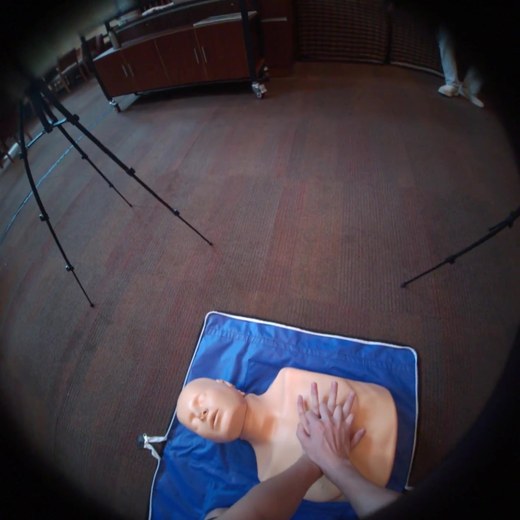}
\end{tabular}}\par\vspace{2pt}
\demostrip{%
\begin{tabular}{@{}c@{\hspace{\demogap}}c@{\hspace{\demogap}}c@{\hspace{\demogap}}c@{}}
\demoimg{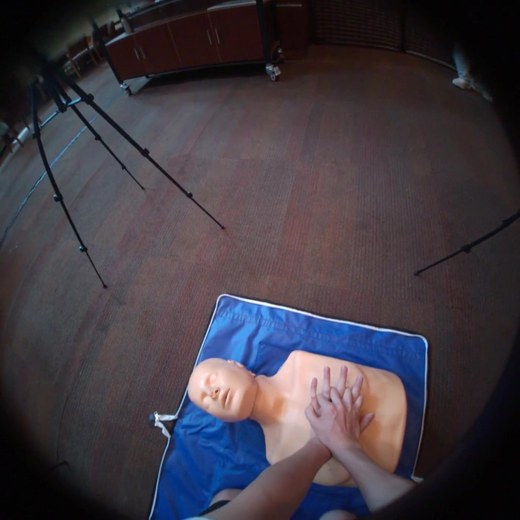} & \demoimg{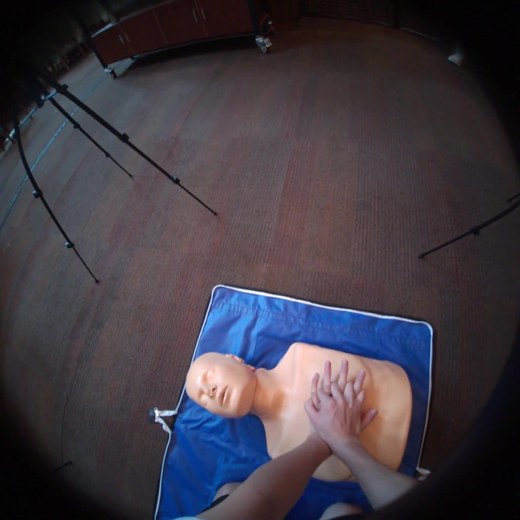} & \demoimg{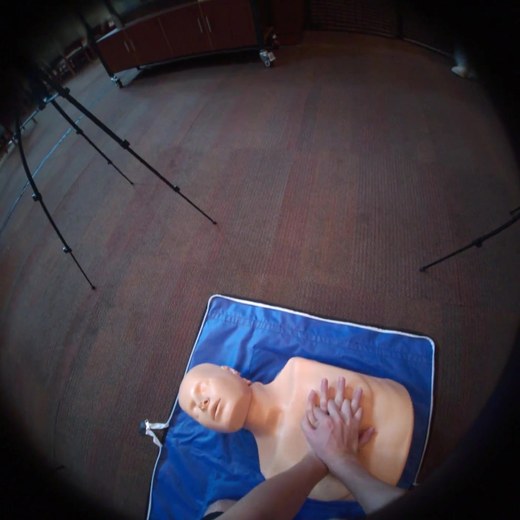} & \demoimg{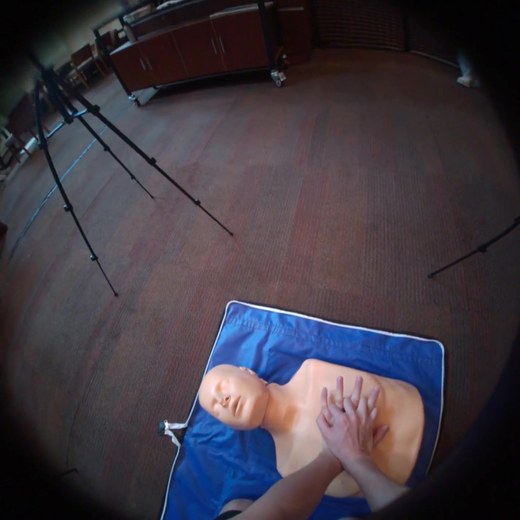}
\end{tabular}}\par}
\demotimeline{21}{0.0,3.0,6.0,9.0,12.0,15.0,18.0,21.0}
\demoq{How many times does the person compress the manikin's chest with their interlocked hands?}
{\centering
\begin{tabular}{@{}l@{\hspace{30pt}}l@{}}
\demopt{A}{23}{} & \demopt{B}{21}{} \\[3pt]
\demopt{C}{15}{} & \demopt{D}{18}{motiondynamicscolor}
\end{tabular}\par}
\end{democard}
\vspace{-2pt}\captionof{figure}{\small\textbf{Repetition Counting.} Ego-Exo4D clip, 21\,s.}
\label{fig:ex-repetition-counting-2}
\end{excard}

\begin{excard}
\begin{democard}{motiondynamicscolor}{Repetition Counting}{nymeria_mark.png}{49s}
{\centering\setlength{\tabcolsep}{0pt}\renewcommand{\arraystretch}{1}%
\demostrip{%
\begin{tabular}{@{}c@{\hspace{\demogap}}c@{\hspace{\demogap}}c@{\hspace{\demogap}}c@{}}
\demoimg{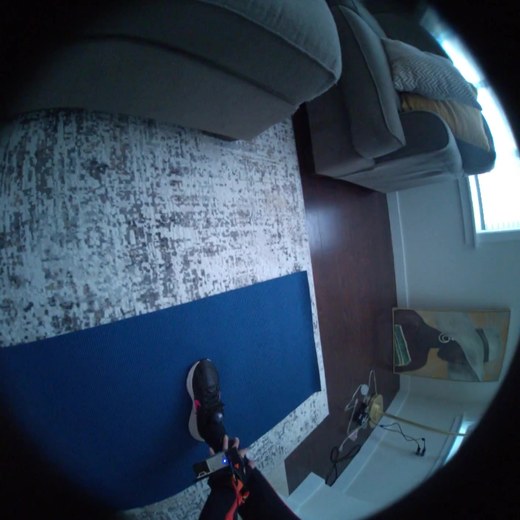} & \demoimg{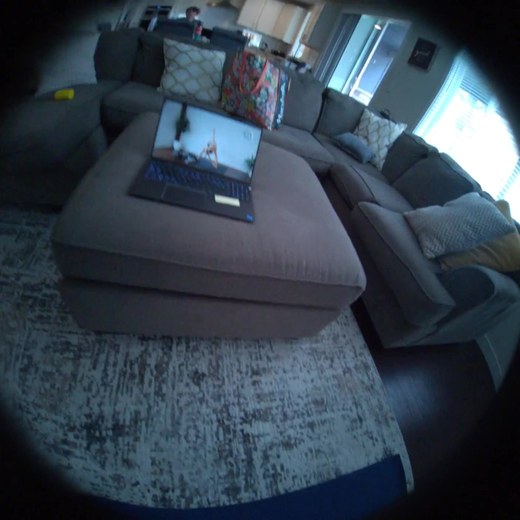} & \demoimg{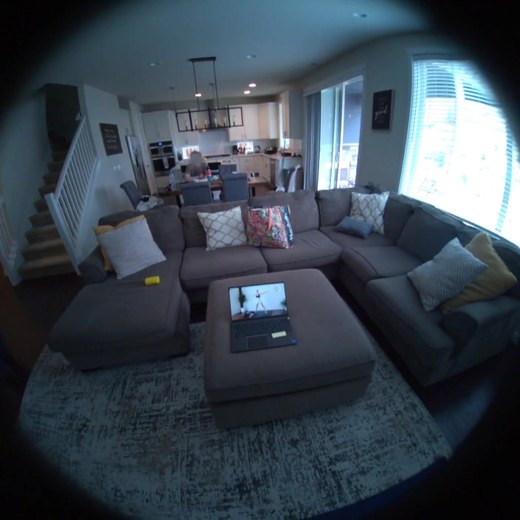} & \demoimg{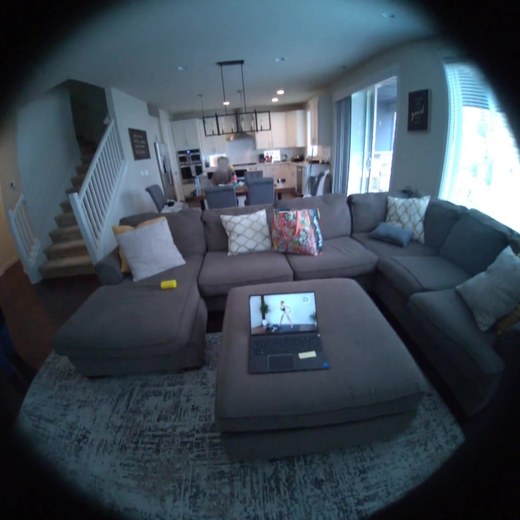}
\end{tabular}}\par\vspace{2pt}
\demostrip{%
\begin{tabular}{@{}c@{\hspace{\demogap}}c@{\hspace{\demogap}}c@{\hspace{\demogap}}c@{}}
\demoimg{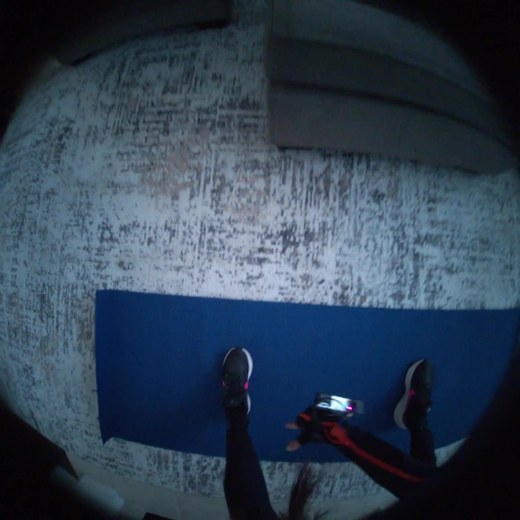} & \demoimg{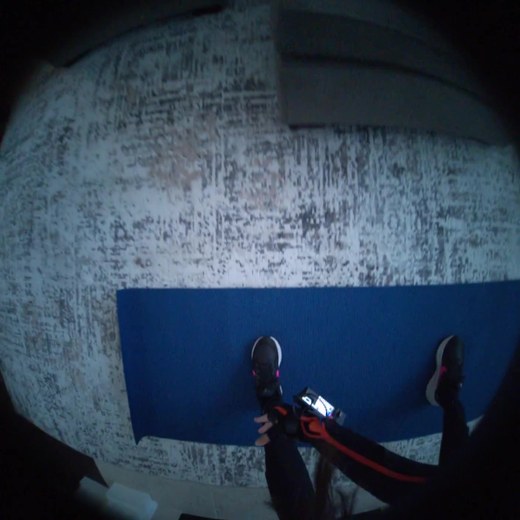} & \demoimg{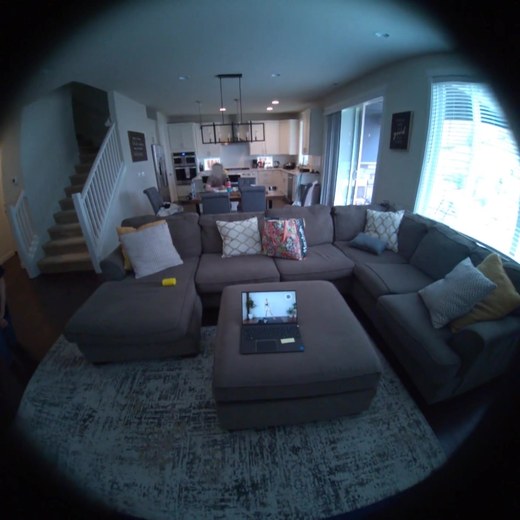} & \demoimg{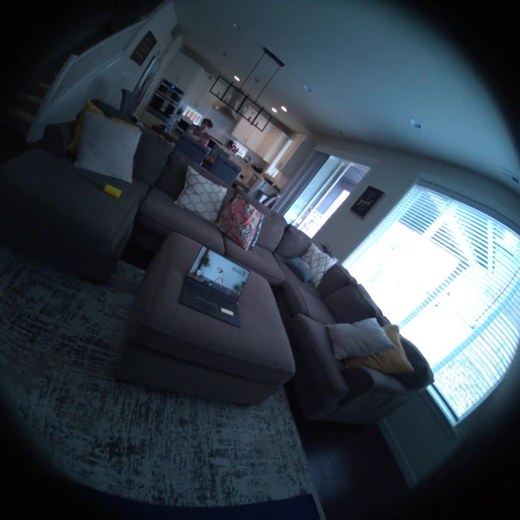}
\end{tabular}}\par}
\demotimeline{49}{0.0,7.0,14.0,21.0,28.0,35.0,42.0,49.0}
\demoq{How many times does the person mend over to touch their left leg?}
{\centering
\begin{tabular}{@{}l@{\hspace{30pt}}l@{}}
\demopt{A}{7}{} & \demopt{B}{4}{} \\[3pt]
\demopt{C}{5}{motiondynamicscolor} & \demopt{D}{6}{}
\end{tabular}\par}
\end{democard}
\vspace{-2pt}\captionof{figure}{\small\textbf{Repetition Counting.} Nymeria clip, 49\,s.}
\label{fig:ex-repetition-counting-3}
\end{excard}

\begin{excard}
\begin{democard}{motiondynamicscolor}{Movement Direction}{nymeria_mark.png}{35s}
{\centering\setlength{\tabcolsep}{0pt}\renewcommand{\arraystretch}{1}%
\demostrip{%
\begin{tabular}{@{}c@{\hspace{\demogap}}c@{\hspace{\demogap}}c@{\hspace{\demogap}}c@{}}
\demoimg{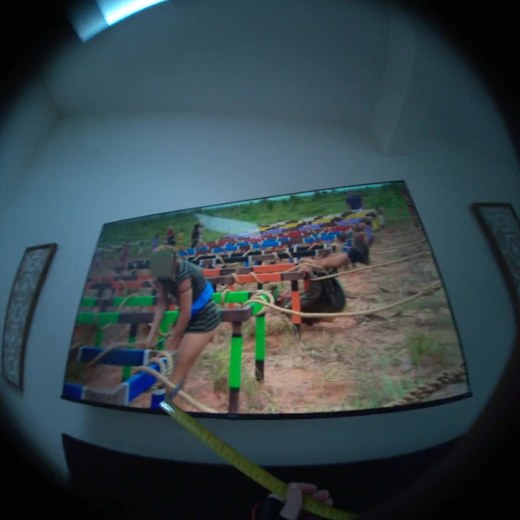} & \demoimg{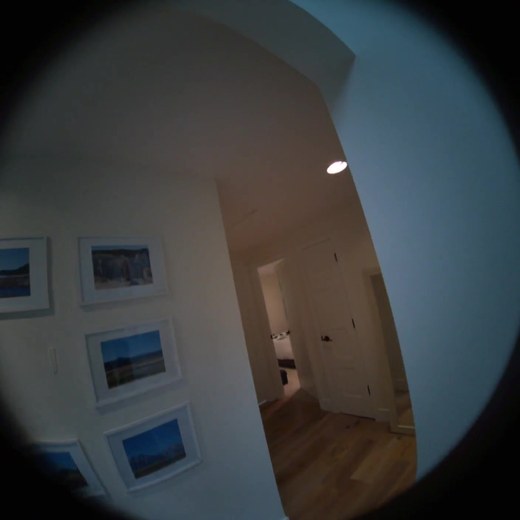} & \demoimg{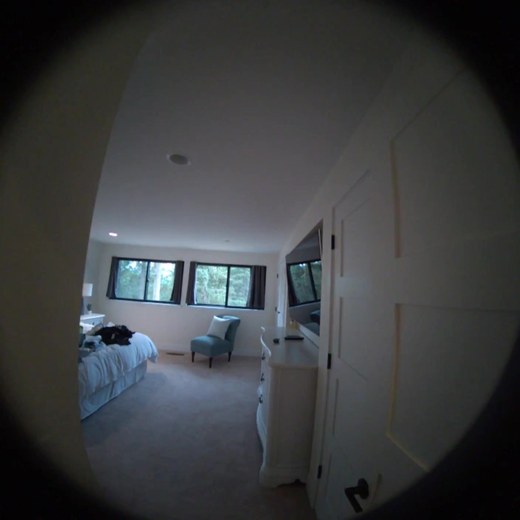} & \demoimg{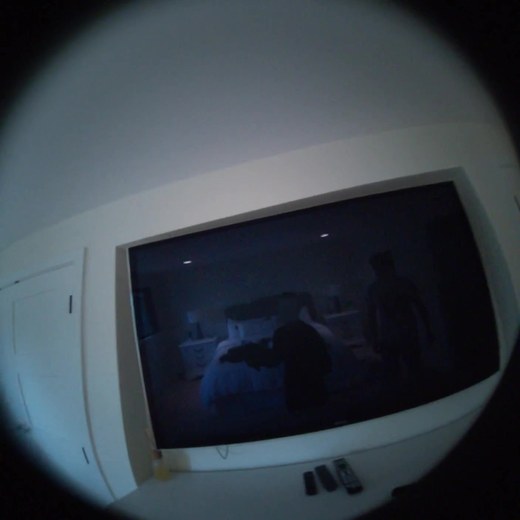}
\end{tabular}}\par\vspace{2pt}
\demostrip{%
\begin{tabular}{@{}c@{\hspace{\demogap}}c@{\hspace{\demogap}}c@{\hspace{\demogap}}c@{}}
\demoimg{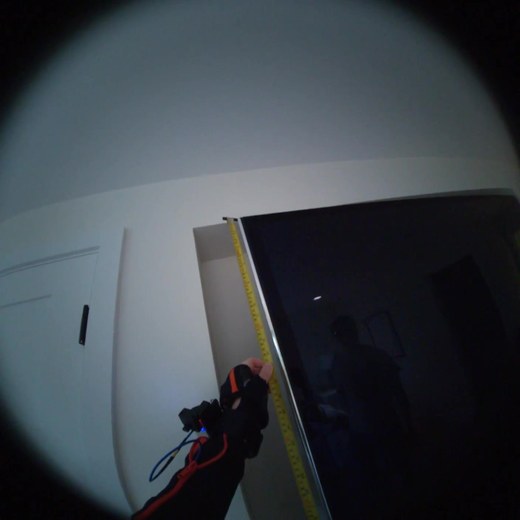} & \demoimg{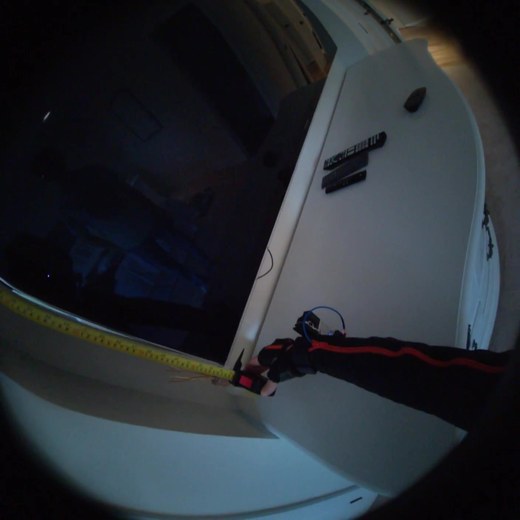} & \demoimg{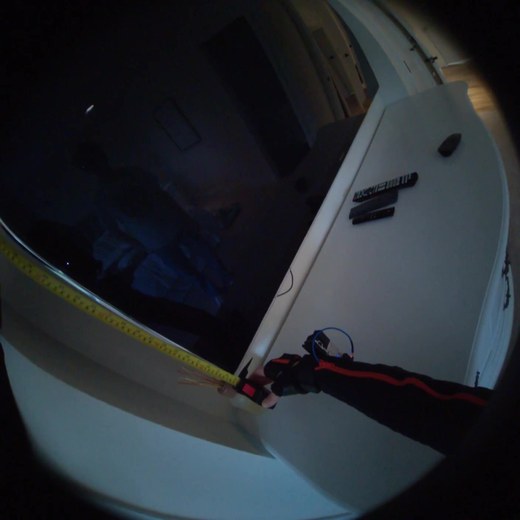} & \demoimg{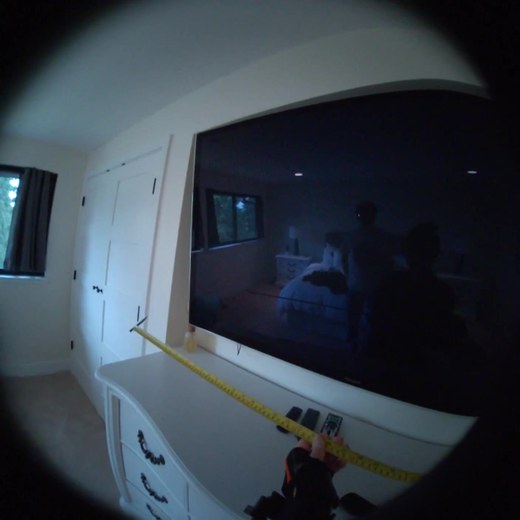}
\end{tabular}}\par}
\demotimeline{35}{0.0,5.0,9.8,14.8,19.8,24.8,29.5,34.5}
\demoq{After the person measures the TV in the bedroom, what is the sequence of their next two movements?}
{\centering\small
\begin{tabularx}{0.95\textwidth}{@{}X@{}}
\demopt{A}{They step forward, then step to their right}{} \\[2pt]
\demopt{B}{They take a backward step, then step to their right}{motiondynamicscolor} \\[2pt]
\demopt{C}{They step forward, then step to their left}{} \\[2pt]
\demopt{D}{They take a backward step, then step to their left}{}
\end{tabularx}\par}
\end{democard}
\vspace{-2pt}\captionof{figure}{\small\textbf{Movement Direction.} Nymeria clip, 35\,s.}
\label{fig:ex-movement-direction-1}
\end{excard}

\begin{excard}
\begin{democard}{motiondynamicscolor}{Movement Direction}{nymeria_mark.png}{42s}
{\centering\setlength{\tabcolsep}{0pt}\renewcommand{\arraystretch}{1}%
\demostrip{%
\begin{tabular}{@{}c@{\hspace{\demogap}}c@{\hspace{\demogap}}c@{\hspace{\demogap}}c@{}}
\demoimg{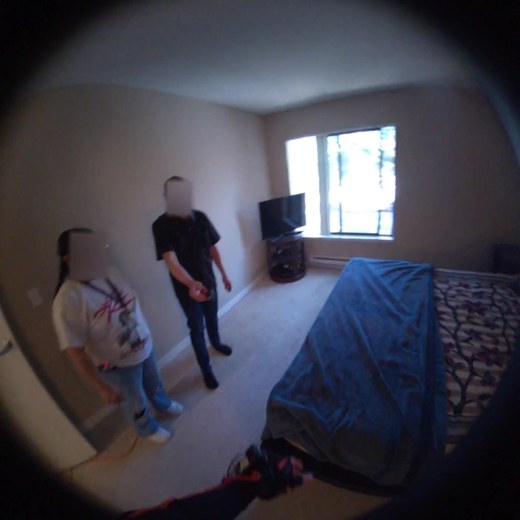} & \demoimg{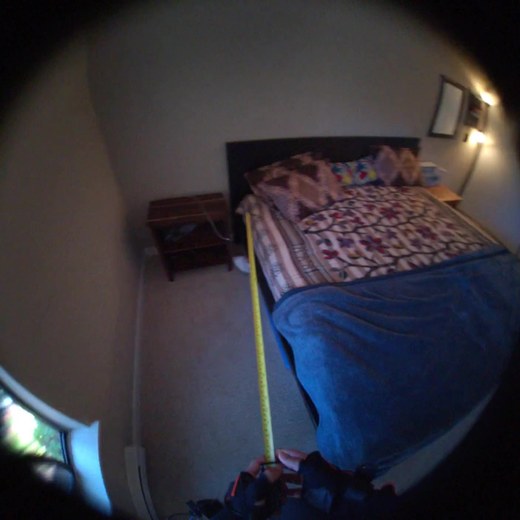} & \demoimg{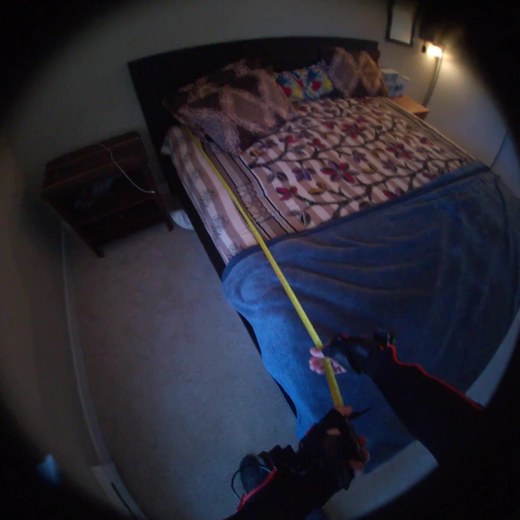} & \demoimg{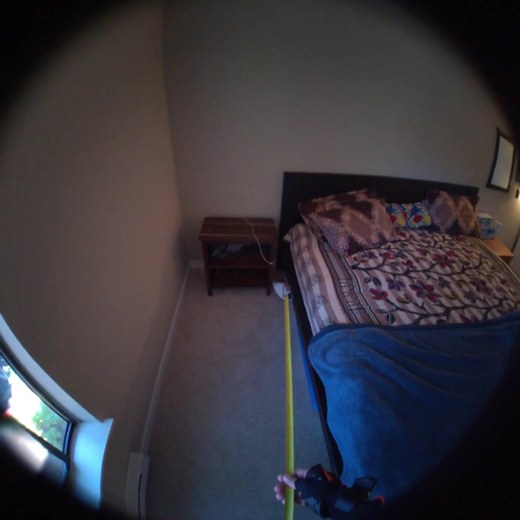}
\end{tabular}}\par\vspace{2pt}
\demostrip{%
\begin{tabular}{@{}c@{\hspace{\demogap}}c@{\hspace{\demogap}}c@{\hspace{\demogap}}c@{}}
\demoimg{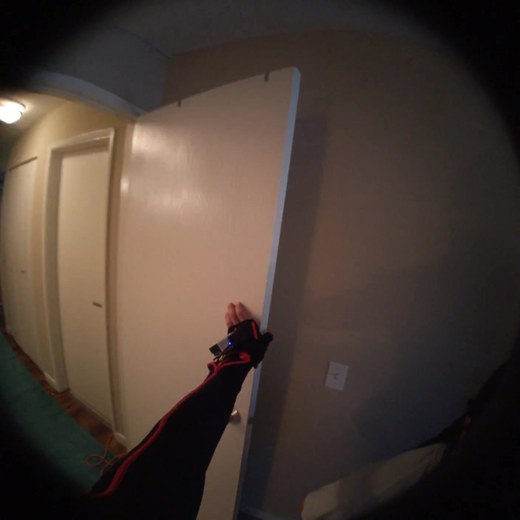} & \demoimg{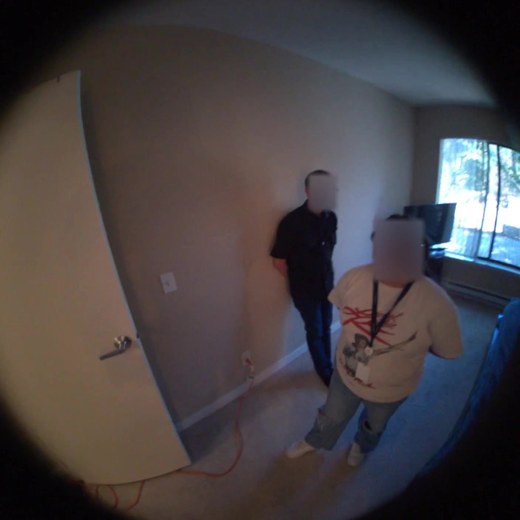} & \demoimg{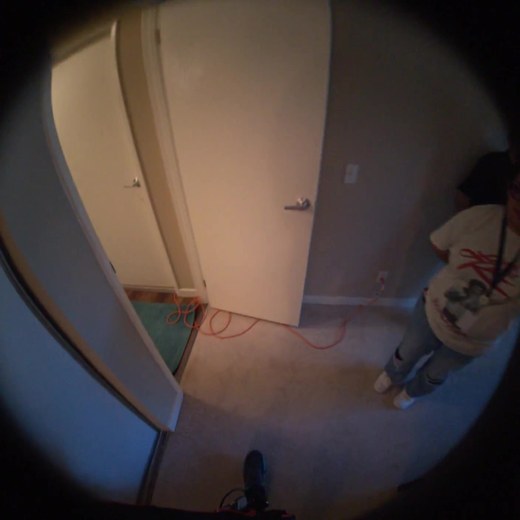} & \demoimg{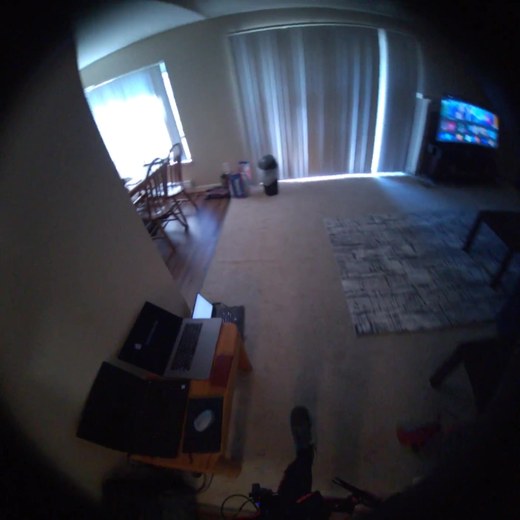}
\end{tabular}}\par}
\demotimeline{42}{0.0,6.0,12.0,18.0,23.8,29.8,35.8,41.8}
\demoq{After the person gets the measuring tape from another person, what is the sequence of their next two movements?}
{\centering\small
\begin{tabular}{@{}l@{\hspace{8pt}}l@{}}
\demopt{A}{They step to the right, then walk forward}{} & \demopt{B}{They step to the left, then walk backward}{} \\[3pt]
\demopt{C}{They step to the left, then walk forward}{motiondynamicscolor} & \demopt{D}{They step to the right, then walk backward}{}
\end{tabular}\par}
\end{democard}
\vspace{-2pt}\captionof{figure}{\small\textbf{Movement Direction.} Nymeria clip, 42\,s.}
\label{fig:ex-movement-direction-2}
\end{excard}

\begin{excard}
\begin{democard}{motiondynamicscolor}{Movement Direction}{nymeria_mark.png}{20s}
{\centering\setlength{\tabcolsep}{0pt}\renewcommand{\arraystretch}{1}%
\demostrip{%
\begin{tabular}{@{}c@{\hspace{\demogap}}c@{\hspace{\demogap}}c@{\hspace{\demogap}}c@{}}
\demoimg{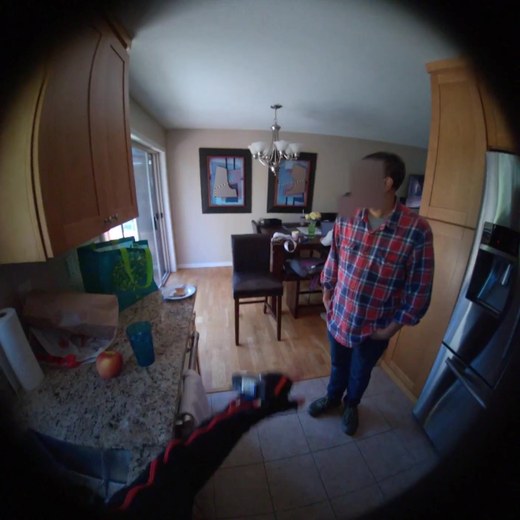} & \demoimg{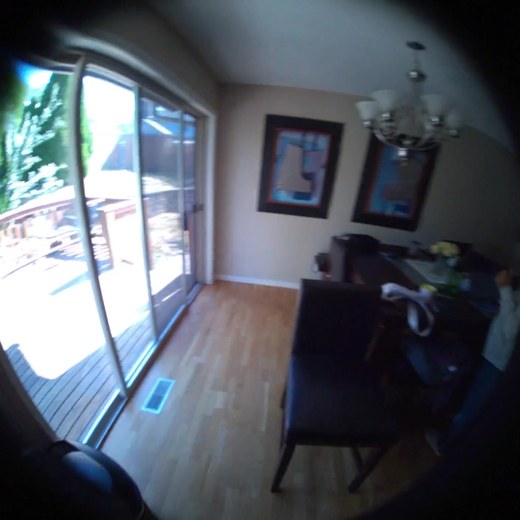} & \demoimg{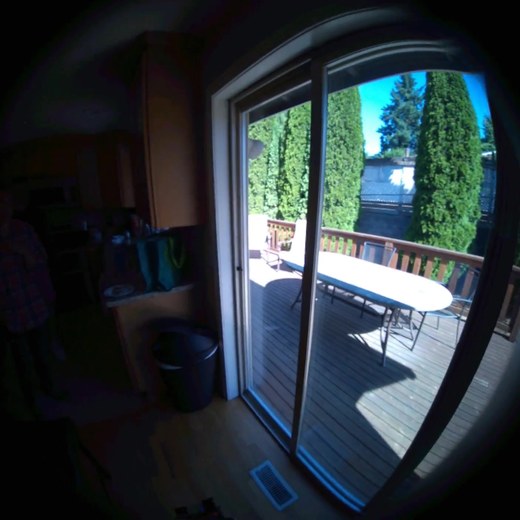} & \demoimg{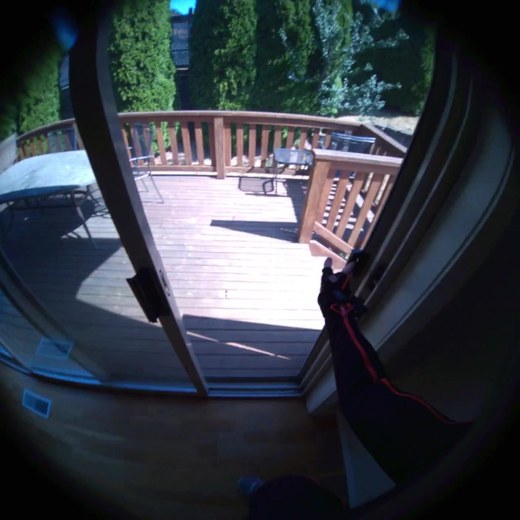}
\end{tabular}}\par\vspace{2pt}
\demostrip{%
\begin{tabular}{@{}c@{\hspace{\demogap}}c@{\hspace{\demogap}}c@{\hspace{\demogap}}c@{}}
\demoimg{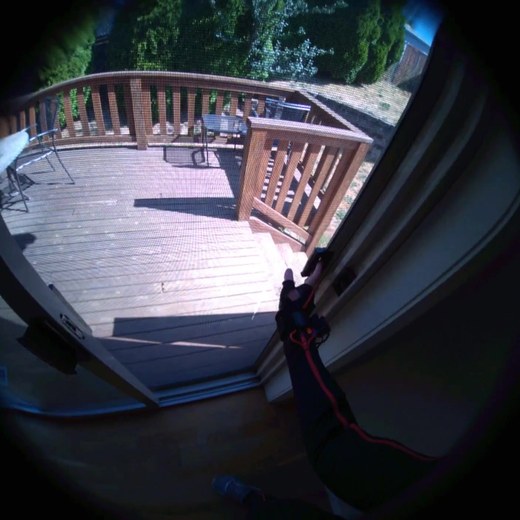} & \demoimg{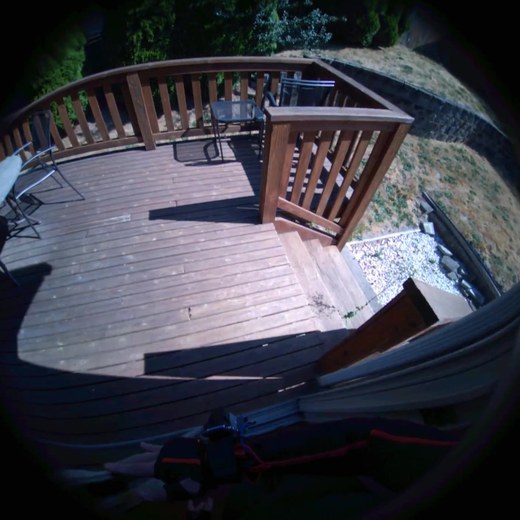} & \demoimg{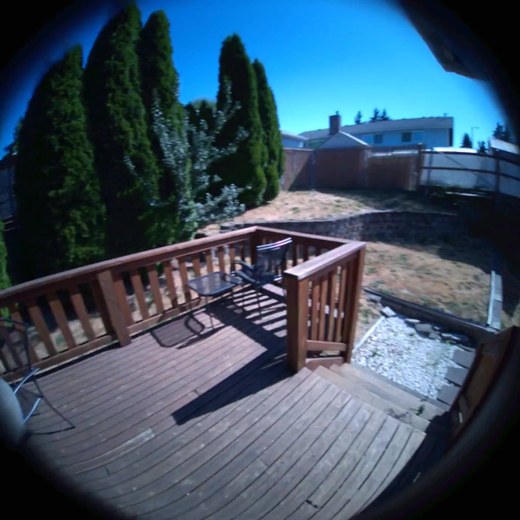} & \demoimg{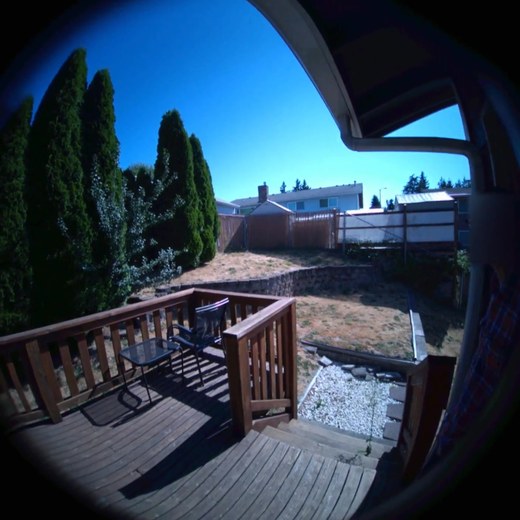}
\end{tabular}}\par}
\demotimeline{20}{0.0,2.8,5.8,8.5,11.2,14.0,17.0,19.8}
\demoq{After the person walks at the dining area, what is the sequence of their next two turns?}
{\centering\small
\begin{tabularx}{0.95\textwidth}{@{}X@{}}
\demopt{A}{They turn clockwise, then turn clockwise}{} \\[2pt]
\demopt{B}{They turn clockwise, then turn counterclockwise}{} \\[2pt]
\demopt{C}{They turn counterclockwise, then turn clockwise}{motiondynamicscolor} \\[2pt]
\demopt{D}{They turn counterclockwise, then turn counterclockwise}{}
\end{tabularx}\par}
\end{democard}
\vspace{-2pt}\captionof{figure}{\small\textbf{Movement Direction.} Nymeria clip, 20\,s.}
\label{fig:ex-movement-direction-3}
\end{excard}

\begin{excard}
\begin{democard}{motiondynamicscolor}{Opposite Motion Discrimination}{nymeria_mark.png}{14s}
{\centering\setlength{\tabcolsep}{0pt}\renewcommand{\arraystretch}{1}%
\demostrip{%
\begin{tabular}{@{}c@{\hspace{\demogap}}c@{\hspace{\demogap}}c@{\hspace{\demogap}}c@{}}
\demoimg{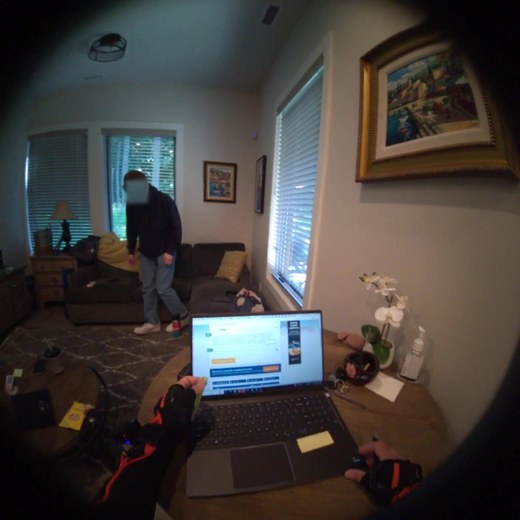} & \demoimg{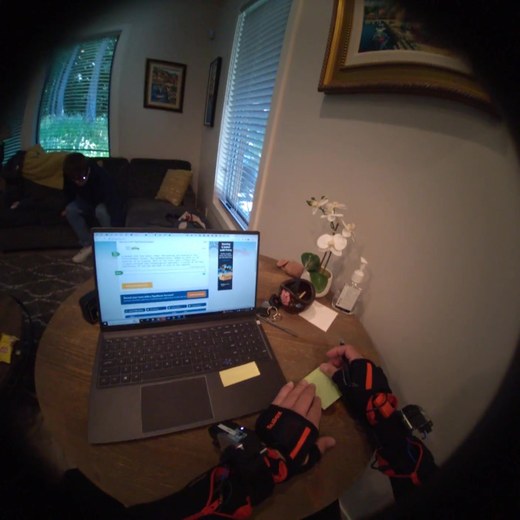} & \demoimg{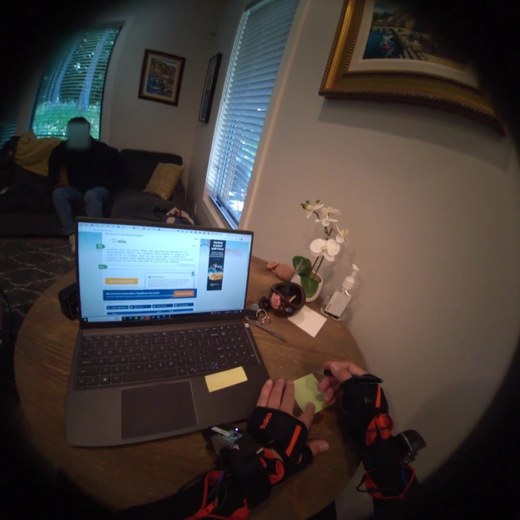} & \demoimg{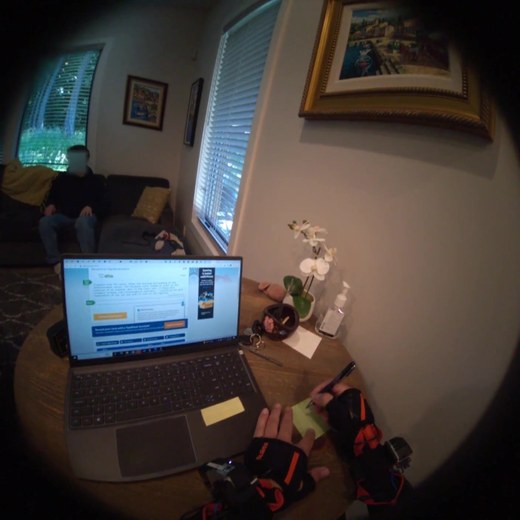}
\end{tabular}}\par\vspace{2pt}
\demostrip{%
\begin{tabular}{@{}c@{\hspace{\demogap}}c@{\hspace{\demogap}}c@{\hspace{\demogap}}c@{}}
\demoimg{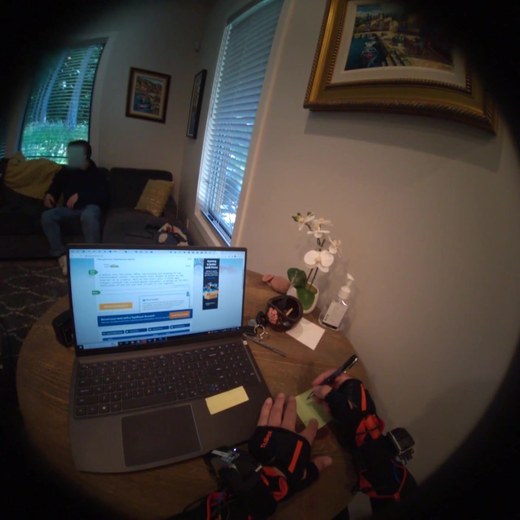} & \demoimg{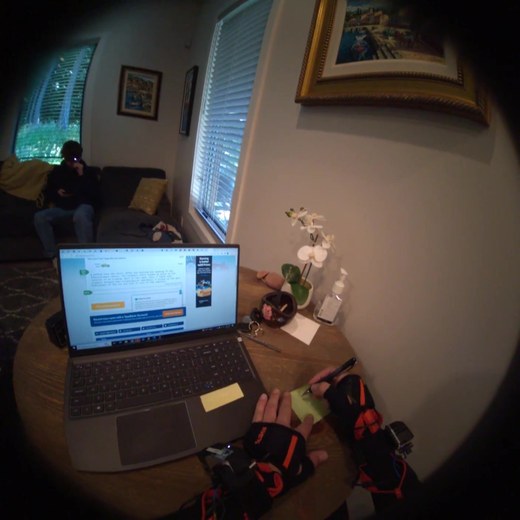} & \demoimg{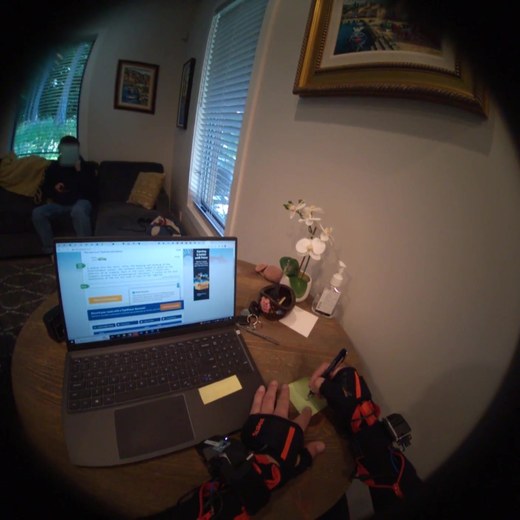} & \demoimg{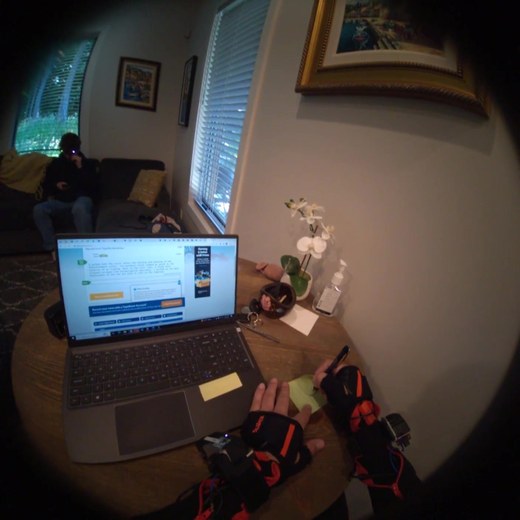}
\end{tabular}}\par}
\demotimeline{14}{0.0,2.0,4.0,6.0,7.8,9.8,11.8,13.8}
\demoq{What is the person doing with the pen?}
{\centering
\begin{tabular}{@{}l@{\hspace{30pt}}l@{}}
\demopt{A}{Drawing lines from left to right}{} & \demopt{B}{Drawing lines from top to bottom}{motiondynamicscolor} \\[3pt]
\demopt{C}{Drawing lines from right to left}{} & \demopt{D}{Drawing lines from bottom to top}{}
\end{tabular}\par}
\end{democard}
\vspace{-2pt}\captionof{figure}{\small\textbf{Opposite Motion Discrimination.} Nymeria clip, 14\,s.}
\label{fig:ex-opposite-motion-discrimination-1}
\end{excard}

\begin{excard}
\begin{democard}{motiondynamicscolor}{Opposite Motion Discrimination}{nymeria_mark.png}{10s}
{\centering\setlength{\tabcolsep}{0pt}\renewcommand{\arraystretch}{1}%
\demostrip{%
\begin{tabular}{@{}c@{\hspace{\demogap}}c@{\hspace{\demogap}}c@{\hspace{\demogap}}c@{}}
\demoimg{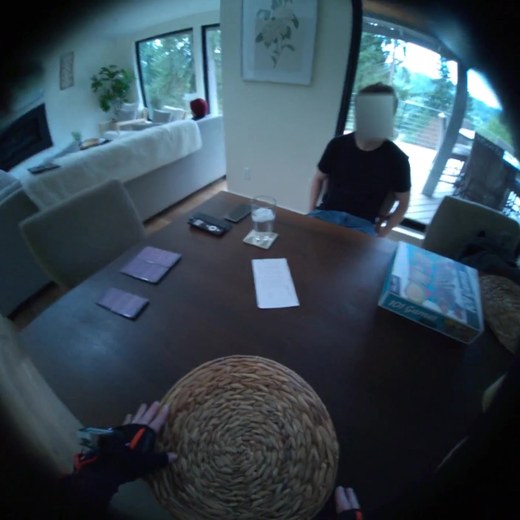} & \demoimg{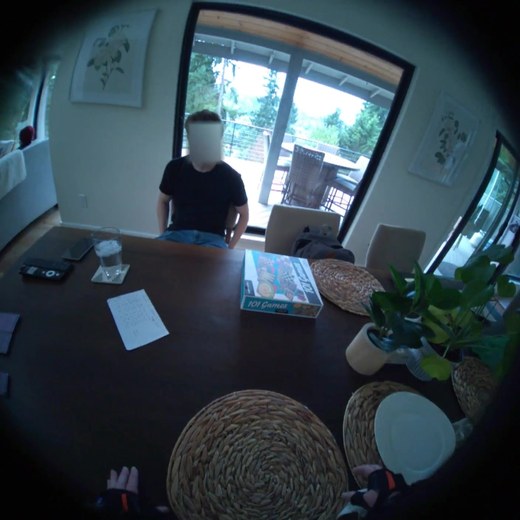} & \demoimg{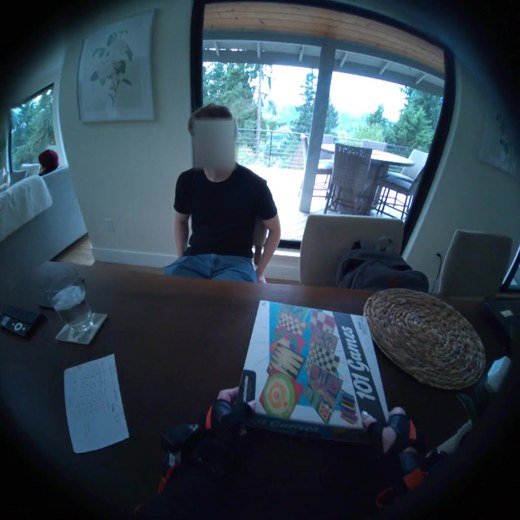} & \demoimg{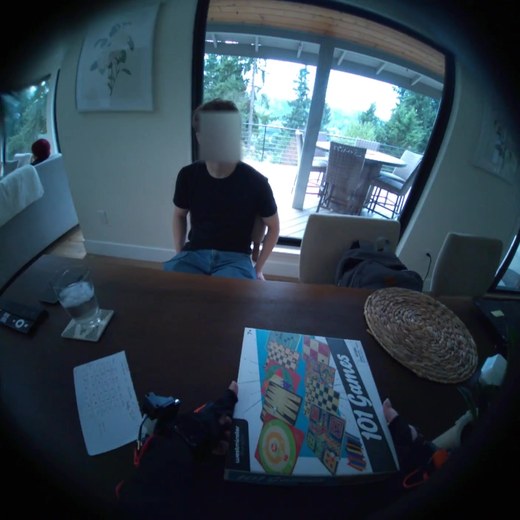}
\end{tabular}}\par\vspace{2pt}
\demostrip{%
\begin{tabular}{@{}c@{\hspace{\demogap}}c@{\hspace{\demogap}}c@{\hspace{\demogap}}c@{}}
\demoimg{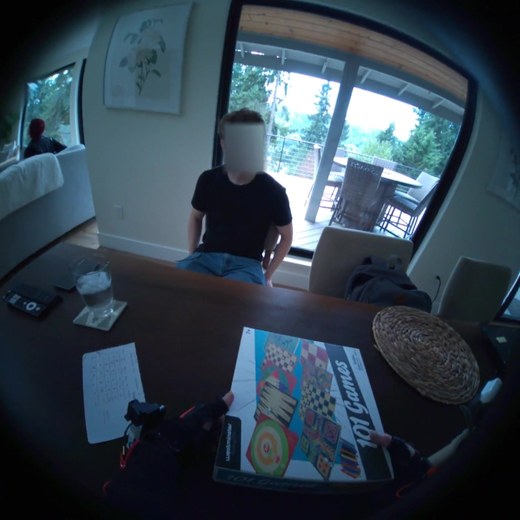} & \demoimg{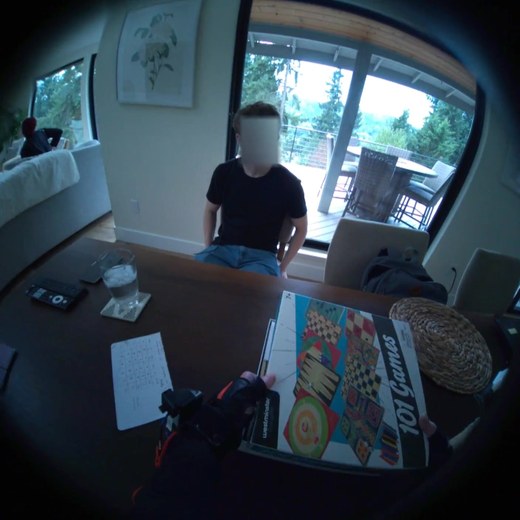} & \demoimg{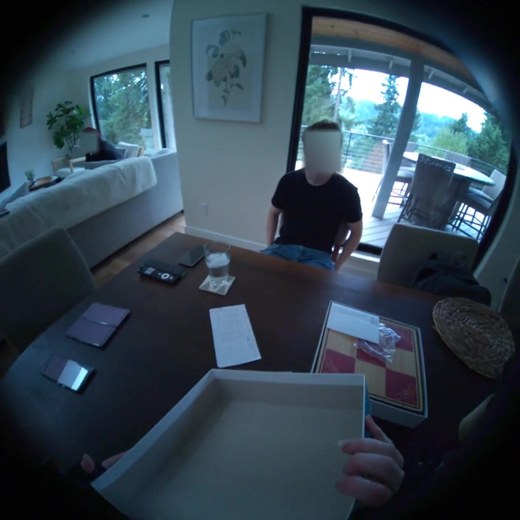} & \demoimg{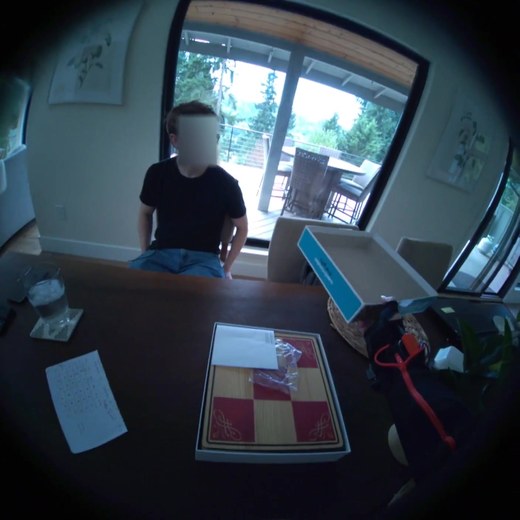}
\end{tabular}}\par}
\demotimeline{10}{0.0,1.5,2.8,4.2,5.5,7.0,8.2,9.8}
\demoq{What is the person doing with the placemat?}
{\centering
\begin{tabular}{@{}l@{\hspace{30pt}}l@{}}
\demopt{A}{Sliding the placemat to the right}{motiondynamicscolor} & \demopt{B}{Sliding the placemat down}{} \\[3pt]
\demopt{C}{Sliding the placemat to the left}{} & \demopt{D}{Sliding the placemat up}{}
\end{tabular}\par}
\end{democard}
\vspace{-2pt}\captionof{figure}{\small\textbf{Opposite Motion Discrimination.} Nymeria clip, 10\,s.}
\label{fig:ex-opposite-motion-discrimination-2}
\end{excard}

\begin{excard}
\begin{democard}{motiondynamicscolor}{Opposite Motion Discrimination}{nymeria_mark.png}{11s}
{\centering\setlength{\tabcolsep}{0pt}\renewcommand{\arraystretch}{1}%
\demostrip{%
\begin{tabular}{@{}c@{\hspace{\demogap}}c@{\hspace{\demogap}}c@{\hspace{\demogap}}c@{}}
\demoimg{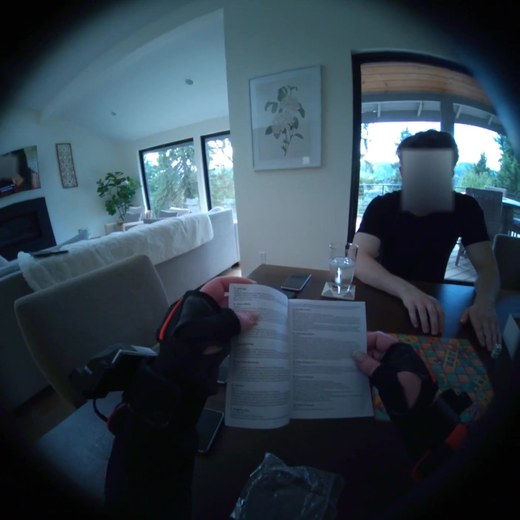} & \demoimg{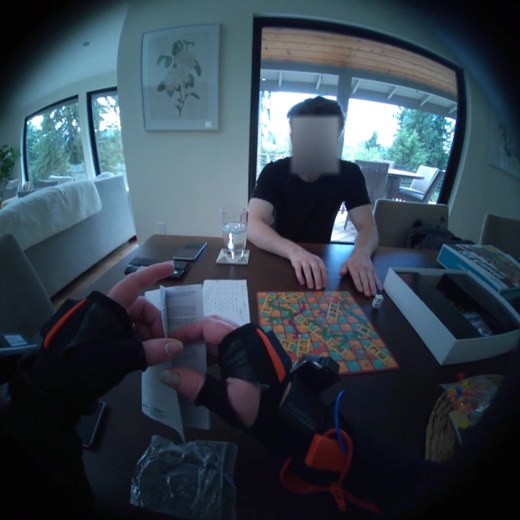} & \demoimg{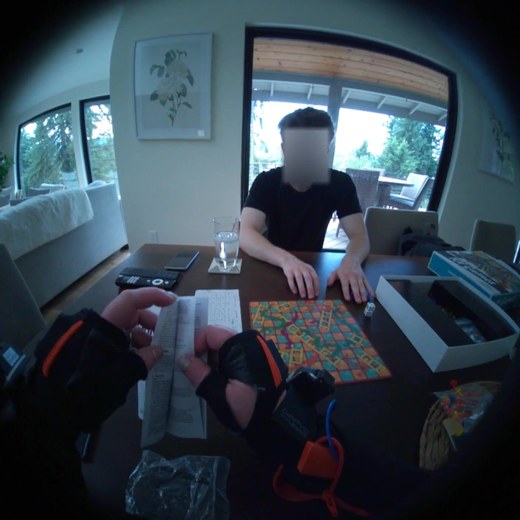} & \demoimg{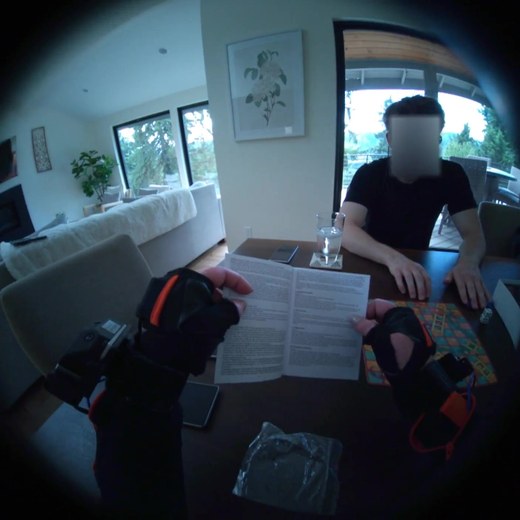}
\end{tabular}}\par\vspace{2pt}
\demostrip{%
\begin{tabular}{@{}c@{\hspace{\demogap}}c@{\hspace{\demogap}}c@{\hspace{\demogap}}c@{}}
\demoimg{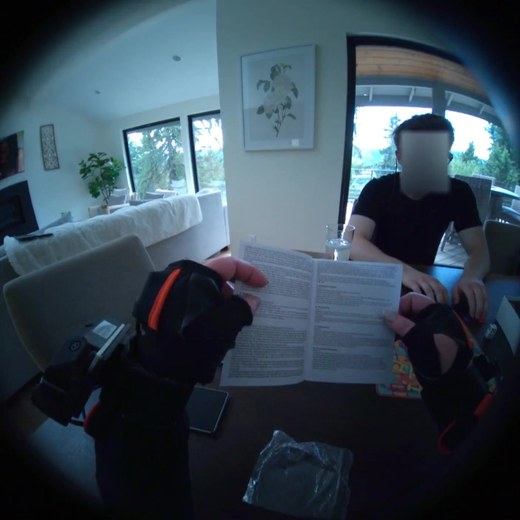} & \demoimg{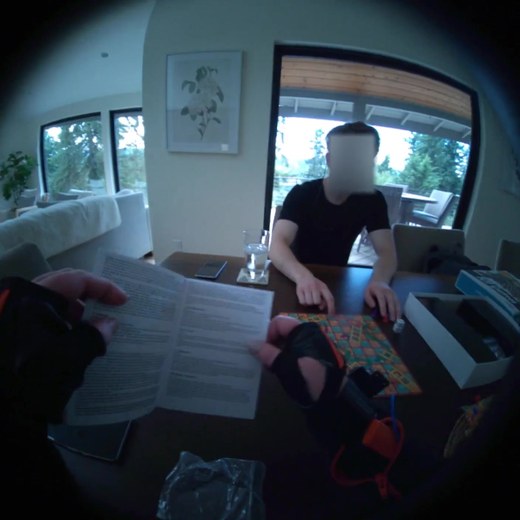} & \demoimg{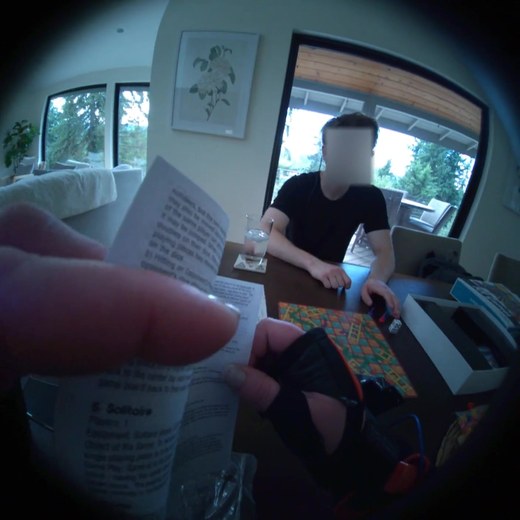} & \demoimg{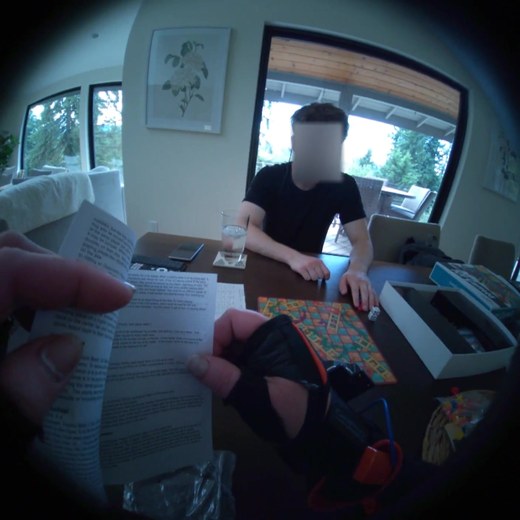}
\end{tabular}}\par}
\demotimeline{11}{0.0,1.5,3.2,4.8,6.5,8.0,9.8}
\demoq{What is the person doing with the board game manual?}
{\centering
\begin{tabular}{@{}l@{\hspace{30pt}}l@{}}
\demopt{A}{Closing the board game manual}{} & \demopt{B}{Flipping back one page}{motiondynamicscolor} \\[3pt]
\demopt{C}{Opening the board game manual}{} & \demopt{D}{Flipping ahead one page}{}
\end{tabular}\par}
\end{democard}
\vspace{-2pt}\captionof{figure}{\small\textbf{Opposite Motion Discrimination.} Nymeria clip, 11\,s.}
\label{fig:ex-opposite-motion-discrimination-3}
\end{excard}

\begin{excard}
\begin{democard}{temporalcolor}{Adjacent Action Identification}{nymeria_mark.png}{45s}
{\centering\setlength{\tabcolsep}{0pt}\renewcommand{\arraystretch}{1}%
\demostrip{%
\begin{tabular}{@{}c@{\hspace{\demogap}}c@{\hspace{\demogap}}c@{\hspace{\demogap}}c@{}}
\demoimg{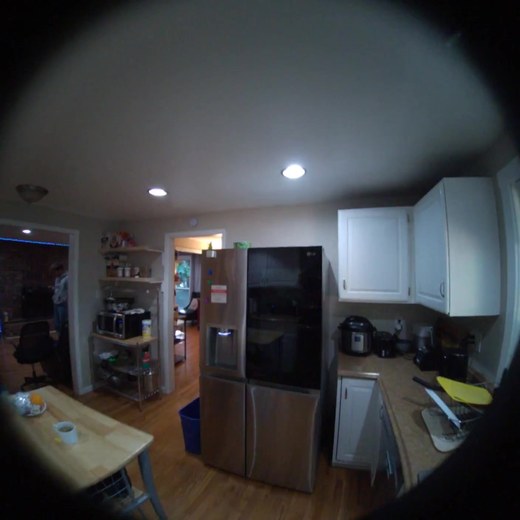} & \demoimg{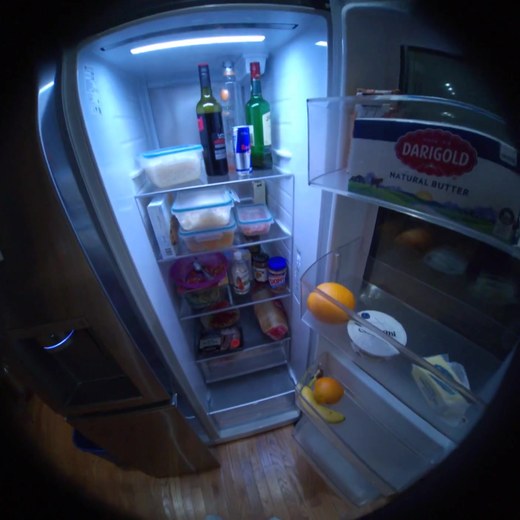} & \demoimg{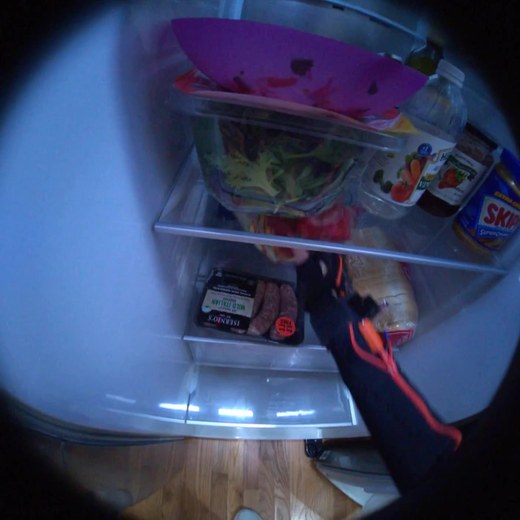} & \demoimg{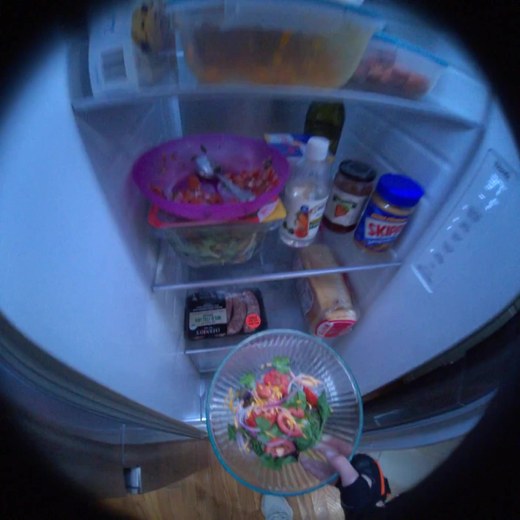}
\end{tabular}}\par\vspace{2pt}
\demostrip{%
\begin{tabular}{@{}c@{\hspace{\demogap}}c@{\hspace{\demogap}}c@{\hspace{\demogap}}c@{}}
\demoimg{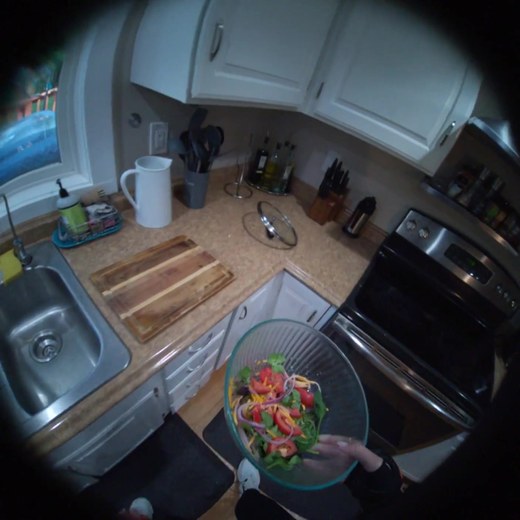} & \demoimg{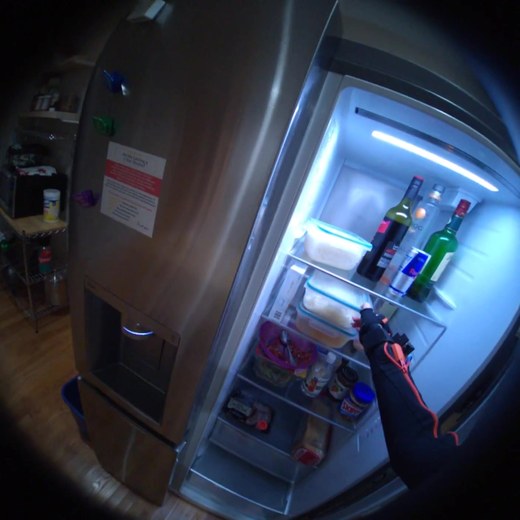} & \demoimg{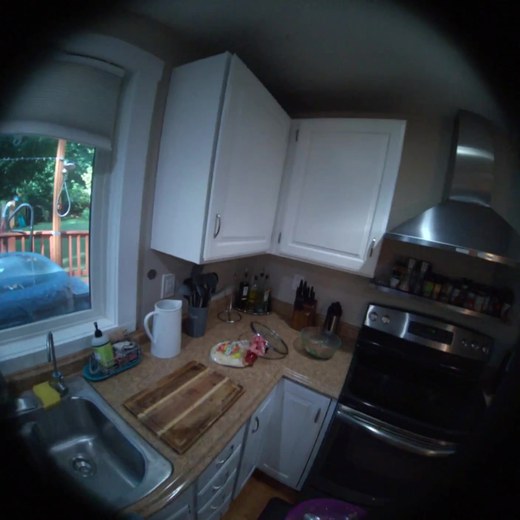} & \demoimg{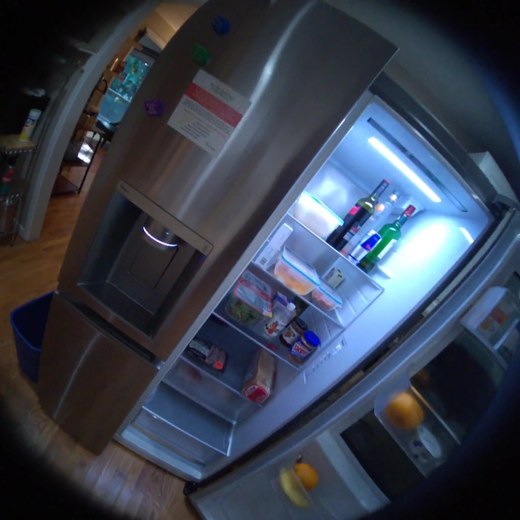}
\end{tabular}}\par}
\demotimeline{45}{0.0,6.5,12.8,19.2,25.5,32.0,38.2,44.8}
\demoq{What does the person do immediately before pulling out a bowl of vegetables?}
{\centering\small
\begin{tabular}{@{}l@{\hspace{10pt}}l@{}}
\demopt{A}{Pushes the pan lid forward}{} & \demopt{B}{Grabs then releases a bag of sliced bread}{temporalcolor} \\[3pt]
\demopt{C}{Lifts and puts down a pack of sausages}{} & \demopt{D}{Pulls out a tupperware}{}
\end{tabular}\par}
\end{democard}
\vspace{-2pt}\captionof{figure}{\small\textbf{Adjacent Action Identification.} Nymeria clip, 45\,s.}
\label{fig:ex-adjacent-action-identification-1}
\end{excard}

\begin{excard}
\begin{democard}{temporalcolor}{Adjacent Action Identification}{ee4d_mark.png}{49s}
{\centering\setlength{\tabcolsep}{0pt}\renewcommand{\arraystretch}{1}%
\demostrip{%
\begin{tabular}{@{}c@{\hspace{\demogap}}c@{\hspace{\demogap}}c@{\hspace{\demogap}}c@{}}
\demoimg{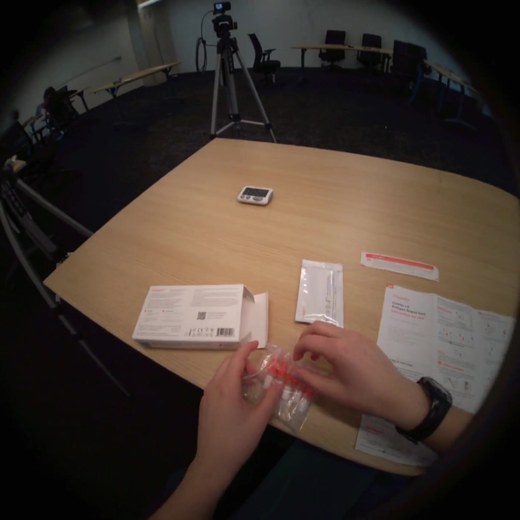} & \demoimg{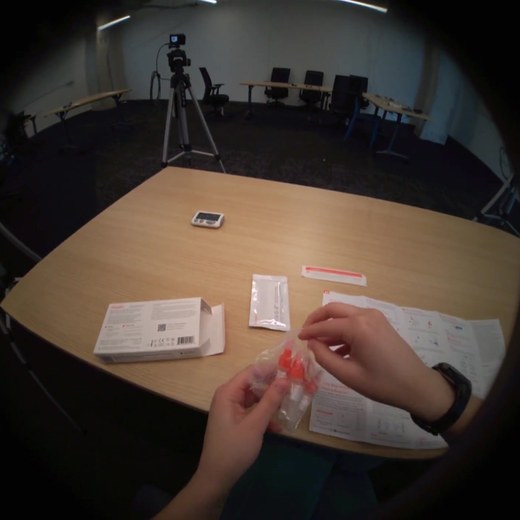} & \demoimg{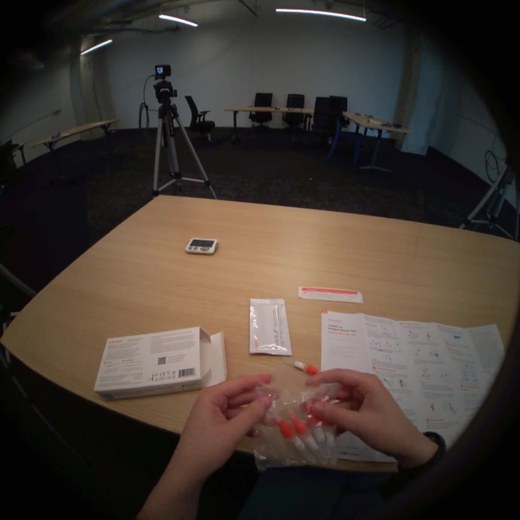} & \demoimg{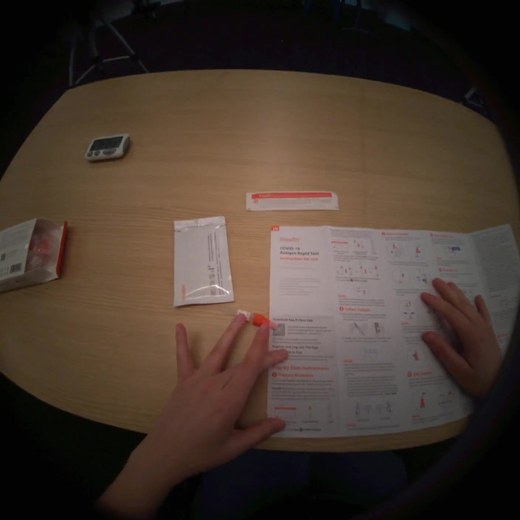}
\end{tabular}}\par\vspace{2pt}
\demostrip{%
\begin{tabular}{@{}c@{\hspace{\demogap}}c@{\hspace{\demogap}}c@{\hspace{\demogap}}c@{}}
\demoimg{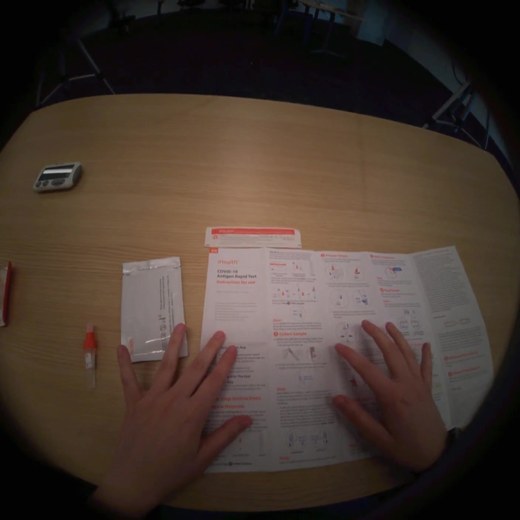} & \demoimg{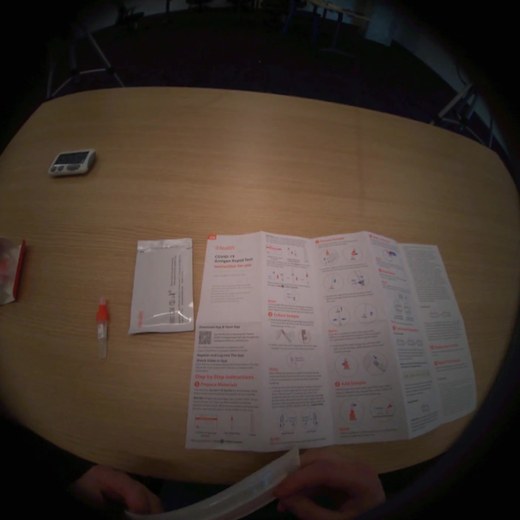} & \demoimg{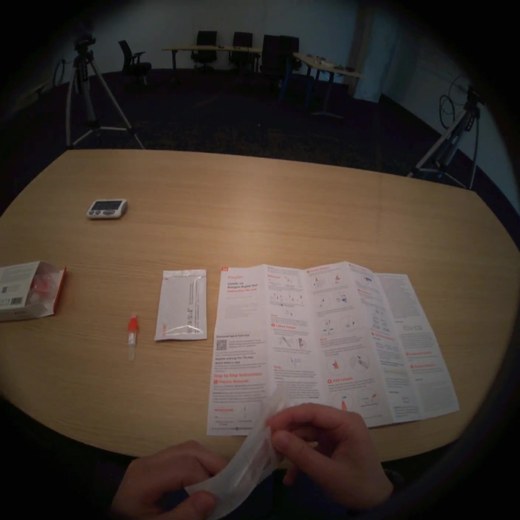} & \demoimg{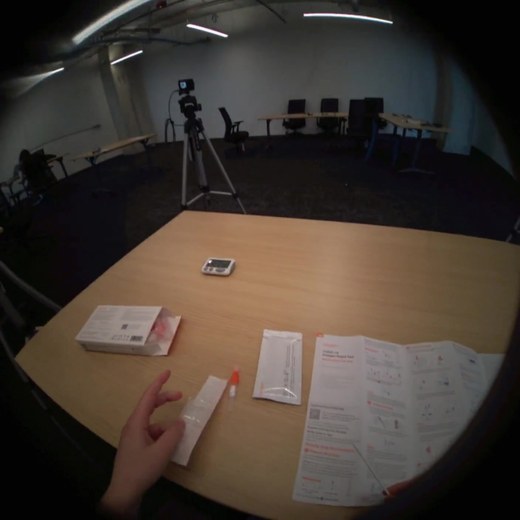}
\end{tabular}}\par}
\demotimeline{49}{0.0,7.0,14.0,21.0,28.2,35.2,42.2}
\demoq{What does the person do immediately after dropping the test box on the table?}
{\centering\small
\begin{tabularx}{0.95\textwidth}{@{}X@{}}
\demopt{A}{Picks the extraction tube on the table}{} \\[2pt]
\demopt{B}{Places both hands on the test manual}{temporalcolor} \\[2pt]
\demopt{C}{Removes the nasal swab from the sachet}{} \\[2pt]
\demopt{D}{Returns the extraction tube sachet into the test box}{}
\end{tabularx}\par}
\end{democard}
\vspace{-2pt}\captionof{figure}{\small\textbf{Adjacent Action Identification.} Ego-Exo4D clip, 49\,s.}
\label{fig:ex-adjacent-action-identification-2}
\end{excard}

\begin{excard}
\begin{democard}{temporalcolor}{Chronological Ordering}{nymeria_mark.png}{12s}
{\centering\setlength{\tabcolsep}{0pt}\renewcommand{\arraystretch}{1}%
\demostrip{%
\begin{tabular}{@{}c@{\hspace{\demogap}}c@{\hspace{\demogap}}c@{\hspace{\demogap}}c@{}}
\demoimg{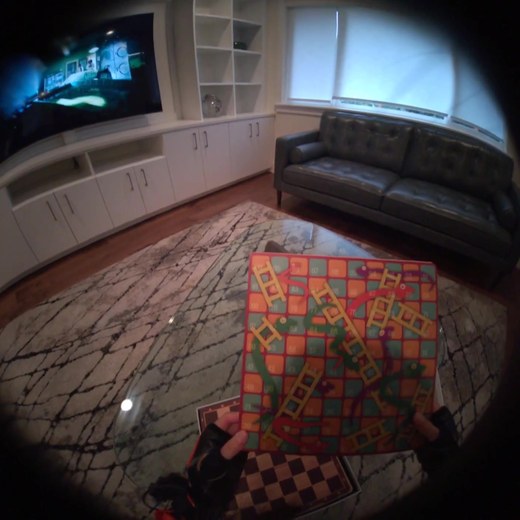} & \demoimg{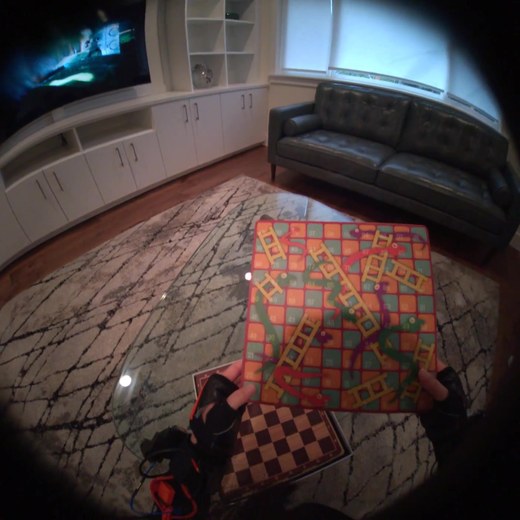} & \demoimg{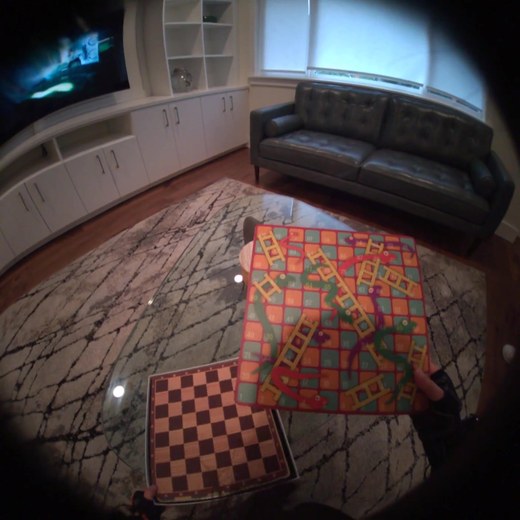} & \demoimg{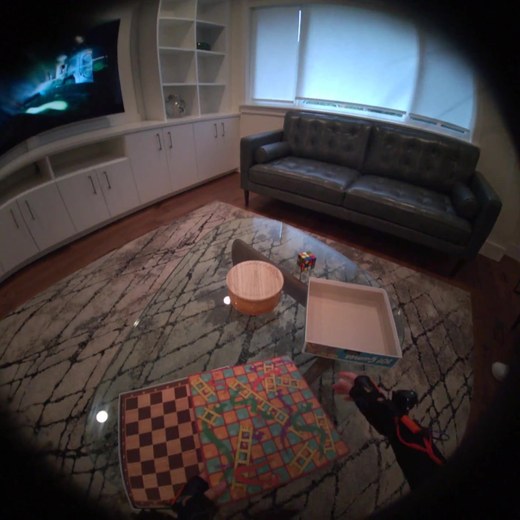}
\end{tabular}}\par\vspace{2pt}
\demostrip{%
\begin{tabular}{@{}c@{\hspace{\demogap}}c@{\hspace{\demogap}}c@{\hspace{\demogap}}c@{}}
\demoimg{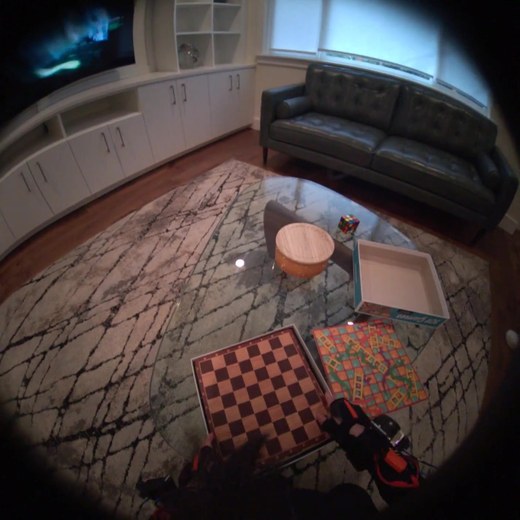} & \demoimg{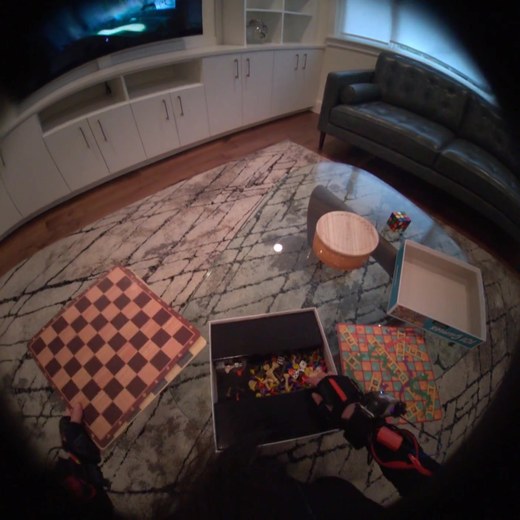} & \demoimg{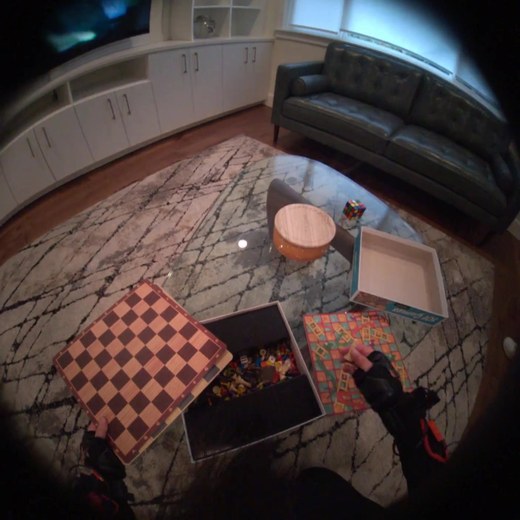} & \demoimg{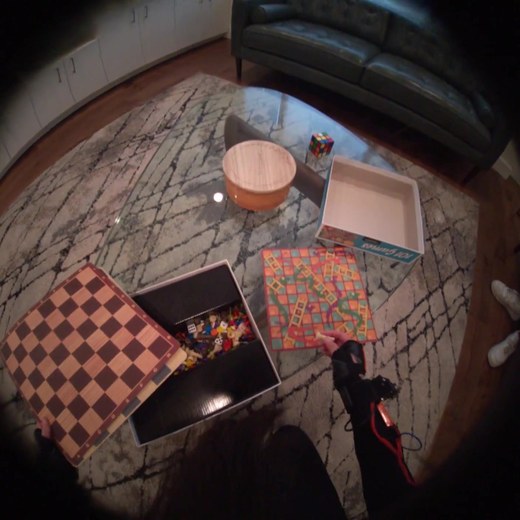}
\end{tabular}}\par}
\demotimeline{12}{0.0,1.5,3.2,4.8,6.5,8.0,9.8,11.2}
\demoq{What is the correct temporal order of these three actions: [takes the chessboard from the box / puts the board game on the table / pushes the box cover]?}
{\centering\small
\begin{tabularx}{0.95\textwidth}{@{}X@{}}
\demopt{A}{Pushes the box cover $\rightarrow$ takes the chessboard from the box $\rightarrow$ puts the board game on the table}{} \\[2pt]
\demopt{B}{Takes the chessboard from the box $\rightarrow$ puts the board game on the table $\rightarrow$ pushes the box cover}{} \\[2pt]
\demopt{C}{Pushes the box cover $\rightarrow$ puts the board game on the table $\rightarrow$ takes the chessboard from the box}{temporalcolor} \\[2pt]
\demopt{D}{Puts the board game on the table $\rightarrow$ pushes the box cover $\rightarrow$ takes the chessboard from the box}{}
\end{tabularx}\par}
\end{democard}
\vspace{-2pt}\captionof{figure}{\small\textbf{Chronological Ordering.} Nymeria clip, 12\,s.}
\label{fig:ex-chronological-ordering-1}
\end{excard}

\begin{excard}
\begin{democard}{temporalcolor}{Missing Event Inference}{nymeria_mark.png}{11s}
{\centering\setlength{\tabcolsep}{0pt}\renewcommand{\arraystretch}{1}%
\demostrip{%
\begin{tabular}{@{}c@{\hspace{\demogap}}c@{\hspace{\demogap}}c@{\hspace{\demogap}}c@{}}
\demoimg{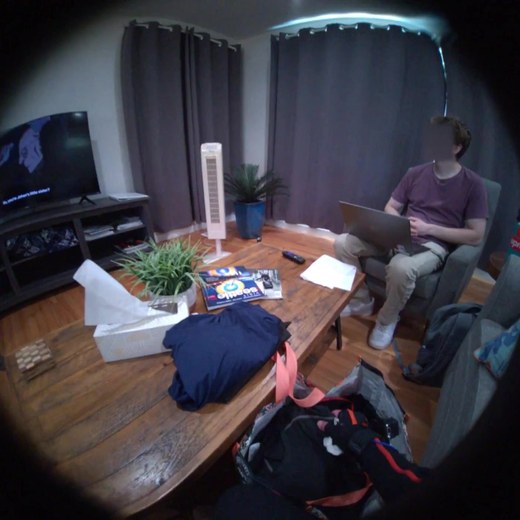} & \demoimg{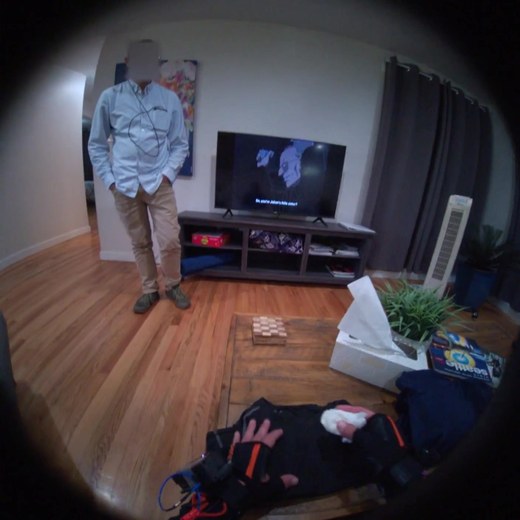} & \demoimg{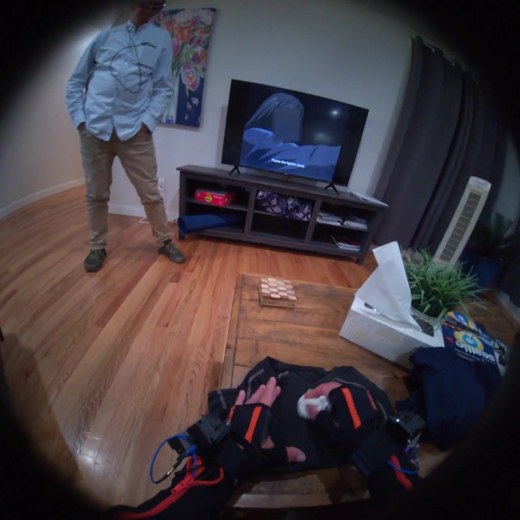} & \demoimg{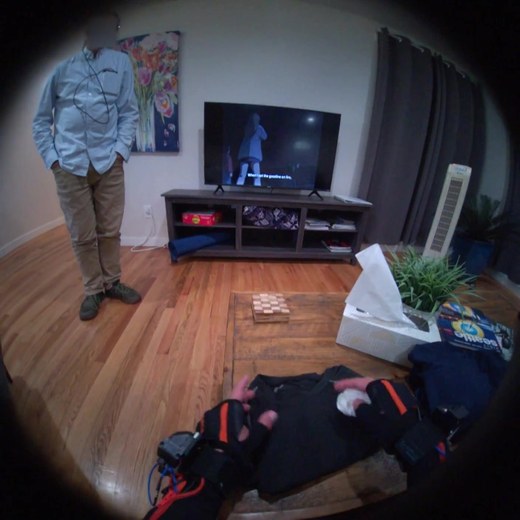}
\end{tabular}}\par\vspace{2pt}
\demostrip{%
\begin{tabular}{@{}c@{\hspace{\demogap}}c@{\hspace{\demogap}}c@{\hspace{\demogap}}c@{}}
\demoimg{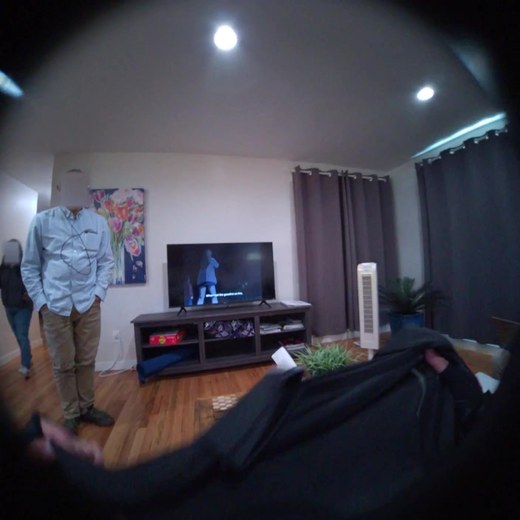} & \demoimg{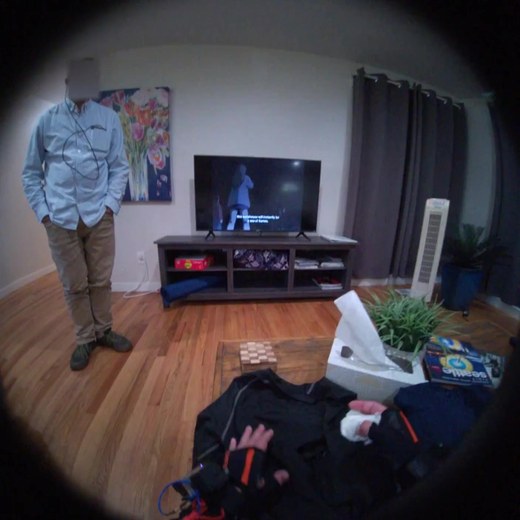} & \demoimg{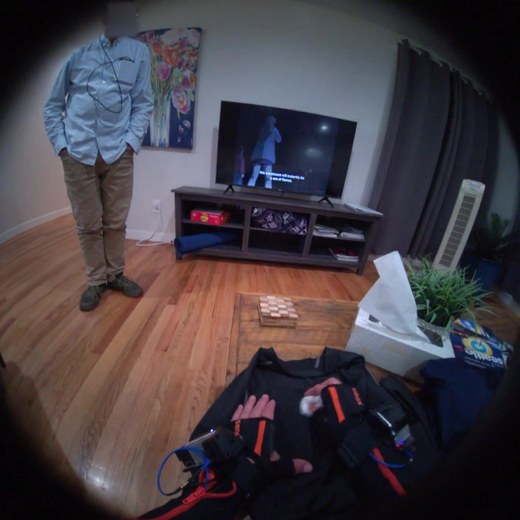} & \demoimg{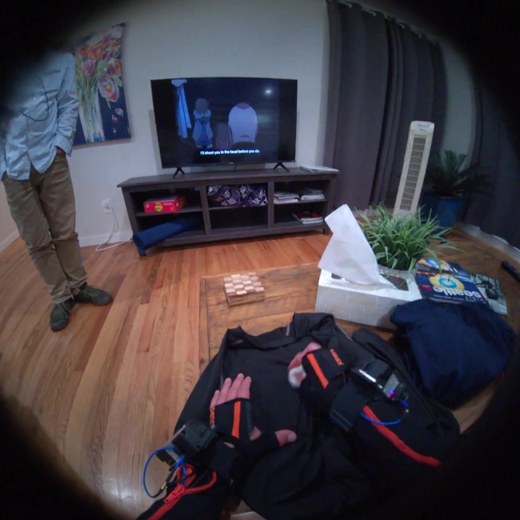}
\end{tabular}}\par}
\demotimeline{11}{0.0,1.5,3.2,4.8,6.5,8.0,9.8}
\demoq{What action is missing from this sequence: places the clothing on the table $\rightarrow$ [?] $\rightarrow$ flips the clothing on the table?}
{\centering\small
\begin{tabular}{@{}l@{\hspace{30pt}}l@{}}
\demopt{A}{Looks at their phone}{} & \demopt{B}{Scrubs the clothing with the tissue}{temporalcolor} \\[3pt]
\demopt{C}{Retrieves a piece of clothing from the bag}{} & \demopt{D}{Folds the clothing on the table}{}
\end{tabular}\par}
\end{democard}
\vspace{-2pt}\captionof{figure}{\small\textbf{Missing Event Inference.} Nymeria clip, 11\,s.}
\label{fig:ex-missing-event-inference-1}
\end{excard}

\begin{excard}
\begin{democard}{temporalcolor}{Next Action Prediction}{nymeria_mark.png}{36s}
{\centering\setlength{\tabcolsep}{0pt}\renewcommand{\arraystretch}{1}%
\demostrip{%
\begin{tabular}{@{}c@{\hspace{\demogap}}c@{\hspace{\demogap}}c@{\hspace{\demogap}}c@{}}
\demoimg{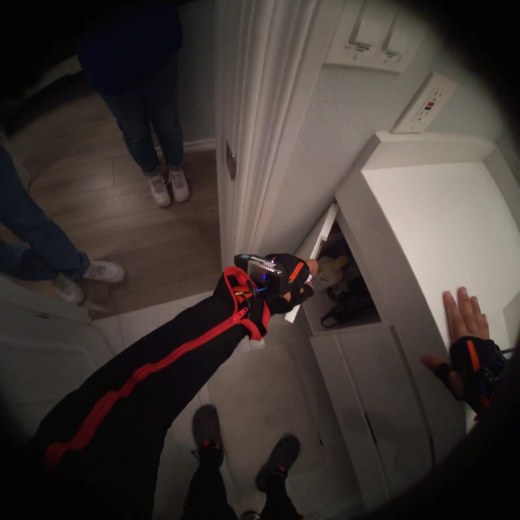} & \demoimg{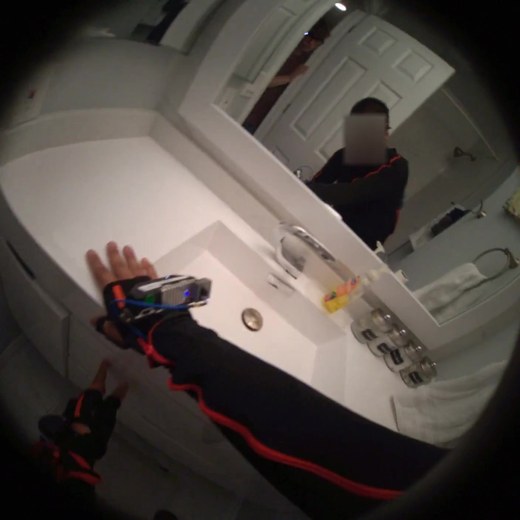} & \demoimg{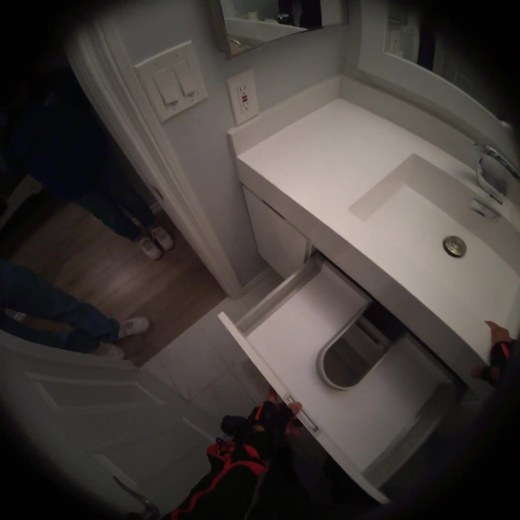} & \demoimg{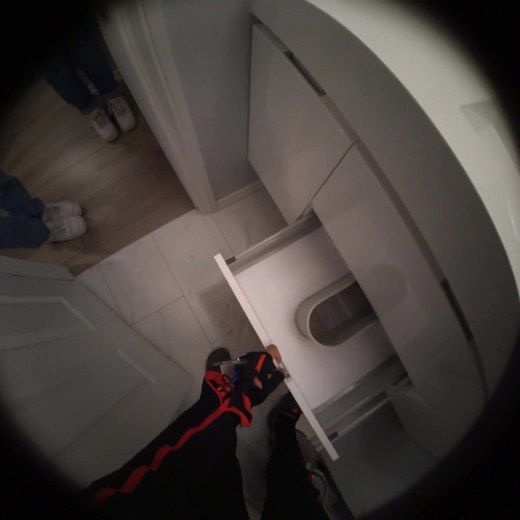}
\end{tabular}}\par\vspace{2pt}
\demostrip{%
\begin{tabular}{@{}c@{\hspace{\demogap}}c@{\hspace{\demogap}}c@{\hspace{\demogap}}c@{}}
\demoimg{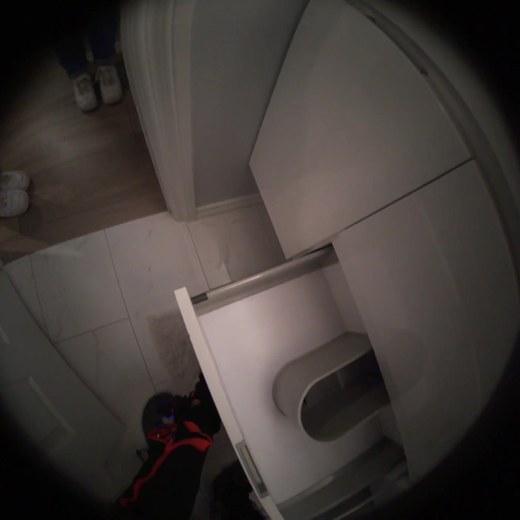} & \demoimg{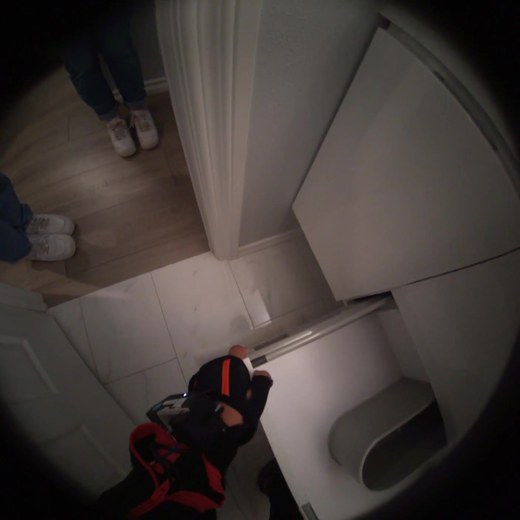} & \demoimg{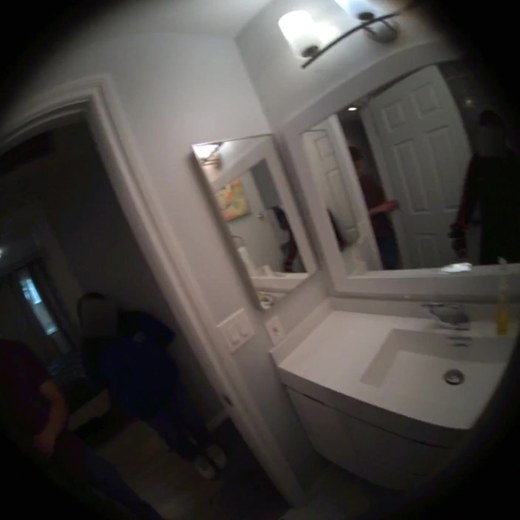} & \demoimg{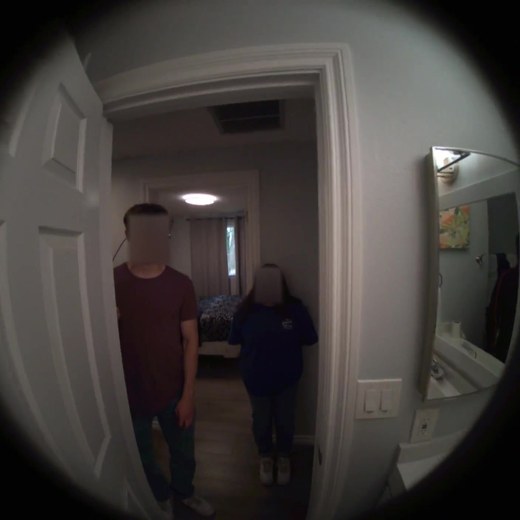}
\end{tabular}}\par}
\demotimeline{36}{0.0,5.0,10.2,15.2,20.5,25.5,30.8,35.8}
\demoq{Based on what the person has just done in the clip, what are they likely to do next?}
{\centering\small
\begin{tabularx}{0.95\textwidth}{@{}X@{}}
\demopt{A}{Take a bottle from the countertop}{} \\[2pt]
\demopt{B}{Adjust the faucet on the sink with their right hand}{} \\[2pt]
\demopt{C}{Pick up a towel from the rack}{} \\[2pt]
\demopt{D}{Turn off the light with their right hand}{temporalcolor}
\end{tabularx}\par}
\end{democard}
\vspace{-2pt}\captionof{figure}{\small\textbf{Next Action Prediction.} Nymeria clip, 36\,s.}
\label{fig:ex-next-action-prediction-1}
\end{excard}

\begin{excard}
\begin{democard}{temporalcolor}{Repeated Event Reasoning}{nymeria_mark.png}{57s}
{\centering\setlength{\tabcolsep}{0pt}\renewcommand{\arraystretch}{1}%
\demostrip{%
\begin{tabular}{@{}c@{\hspace{\demogap}}c@{\hspace{\demogap}}c@{\hspace{\demogap}}c@{}}
\demoimg{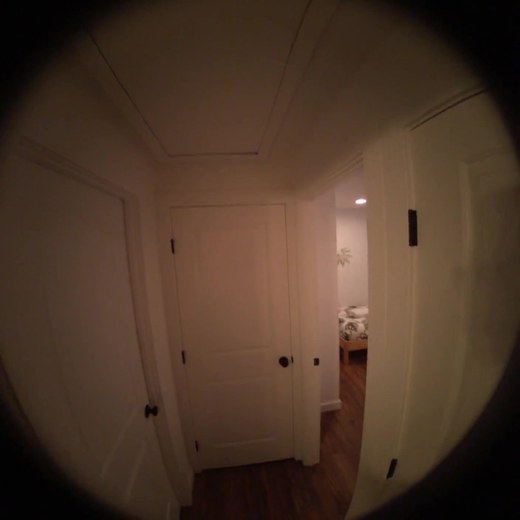} & \demoimg{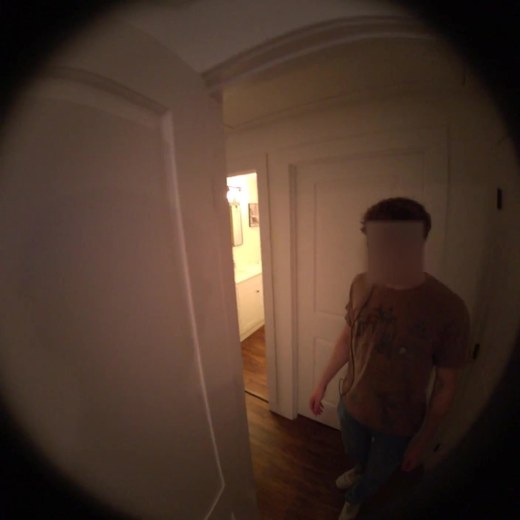} & \demoimg{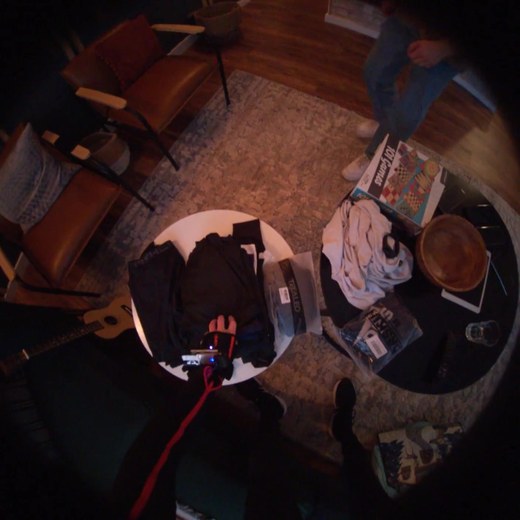} & \demoimg{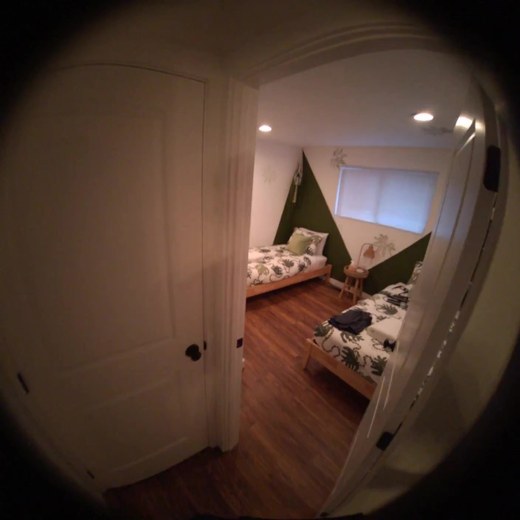}
\end{tabular}}\par\vspace{2pt}
\demostrip{%
\begin{tabular}{@{}c@{\hspace{\demogap}}c@{\hspace{\demogap}}c@{\hspace{\demogap}}c@{}}
\demoimg{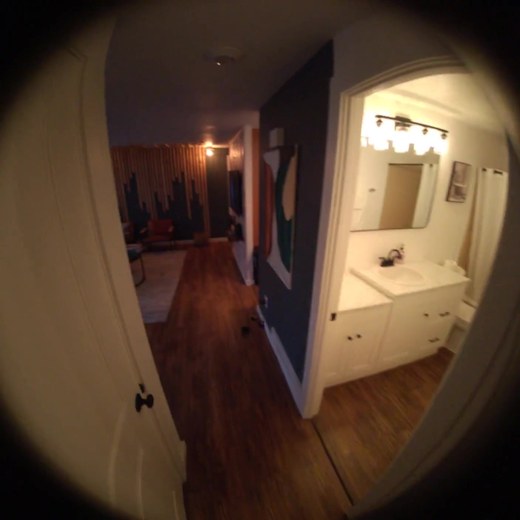} & \demoimg{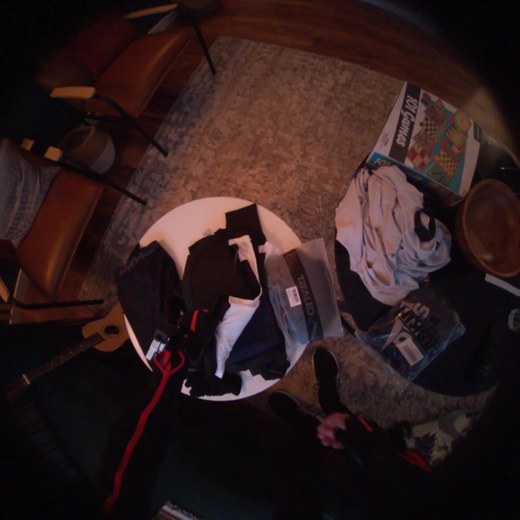} & \demoimg{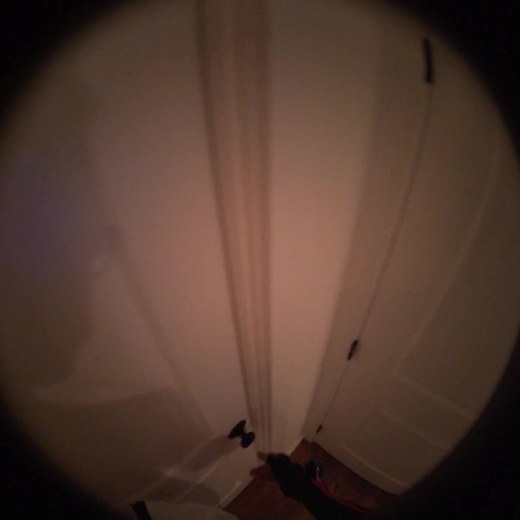} & \demoimg{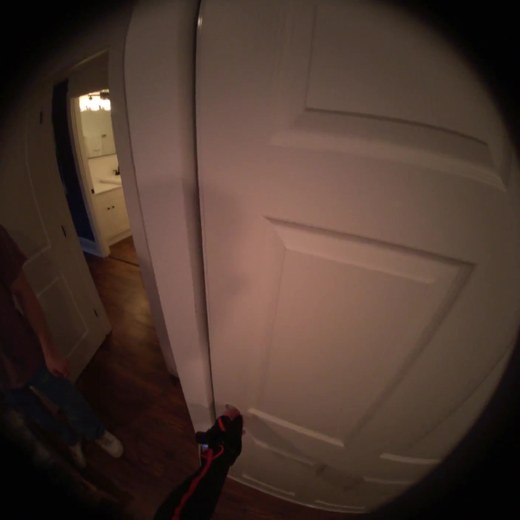}
\end{tabular}}\par}
\demotimeline{57}{0.0,8.0,16.2,24.2,32.5,40.5,48.8,56.8}
\demoq{The person enters the bedroom multiple times in this clip. What does the person do right after the last time they enter the bedroom?}
{\centering\small
\begin{tabularx}{0.95\textwidth}{@{}X@{}}
\demopt{A}{Takes a few steps forward and sits on the bed to the right}{} \\[2pt]
\demopt{B}{Puts the folded laundry on the bed to the right}{} \\[2pt]
\demopt{C}{Puts the folded laundry on the bed to the left}{temporalcolor} \\[2pt]
\demopt{D}{Stands straight, turns right, and exits the bedroom}{}
\end{tabularx}\par}
\end{democard}
\vspace{-2pt}\captionof{figure}{\small\textbf{Repeated Event Reasoning.} Nymeria clip, 57\,s.}
\label{fig:ex-repeated-event-reasoning-1}
\end{excard}
\flushbottom

\section{Question \& Answer Generation}
\label{app:appendix_datacreation}

\subsection{Selector and Generator Prompt Examples}
\label{app:prompts}

For each of the LLM-generated tasks we crafted a pair of task-specific prompts: one for the \textit{selector}, which scans the textual annotation streams of a video and returns sub-sequences that support a valid question, and one for the \textit{generator}, which drafts a single multiple-choice question from the selected sub-sequence. The strategy for constructing distractors is likewise task-specific. \textit{Bimanual Coordination}, for instance, swaps the role assignment between the two hands, \textit{Adjacent Action Identification} substitutes plausible alternative actions from the same activity, and analogous strategies apply to the remaining tasks. We show the two prompts for the \textit{Repeated Event Reasoning} task below as a representative example; the complete prompts for all tasks will be released with the code. Long or repetitive sections (e.g., the full list of refusal conditions) are abbreviated as ``[\ldots]''.

\begin{tcolorbox}[
  breakable, enhanced, sharp corners,
  colback=gray!4, colframe=gray!50!black,
  title=\textbf{Selector Prompt: Repeated Event Reasoning},
  fonttitle=\small\bfseries,
  left=4pt, right=4pt, top=4pt, bottom=4pt,
  boxrule=0.5pt,
]
\ttfamily\footnotesize\raggedright
\# Role\\[2pt]
You are an expert at selecting sub-sequences from egocentric-video atomic-action annotations for generating Repeated-Event Reasoning multiple-choice questions.\\[4pt] \smallskip
\# What this category tests\\[2pt]
Whether a vision-language model can track a RECURRING event across a long video, index into a specific occurrence ("first time", "second time", "last time"), and reason about what happens right after that particular occurrence.
It is not counting, not single-action reasoning, and not generic before/after ordering.\\[2pt] \smallskip
Representative question types:\\
- "The person picks up their phone multiple times during the session. What happens right after the FIRST time they do this?"\\
- "After the SECOND time the person opens the refrigerator, what do they do next?"\\[4pt] \smallskip
\# Input\\[2pt]
Every atomic action annotation for one video, numbered [0], [1], [2], ..., in chronological order. [\ldots]\\[4pt] \smallskip
\# Your job\\[2pt]
Return sub-sequences (contiguous index ranges) that contain a recurring event ($\geq$2 occurrences) AND concrete successor content after at least two of those occurrences.\\[4pt] \smallskip
\#\# What counts as a recurring event\\
A specific, nameable action that occurs $\geq$2 times in the span and is clearly the same action each time (same verb, same object, same intent). Examples: picks up a phone, opens/closes the refrigerator, transfers an object from one hand to the other. The verb/object pair must be stable across occurrences.\\[4pt]
\#\# What makes a good sub-sequence\\
A good span has ALL of:\\
1. The same event appears 3+ times, with clearly different occurrences (3 strongly preferred so the generator has enough material for distractors).\\
2. Concrete successor content after each occurrence (a specific physical action, not just "leans forward" or "turns around").\\
3. Distinct successors across occurrences (if all three are followed by the same thing, there is no interesting distinction to ask about).\\
4. Occurrences meaningfully separated in time ($\geq$10s between consecutive ones).\\[4pt] \smallskip

\#\# What to AVOID\\
- Spans where no event clearly recurs.\\
- Recurring postures or generic motion ("walking", "standing").\\
- Identical successor content after every occurrence.\\
- Back-to-back duplicate annotations (same moment re-described). [\ldots]\\[4pt] \smallskip
\# Output format\\[2pt]
Return ONLY a JSON array, no prose, no code fences:\\
\{"start\_idx": <int>, "end\_idx": <int>\}\\[2pt]
If the video has no recurring event with distinct successor content, return [].
\end{tcolorbox}

\begin{tcolorbox}[
  breakable, enhanced, sharp corners,
  colback=gray!4, colframe=gray!50!black,
  title=\textbf{Generator Prompt: Repeated Event Reasoning},
  fonttitle=\small\bfseries,
  left=4pt, right=4pt, top=4pt, bottom=4pt,
  boxrule=0.5pt,
]
\ttfamily\footnotesize\raggedright
\# Role\\[2pt]
You are an expert in designing multiple-choice questions for the Repeated-Event Reasoning category: questions that reference a RECURRING event (multiple occurrences in the clip) and an INDEXING word (first / second / last) to pick out a particular occurrence, then ask what the person does right after that occurrence.\\[4pt] \smallskip
\# What this benchmark evaluates\\[2pt]
A good Repeated-Event question is one where:\\
- A human watching the clip can answer by identifying all occurrences and looking at what follows the specified one.\\
- A model with weak temporal understanding gets it wrong.\\
- A blind reader (question only, no video) cannot guess the answer.\\[4pt] \smallskip
\# Valid question patterns\\
Every question has the shape: "After the \{INDEX\} time the person \{recurring event\}, what does the person do next?" Main subject: "the person", 3rd-person singular.\\
- Pattern A: First occurrence.\\
- Pattern B: Last occurrence.\\
- Pattern C: Specific middle occurrence (only when the span contains 3+ occurrences).\\[4pt] \smallskip
\# Procedure\\
Step 1. Find the recurring event (same verb/object pair, with multiple separated occurrences).\\
Step 2. Identify the successor of each occurrence.\\
Step 3. Pick the target occurrence whose successor is MOST DIFFERENT from the others, and the pattern (A/B/C).\\
Step 4. Draft the question. Explicitly state that the event recurs ("\ldots multiple times in this clip"), include the INDEX word.\\
Step 5. Blind-reader check: reject if the answer can be guessed from wording alone.\\
Step 6. Draft the correct answer: a 3--10-word phrase describing the FIRST specific action right after the target occurrence, in 3rd-person singular present tense.\\
Step 7. Draft 3 distractors. Each must (a) describe a concrete physical action (no transitional verbs like "moves", "leans", "turns" without a named destination), and (b) be temporally possible at the moment the answer happens.\\
At least TWO distractors must be immediate successors of OTHER occurrences of the SAME recurring event. [\ldots].\\[4pt] \smallskip
\# Worked example for Pattern A (first occurrence)\\[2pt]
Input span:\\
- C walks to the kitchen and opens the refrigerator.\\
- C takes a bottle of milk out of the refrigerator.\\
- C places the bottle on the counter and closes the refrigerator.\\
- C pours milk into a glass.\\
- C walks back to the refrigerator and opens it.\\
- C puts the bottle of milk back into the refrigerator.\\
- C closes the refrigerator.\\
- [\ldots\ one more open/close cycle]\\[2pt]
Recurring event: "open the refrigerator" occurs 3 times. After 1st: takes a bottle out. After 2nd: puts the bottle back. After 3rd: closes it and walks to the living room.\\[2pt]
Expected output:\\
\{"question": "The person opens the refrigerator multiple times in this clip. What does the person do right after the FIRST time they open it?",\\
\phantom{\{}"answer": "Takes a bottle of milk out of the refrigerator",\\
\phantom{\{}"distractors": ["Puts the bottle of milk back into the refrigerator", "Closes the refrigerator and walks to the living room", "Pours milk into a glass"]\}\\[2pt]
[\ldots\ one more worked example for Pattern B (last occurrence)]\\[4pt] \smallskip
\# Output format\\[2pt]
Return exactly:\\
\{"question": "<your question>", "answer": "<correct option>",\\
\phantom{\{}"distractors": [<3 distractors>]\}\\[4pt] \smallskip
\# Refusal conditions\\[2pt]
Return \{"question": null\} if any hold:\\
R1. No event recurs multiple times with the same verb/object pair.\\
R2. All occurrences are followed by the same successor content.\\
R3. The answer is an obvious continuation of the event (blind-reader leak).\\
R4. Cannot produce 3 distractors where at least one is a successor of a DIFFERENT occurrence.  [\ldots]
\end{tcolorbox}

\subsection{Mitigating Generator Bias}
\label{app:generator_bias}

Benchmark questions generated by an LLM may inherit regularities of the generating model in their phrasing, answer formulation, and distractor construction, which could advantage models from the same family at evaluation time, a concern analogous to self-preference in LLM-based evaluation~\citep{panickssery2024llmevaluatorsrecognizefavor}. Since Gemini 2.5 Flash is both our generator and one of the evaluated models, we took two measures to mitigate such bias.
First, the question generation stage is not an open-ended LLM writing process: the model fills fixed task-specific question templates, while both the correct answer and the distractors are derived from the ground-truth annotations under predefined rules (Appendix~\ref{app:prompts}). The generator therefore has limited freedom to introduce stylistic regularities of its own, and the resulting questions largely reflect the source annotations rather than the generating model.
Second, Gemini 2.5 Flash is one of the four blind solvers in the filtering stage (Sec.~\ref{sec:pipeline}). Questions that Gemini answers correctly without the video, together with at least one other blind solver, are therefore removed during construction, before any MLLM is evaluated.

\subsection{Manual Annotation of Repetition Counting}
\label{app:rep_counting}

For each selected take, we used the 3D body-pose data to identify segments containing repetitive motion; a human annotator then inspected each candidate segment, examining the video together with the synchronized 3D body-pose rendering, and created the question and recorded the count for each.
One repetition is defined as a single completed instance of the action named in the question. Segments containing partial repetitions, occluded repetitions, or otherwise unclear repetition boundaries are discarded rather than annotated with an uncertain count, so a repetition is counted only when it is clearly visible. Examples are available in Appendix~\ref{app:examples}.
This adds 101 Repetition Counting questions to the candidate set.

\clearpage

\subsection{Filtering and Review Statistics}
\label{app:filtering_stats}

Across the eleven automatically generated tasks, our pipeline produced 3{,}928 candidate questions. The blind-language filter removed questions answered correctly by at least two of four solvers, using the prompt shown below.

\begin{tcolorbox}[
  breakable, enhanced, sharp corners,
  colback=gray!4, colframe=gray!50!black,
  title=\textbf{Blind Solver Prompt},
  fonttitle=\small\bfseries,
  left=4pt, right=4pt, top=4pt, bottom=4pt,
  boxrule=0.5pt,
]
\ttfamily\footnotesize\raggedright
You are answering a multiple-choice question about a first-person (``egocentric'') video that you have NOT been shown.\\[4pt] \smallskip
You will see only the question text and four options labeled A, B, C, D. Use your world knowledge, common sense, and any clues in the wording of the question itself to pick the option you think is most likely correct. Commit to exactly one letter even if uncertain --- guess if you have to.\\[4pt] \smallskip
Respond with ONLY a JSON object with the single key ``answer'', whose value is one of ``A'', ``B'', ``C'', ``D'':\\[4pt]
\{``answer'': ``X''\}\\[4pt] \smallskip
Do not include any other text, explanation, markdown, or extra keys.
\end{tcolorbox}

This process removed 1{,}801 questions (45.9\%), leaving the remaining 2,127 for human review. Removal rates were highest for Body Posture (62.4\%) and Hand Use (53.7\%), where the answer space is small and natural language priors were strongest, and lowest for Repeated Event Reasoning (30.8\%) and Chronological Ordering (39.8\%), where the correct answer depended on multi-event temporal structure that is difficult to recover from text alone.

During human review, 945 questions were accepted as-is, 586 were modified, and 596 were deleted. Combined with the 101 manually annotated Repetition Counting questions (Appendix~\ref{app:rep_counting}), the final benchmark contains 1{,}632 multiple-choice questions.

\section{Evaluation Details}
\label{app:eval_details}

\subsection{Details on Human evaluation}
\label{app:human_eval}
Human performance was measured as the average accuracy of five human subjects who were not involved in the construction or annotation of the benchmark. Each subject independently evaluated the full set of 1,632 samples. For every sample, annotators were presented with the video, question, and four answer options, and were allowed unlimited time to select the single best answer based on the visual information in the video. When the camera wearer’s body is not directly visible, annotators were instructed to rely on indirect visual cues, such as camera motion, viewpoint changes, and hand-object interactions. Completing the full evaluation required 89, 90, 94, 97, and 95 hours for the five annotators, respectively, corresponding to an average of 93 hours per annotator and 465 hours of human evaluation in total. Annotators were compensated for their time spent completing the evaluation. 

We report the standard deviation in accuracy across the five human annotators for each task (Table~\ref{tab:human_std}). Human performance is most consistent on Hand Use and varies most on Chronological Ordering and Missing Event Inference. 

\setlength{\tabcolsep}{3.5pt}
\newcolumntype{?}{!{\color{black!15}\vrule width 0.5pt}}
\begin{table}[h]
\caption{%
\textbf{Variability of Human Performance.}
Accuracy (\%) on each SOMA task, reported as mean\,$\pm$\,standard deviation
across the five annotators. Task abbreviations follow
Table~\ref{tab:SOMA_results}.
}
\label{tab:human_std}
\vspace{0.3\baselineskip}
\centering
\resizebox{\textwidth}{!}{
\begin{tabular}{c?c?c?c|c?c?c|c?c?c?c?c|c}
\toprule
\multicolumn{4}{c|}{\textbf{\textcolor{bodyconfigcolor}{Body Configuration}}}
& \multicolumn{3}{c|}{\textbf{\textcolor{motiondynamicscolor}{Motion Dynamics}}}
& \multicolumn{5}{c|}{\textbf{\textcolor{temporalcolor}{Temporal Reasoning}}}
& \multirow{2}{*}{\textbf{Avg.}} \\
HU & BC & BP & CBS
& RC & MD & OMD
& AAI & CO & MEI & NAP & RER
& \\
\midrule
90.5\,$\pm$\,1.8 & 78.3\,$\pm$\,3.7 & 61.9\,$\pm$\,8.0 & 50.3\,$\pm$\,9.2
& 88.9\,$\pm$\,4.3 & 66.3\,$\pm$\,9.1 & 86.1\,$\pm$\,8.5
& 70.1\,$\pm$\,7.3 & 53.2\,$\pm$\,14.4 & 72.7\,$\pm$\,15.6 & 34.5\,$\pm$\,4.9 & 67.4\,$\pm$\,5.9
& 68.5\,$\pm$\,5.5 \\
\bottomrule
\end{tabular}}
\end{table}

\subsection{Evaluated Models}
\label{app:evaluated-models}

We conducted a comprehensive evaluation spanning closed-source, open-source, and specialized egocentric MLLMs, covering diverse model scales and recent architectural advances. For closed-source models, we included Gemini 2.5 Flash and Gemini 2.5 Pro~\citep{comanici2025gemini}, Gemini 3.5 Flash~\citep{gemini35flash2026}, GPT-5.1~\citep{openai_gpt51_2025} and GPT-5.6 Luna~\citep{gpt56luna2026}, and Claude Sonnet 5~\citep{claudesonnet52026} and Claude Opus 4.8~\citep{claudeopus482026}. For open-source models, we evaluated Qwen3-VL (2B--32B)~\citep{bai2025qwen3vl}, Qwen3.5 (0.8B--27B)~\citep{qwen3.5}, Qwen3.6-27B~\citep{qwen3.6-27b}, and Qwen3.8-27B~\citep{qwen38}, InternVL3.5 (1B--38B)~\citep{wang2025internvl3_5}, Gemma-4 (12B, 31B)~\citep{team2026gemma}, GLM-4.6V and GLM-4.6V-Flash~\citep{vteam2025glm45vglm41vthinkingversatilemultimodal}, Molmo2 (4B, 8B)~\citep{clark2026molmo2}, Kimi-VL-A3B~\citep{kimiteam2025kimivltechnicalreport}, MiniCPM-o-4.5-9B ~\citep{cui2026minicpmo45realtimefullduplex}, LLaVA-OneVision-1.5-8B ~\citep{an2025llava}, and VideoLLaMA3-7B ~\citep{damonlpsg2025videollama3}. For specialized egocentric models, we included EgoGPT-7B ~\citep{yang_egolife_2025}, EgoVLM-3B ~\citep{vinod2025egovlmpolicyoptimizationegocentric}, and EgoThinker-7B ~\citep{pei2026egothinker}.

\subsection{Implementation Details}
\label{app:implementation_details}

We uniformly sampled 16 frames per video segment across all models to balance computational cost and comparability across models with different context lengths; Sec.~\ref{sec:ablations} studies the effect of denser sampling on performance. We used \texttt{lmms-eval}~\citep{zhang2024lmmsevalrealitycheckevaluation}, a standardized framework for evaluating multimodal models across a wide range of tasks. We instructed all models to respond with a single letter corresponding to their chosen option; when a response requires further parsing, we used the common method that other multiple-choice tasks in \texttt{lmms-eval} apply, which checks the response against a set of common answer-formatting patterns to identify the selected option. All open-source model experiments were run on either 4 H100 or 4 A100 GPUs.

\subsection{Zero-Shot Evaluation Prompt Template}
\label{app:eval_prompt}

Every model received the same user prompt, shown below, instantiated per question by inserting the question text and the four answer options. Options were presented in the order stored in the benchmark, which is shuffled once at benchmark-creation time.

\begin{tcolorbox}[
  enhanced, sharp corners,
  colback=gray!4, colframe=gray!50!black,
  title=\textbf{Evaluation Prompt Template},
  fonttitle=\small\bfseries,
  left=4pt, right=4pt, top=4pt, bottom=4pt,
  boxrule=0.5pt,
]
\ttfamily\footnotesize\raggedright
Answer the following multiple-choice question using the given video.\\[\baselineskip]
Question: \{question\}\\
Choices:\\
A. \{option A\}\\
B. \{option B\}\\
C. \{option C\}\\
D. \{option D\}\\[\baselineskip]
Respond only with the letter of the correct answer.
\end{tcolorbox}

\noindent The model's free-form response is then mapped to one of \{A, B, C, D\} by the standard \texttt{lmms-eval} multiple-choice parser, and counted as correct if it matches the correct answer.

\subsection{Chain-of-Thought Evaluation Prompt Template}
\label{app:cot_prompt}

The chain-of-thought (CoT) ablation keeps the original question and answer options unchanged from the zero-shot prompt (Appendix~\ref{app:eval_prompt}), and replaces the final instruction with the reasoning instruction shown below.

\begin{tcolorbox}[
  enhanced, sharp corners,
  colback=gray!4, colframe=gray!50!black,
  title=\textbf{Chain-of-Thought Prompt Template},
  fonttitle=\small\bfseries,
  left=4pt, right=4pt, top=4pt, bottom=4pt,
  boxrule=0.5pt,
]
\ttfamily\footnotesize\raggedright
Answer the following multiple-choice question using the given video.\\[\baselineskip]
Question: \{question\}\\
Choices:\\
A. \{option A\}\\
B. \{option B\}\\
C. \{option C\}\\
D. \{option D\}\\[\baselineskip]
Before answering, work through the question step by step. First, identify what the question is asking. Then carefully observe the video and briefly describe the observations relevant to the question. Combine these observations to evaluate each option, explaining why it is or is not supported. Finally, choose the option best supported by your analysis and end your response with a final line in the format "Answer: X", where X is the letter of the correct answer.
\end{tcolorbox}

\section{Additional Results}
\label{app:additional_results}

\subsection{Effect of Chain-of-Thought Prompting}
\label{app:cot-prompting}

Test-time reasoning can also be invoked without a dedicated thinking mode, by instructing a model to reason step by step before answering~\citep{wei2022chain}. We evaluated five open-source models by extending the original prompt with a CoT instruction that asks the model to observe, reason, then answer (Appendix~\ref{app:cot_prompt}).

Reasoning step-by-step did not systematically help on SOMA (Table~\ref{tab:SOMA_cot}). Only Qwen3.8-27B benefited, gaining 3.9 points and improving on 11 of 12 tasks, with the largest gains in temporal reasoning and motion dynamics. For the remaining models, CoT had a neutral effect or reduced performance. This mirrors the thinking-mode result (Figure~\ref{fig:effect-analyis}) and suggests that current reasoning mechanisms do not address the difficulty of perceiving body motion, and additional reasoning tokens do little to help.

\begin{table*}[h]
\setlength{\tabcolsep}{4pt}
\renewcommand{\arraystretch}{1.03}
\caption{
\textbf{Effect of Chain-of-Thought Prompting on SOMA.}
Accuracy (\%) across the three capability axes.
For each model, the top row is direct answering (\textit{w/o CoT}) and the bottom row asks the model to reason step by step before answering (\textit{w/ CoT}).
The better of the two is marked in \textbf{bold}.
Task abbreviations follow Table~\ref{tab:SOMA_results}.
}
\label{tab:SOMA_cot}
\vspace{0.5\baselineskip}
\centering
\scriptsize
\resizebox{\textwidth}{!}{
\begin{tabular}{ll|cccc|ccc|ccccc|c}
\toprule
\multirow{2}{*}{\textbf{Models}} & \multirow{2}{*}{\textbf{Setting}}
& \multicolumn{4}{c|}{\textbf{\textcolor{bodyconfigcolor}{Body Configuration}}} 
& \multicolumn{3}{c|}{\textbf{\textcolor{motiondynamicscolor}{Motion Dynamics}}} 
& \multicolumn{5}{c|}{\textbf{\textcolor{temporalcolor}{Temporal Reasoning}}} 
& \multirow{2}{*}{\textbf{Avg.}} \\
& & HU & BC & BP & CBS
& RC & MD & OMD
& AAI & CO & MEI & NAP & RER
& \\
\midrule
\multirow{2}{*}{Qwen3.8-27B}
& w/o CoT 
& 72.1 & 47.7 & \textbf{46.1} & 44.1
& 36.6 & 40.5 & 58.6
& 44.3 & 34.7 & 48.3 & 39.0 & 37.0
& 47.0 \\
& w/\hphantom{o} CoT 
& \textbf{73.1} & \textbf{49.0} & 43.5 & \textbf{49.5}
& \textbf{41.6} & \textbf{49.0} & \textbf{59.3}
& \textbf{51.1} & \textbf{46.3} & \textbf{52.5} & \textbf{40.9} & \textbf{40.0}
& \textbf{50.9} \\
\midrule
\multirow{2}{*}{Gemma-4-31B}
& w/o CoT 
& \textbf{67.2} & 41.1 & 39.1 & \textbf{46.8}
& \textbf{37.6} & 34.6 & \textbf{50.7}
& 38.2 & 38.8 & 30.5 & \textbf{40.2} & \textbf{38.0}
& \textbf{43.1} \\
& w/\hphantom{o} CoT 
& 58.2 & \textbf{42.4} & \textbf{42.6} & 44.1
& 25.7 & \textbf{39.2} & 47.9
& \textbf{43.5} & \textbf{41.5} & \textbf{39.8} & 33.5 & 36.0
& 42.2 \\
\midrule
\multirow{2}{*}{InternVL3.5-38B}
& w/o CoT 
& \textbf{52.2} & \textbf{44.4} & 37.4 & 32.4
& \textbf{36.6} & 30.1 & \textbf{59.3}
& \textbf{42.0} & 29.9 & \textbf{45.8} & 17.7 & \textbf{38.0}
& \textbf{39.0} \\
& w/\hphantom{o} CoT 
& 45.8 & 36.4 & \textbf{42.6} & \textbf{36.0}
& 35.6 & 30.1 & 48.6
& 40.5 & \textbf{32.0} & 36.4 & \textbf{21.3} & 32.0
& 36.5 \\
\midrule    
\multirow{2}{*}{MiniCPM-o-4.5-9B}
& w/o CoT 
& \textbf{55.2} & 35.8 & 40.9 & \textbf{46.8}
& \textbf{32.7} & 29.4 & 42.9
& 35.1 & \textbf{30.6} & 26.3 & \textbf{23.2} & 30.0
& \textbf{36.3} \\
& w/\hphantom{o} CoT 
& 47.8 & \textbf{37.7} & \textbf{47.0} & 42.3
& 31.7 & \textbf{35.3} & \textbf{43.6}
& \textbf{35.9} & 29.9 & \textbf{27.1} & 18.9 & \textbf{36.0}
& 36.2 \\
\midrule
\multirow{2}{*}{Molmo2-8B}
& w/o CoT 
& \textbf{54.2} & \textbf{42.4} & \textbf{51.3} & \textbf{41.4}
& \textbf{33.7} & \textbf{36.6} & \textbf{39.3}
& \textbf{39.7} & \textbf{38.1} & \textbf{37.3} & \textbf{36.0} & 29.0
& \textbf{40.6} \\
& w/\hphantom{o} CoT 
& 37.3 & 32.5 & 37.4 & 34.2
& 30.7 & 20.9 & 37.1
& 30.5 & 27.2 & 20.3 & 26.8 & \textbf{34.0}
& 30.8 \\
\bottomrule
\end{tabular}
}
\end{table*}

\clearpage
\subsection{Inference-Time Analysis per Capability Axis}
\label{app:inference-time-per-category}

\subsubsection{Effect of Temporal Sampling Density}
\label{app:sampling-density-category}
Denser frame sampling does not help the three capability axes equally. For Body Configuration tasks, where questions ask about the body's state at a specific moment, accuracy saturates early. A possible explanation is that the relevant evidence for these questions is concentrated in a small number of frames, so additional frames beyond that point add little new information, even though correctly extracting that evidence remains difficult. For Motion Dynamics tasks, which ask how the body moves and are hard to answer from only a few frames, accuracy continues to improve for most models, though only marginally beyond 16 frames. Temporal Reasoning remains the weakest axis at every sampling density, suggesting that regardless of how many frames the model is given, it struggles to integrate that information over time.

\begin{figure}[h]
    \centering
    \includegraphics[width=\textwidth]{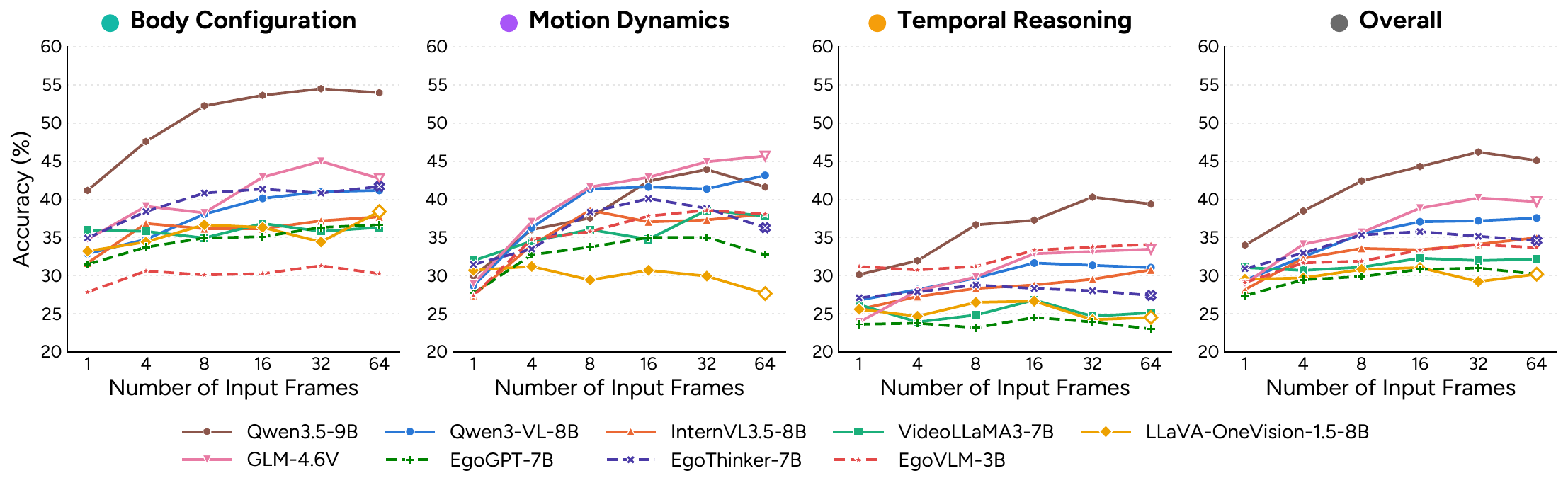}
    \vspace{-.3cm}
    \caption{Effect of temporal sampling density on accuracy, per capability axis.}
    \label{fig:frames-category}
\end{figure}

\subsubsection{Effect of Test-Time Reasoning}
\label{app:test-time-reasoning-category}

The effect of test-time reasoning also varies by axis. Gains are largest on Motion Dynamics, where most models improve; these tasks often require counting or distinguishing observed motion, which can benefit from explicit reasoning once the motion itself has been perceived. On Body Configuration, which is the most perceptual of the three axes, most models show negligible change or a decline. Changes on Temporal Reasoning are modest overall. This suggests that reasoning helps once the relevant evidence has been observed, but does not compensate for a failure to perceive it in the first place.

\begin{figure}[h]
    \centering
    \includegraphics[width=\textwidth]{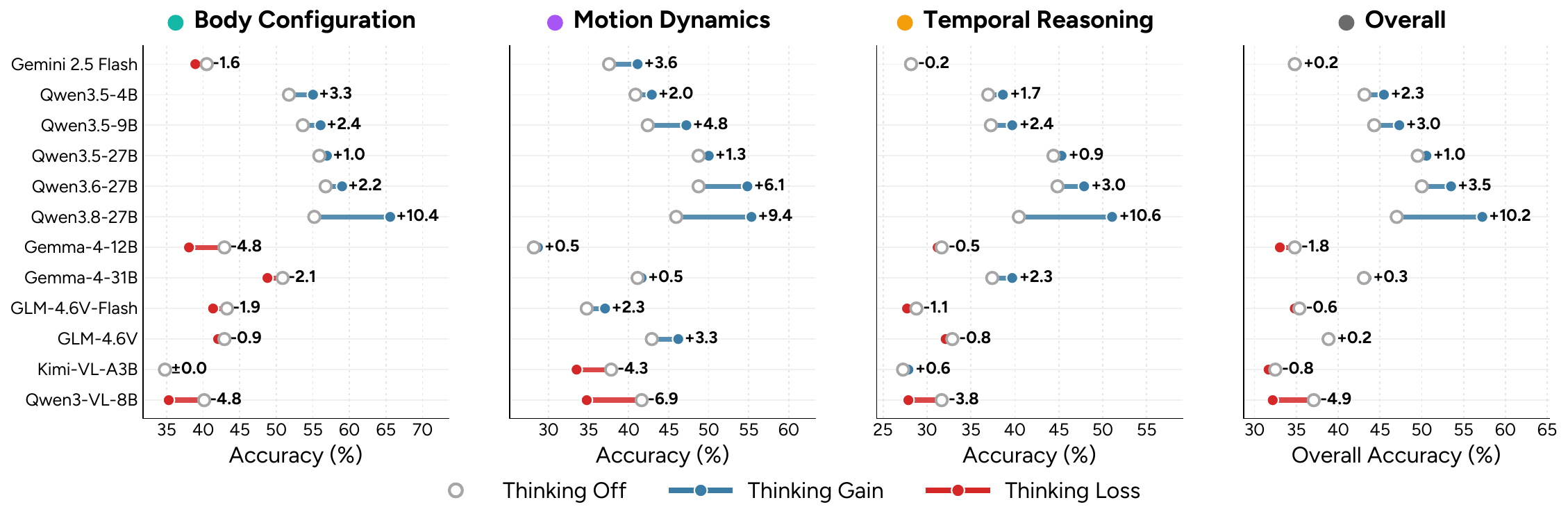}
    \vspace{-.3cm}
    \caption{Effect of test-time reasoning on accuracy, per capability axis.}
    \label{fig:thinking-category}
\end{figure}

\newpage
\subsection{Performance by Question Difficulty}
\label{app:question-difficulty}

\begin{wrapfigure}{r}{0.52\textwidth}
    \vspace{-.7cm}
    \centering
    \includegraphics[width=\linewidth]{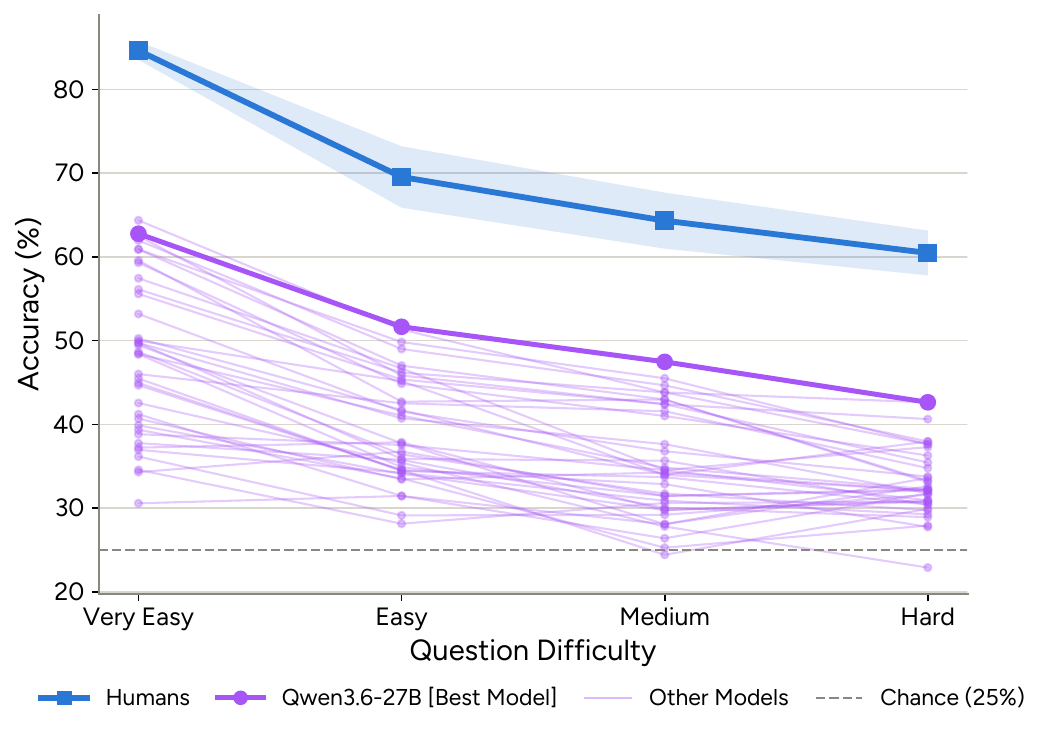}
    \vspace{-.3cm}
    \caption{\textbf{Human and model performance across question difficulty levels.} The human curve is the mean over five annotators with the band showing $\pm 1$ SE.}
    \label{fig:question-difficulty}
    \vspace{-.4cm}
\end{wrapfigure}

Questions in SOMA differ in how demanding they are. To understand how models and humans perform across different difficulty levels, we adopted the concept of psychometric curves~\citep{boring1917chart}, following~\citet{qi2025llavaction}. We used Gemini~2.5~Pro to rate each question from 1 (very easy) to 4 (hard) according to how difficult it is to distinguish between the options, using the prompt shown below, and we report the accuracy of every evaluated model and our human annotators at each level (Figure~\ref{fig:question-difficulty}). Accuracy falls with difficulty for both humans and all models, with humans remaining far ahead at every level. The gap is largest on the easiest questions, showing that models fall short even on simple cases and across the full difficulty range of the benchmark.

\begin{tcolorbox}[
  float=tbp,
  enhanced, sharp corners,
  colback=gray!4, colframe=gray!50!black,
  title=\textbf{Difficulty Rating Prompt Template},
  fonttitle=\small\bfseries,
  left=4pt, right=4pt, top=4pt, bottom=4pt,
  boxrule=0.5pt,
  segmentation style={dashed, gray!50!black},
]
\ttfamily\footnotesize\raggedright
{\normalfont\small\bfseries System prompt}\\[0.5\baselineskip]
You are analyzing a multiple-choice question from an egocentric (first-person) video benchmark on human motion and activity. Your task is to rate the difficulty of distinguishing between the options based on how similar they are to each other and to the ground truth answer.\\[\baselineskip]
You will be given the question, the four options, and the ground truth answer. Rate how fine-grained an observation is needed to separate the correct answer from the distractors, not how guessable the answer is from the wording alone.\\[\baselineskip]
Please analyze the semantic similarity between the options and rate the difficulty on a scale of 1-4:\\
- 1 (Very Easy): Options are very different from each other and the correct answer is obvious\\
- 2 (Easy): Options have clear differences, the correct answer is fairly obvious\\
- 3 (Medium): Options have moderate similarity, requires some careful consideration\\
- 4 (Hard): Options are quite similar, subtle differences make it challenging\\[\baselineskip]
Consider factors like:\\
- Semantic similarity between the action descriptions\\
- Specificity vs generality of the actions\\
- Whether options describe similar but distinct actions\\
- How confusable the distractors are with the ground truth\\
- How fine-grained an observation is needed to rule out the closest distractor\\
- Whether the answer is visible from an egocentric perspective or requires inference\\[\baselineskip]
Respond with only a single number (1-4) representing the difficulty score.
\tcblower
\ttfamily\footnotesize\raggedright
{\normalfont\small\bfseries User message}\\[0.5\baselineskip]
Category: \{category\}\\[\baselineskip]
Question: \{question\}\\[\baselineskip]
Options:\\
A. \{option A\}\\
B. \{option B\}\\
C. \{option C\}\\
D. \{option D\}\\[\baselineskip]
Ground truth answer: \{answer\}\\[\baselineskip]
Rate the difficulty of distinguishing these options (1-4) as instructed.
\end{tcolorbox}

\subsection{Error Analysis}
\label{app:error-analysis-details}

Here we provide the full definitions of the four error types used in Sec.~\ref{sec:error_analysis}, which we identified through manual inspection of a subset of incorrect predictions, and describe the annotation procedure. Each incorrect answer is assigned the single error type that most directly leads to the chosen option:
\begin{itemize}
\item \textbf{Visual Grounding Error.} The model misidentifies what or who is involved. For example, it confuses or overlooks an object or scene element, or attributes the action to another visible person instead of the camera wearer.
\item \textbf{Motion and Body-State Perception Error.} The model misperceives what the wearer's own body does, such as their locomotion, movement direction, posture, the limb used, or the action performed, as well as body actions it fails to notice at all.
\item \textbf{Temporal Localization Error.} The model recognizes the relevant events but places them wrongly in time. For instance, it anchors on the wrong moment or reverses their order.
\item \textbf{Prior-Driven Inference.} The answer follows from expectation rather than evidence, such as procedural common sense, default assumptions, reinterpreted question or option wording, or assumptions about the dataset, and may rely on observations invented to support that expectation.
\end{itemize}

Since manually reading all 400 long reasoning traces is extremely labor-intensive, we combined automated labeling with manual verification, following similar approaches in the literature~\citep{yang2026mmsi, nagrani2025minerva}. Moreover, every clip in SOMA comes with ground-truth atomic action and motion narration annotations, so each claim in a reasoning trace can be compared with the annotated ground truth. We therefore used LLM judges: given the question, the judge compares the model's reasoning and chosen answer against the ground-truth annotations, together with the correct answer, and assigns one of the four error types, quoting the excerpt of the reasoning that shows the error.

To ensure that the labels do not depend on the biases of a single judge, we used two judges from different model families, GPT-5.1 and Gemini~2.5~Pro, each labeling all 400 traces independently. The agreement between the two judges is high, as they assign the same error type to 80.0\% of the traces (Cohen's $\kappa = 0.69$, 95\% bootstrap CI 0.63--0.75). We additionally verified a sample of the labels manually against the video clip and ground-truth annotations and found the assigned types and the quoted evidence consistent with the definitions above.

\clearpage
\raggedbottom
\subsubsection{Qualitative Examples}
\label{app:error-qualitative-examples}
For each error type, we show a representative failure case in Figures~\ref{app:visual_grounding_error}--\ref{app:prior_driven_inference_error}.

\usetikzlibrary{calc}
\definecolor{eataskadjacentactionidentification}{HTML}{D2890C}
\definecolor{eataskchronologicalordering}{HTML}{B97808}
\definecolor{eataskmissingeventinference}{HTML}{9F6805}
\definecolor{eatasknextactionprediction}{HTML}{875705}
\definecolor{eataskrepeatedeventreasoning}{HTML}{6E4706}
\definecolor{eataskrepetitioncounting}{HTML}{9049D3}
\definecolor{eataskmovementdirection}{HTML}{6F38A1}
\definecolor{eataskoppositemotiondiscrimination}{HTML}{4D256E}
\definecolor{eataskhanduse}{HTML}{0D9F8F}
\definecolor{eataskbimanualcoordination}{HTML}{0C8578}
\definecolor{eataskbodyposture}{HTML}{0C6B61}
\definecolor{eataskconcurrentbodypartstate}{HTML}{0D534C}
\definecolor{eamuted}{HTML}{6B747D}
\definecolor{eafilm}{HTML}{15191C}
\definecolor{eaframe}{HTML}{1E3A40}
\definecolor{eawrong}{HTML}{C0392B}
\definecolor{earight}{HTML}{1E8449}
\newlength{\eaframewidth}

\newcommand{\eaimg}[2]{%
  \begin{tikzpicture}
    \path[clip, rounded corners=5pt]
      (0,0) rectangle (\eaframewidth,\eaframewidth);
    \node[anchor=south west, inner sep=0pt] at (0,0)
      {\includegraphics[width=\eaframewidth]{#1}};
    \ifx\relax#2\relax\else
      \fill[#2] (0,0) rectangle (\eaframewidth,2pt);
    \fi
  \end{tikzpicture}}

\newcommand{\eastrip}[1]{%
  \begin{tikzpicture}
    \node[inner xsep=5pt, inner ysep=6pt, fill=eafilm, rounded corners=3pt] (g) {\raisebox{0pt}{#1}};
    \foreach \p in {0.018,0.058,...,0.99}{%
      \fill[white, rounded corners=0.9pt]
        ($(g.north west)!\p!(g.north east) + (-1.7pt,-4.6pt)$) rectangle ++(3.4pt,3.2pt);
      \fill[white, rounded corners=0.9pt]
        ($(g.south west)!\p!(g.south east) + (-1.7pt,1.4pt)$) rectangle ++(3.4pt,3.2pt);}
  \end{tikzpicture}}

\newcommand{\earule}{\par\vspace{3pt}\noindent\textcolor{black!12}{\rule{\linewidth}{0.6pt}}\par\vspace{3pt}}

\newenvironment{eaoptions}
  {\noindent\begin{tabular}{@{}p{0.47\linewidth}@{\hspace{0.04\linewidth}}p{0.47\linewidth}@{}}}
  {\end{tabular}\par}

\newcommand{\eatrace}[1]{%
  \noindent{\scriptsize\textcolor{eamuted}{\textsc{Reasoning trace}}}\par\vspace{2pt}%
  \noindent{\scriptsize\setlength{\parskip}{0pt}#1}}

\newenvironment{eacard}
  {\par\noindent\begin{minipage}{\linewidth}\centering}
  {\end{minipage}\par\vspace{7pt plus 2pt minus 4pt}}

\newcommand{\eanum}[2]{%
  \tikz[baseline=-0.5ex]{\node[circle, fill=#1, inner sep=0pt, minimum size=11pt,
    font=\scriptsize\sffamily\bfseries, text=white] {#2};}}

\newcommand{\ealogo}[1]{%
  \raisebox{-0.45ex}[0pt][0pt]{\includegraphics[height=13pt]{figures/logos/#1.png}}}
\newcommand{\eabadge}[1]{%
  \tikz[baseline=-0.6ex]{\node[rounded corners=2pt, draw=eawrong, line width=0.6pt,
    fill=eawrong!8, inner xsep=3pt, inner ysep=1.5pt,
    font=\tiny\sffamily\bfseries, text=eawrong] {#1};}}

\newcommand{\eaopt}[2]{\strut\textbf{#1.}~#2}
\newcommand{\eaoptright}[2]{%
  \strut\textbf{#1.}~\textcolor{earight}{\textbf{#2}}~\textcolor{earight}{\checkmark}}
\newcommand{\eaoptmodel}[3]{%
  \strut\textbf{#1.}~\textcolor{eawrong}{\textbf{#2}}~#3}

\begin{eacard}
\begin{tcolorbox}[enhanced, arc=3mm, boxrule=1.1pt,
  colframe=visualgroundingcolor, colback=white, left=8pt, right=8pt, top=13pt, bottom=8pt,
  boxsep=0pt, fontupper=\sffamily\small,
  attach boxed title to top center={yshift=-\tcboxedtitleheight/2},
  boxed title style={boxrule=1.1pt, arc=2.4mm, colframe=visualgroundingcolor,
                     colback=white, left=7pt, right=7pt, top=1.5pt, bottom=1.5pt},
  coltitle=visualgroundingcolor, fonttitle=\small\sffamily\bfseries,
  title={Visual Grounding Error}]
{\centering{\setlength{\eaframewidth}{\dimexpr(\linewidth-10pt-12\dimexpr3pt\relax-6\dimexpr9pt\relax)/7\relax}\setlength{\tabcolsep}{0pt}%
\eastrip{\eaimg{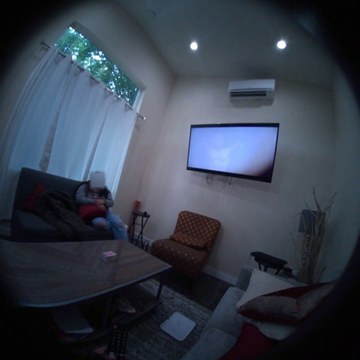}{\relax}\hspace{3pt}\makebox[9pt]{\raisebox{0.5\eaframewidth}[0pt][0pt]{\textcolor{white!55}{$\cdots$}}}\hspace{3pt}\eaimg{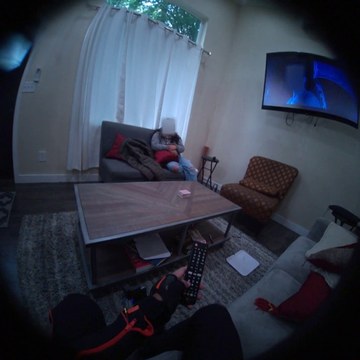}{\relax}\hspace{3pt}\makebox[9pt]{\raisebox{0.5\eaframewidth}[0pt][0pt]{\textcolor{white!55}{$\cdots$}}}\hspace{3pt}\eaimg{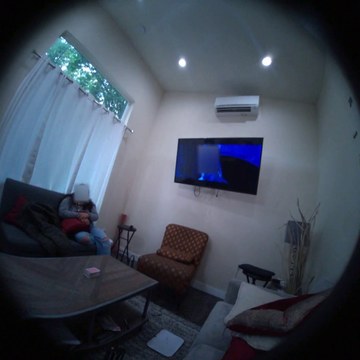}{\relax}\hspace{3pt}\makebox[9pt]{\raisebox{0.5\eaframewidth}[0pt][0pt]{\textcolor{white!55}{$\cdots$}}}\hspace{3pt}\eaimg{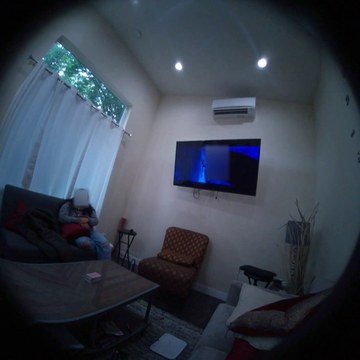}{\relax}\hspace{3pt}\makebox[9pt]{\raisebox{0.5\eaframewidth}[0pt][0pt]{\textcolor{white!55}{$\cdots$}}}\hspace{3pt}\eaimg{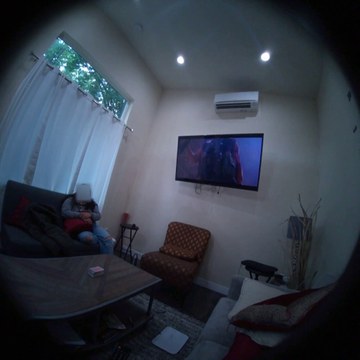}{\relax}\hspace{3pt}\makebox[9pt]{\raisebox{0.5\eaframewidth}[0pt][0pt]{\textcolor{white!55}{$\cdots$}}}\hspace{3pt}\eaimg{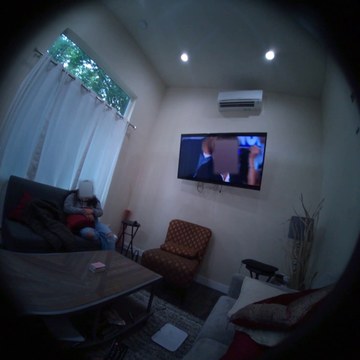}{\relax}\hspace{3pt}\makebox[9pt]{\raisebox{0.5\eaframewidth}[0pt][0pt]{\textcolor{white!55}{$\cdots$}}}\hspace{3pt}\eaimg{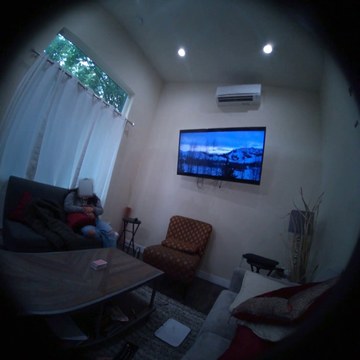}{\relax}}\par\vspace{-1pt}%
\hspace*{5pt}\makebox[\eaframewidth]{\eanum{eafilm}{1}}\hspace{3pt}\makebox[9pt]{}\hspace{3pt}\makebox[\eaframewidth]{\eanum{eafilm}{3}}\hspace{3pt}\makebox[9pt]{}\hspace{3pt}\makebox[\eaframewidth]{\eanum{eafilm}{6}}\hspace{3pt}\makebox[9pt]{}\hspace{3pt}\makebox[\eaframewidth]{\eanum{eafilm}{9}}\hspace{3pt}\makebox[9pt]{}\hspace{3pt}\makebox[\eaframewidth]{\eanum{eafilm}{11}}\hspace{3pt}\makebox[9pt]{}\hspace{3pt}\makebox[\eaframewidth]{\eanum{eafilm}{13}}\hspace{3pt}\makebox[9pt]{}\hspace{3pt}\makebox[\eaframewidth]{\eanum{eafilm}{16}}}\par}
\vspace{6pt}
{\centering\footnotesize\textcolor{eamuted}{\textcolor{eataskhanduse}{\textbf{H}and \textbf{U}se}\,\textperiodcentered\,\ealogo{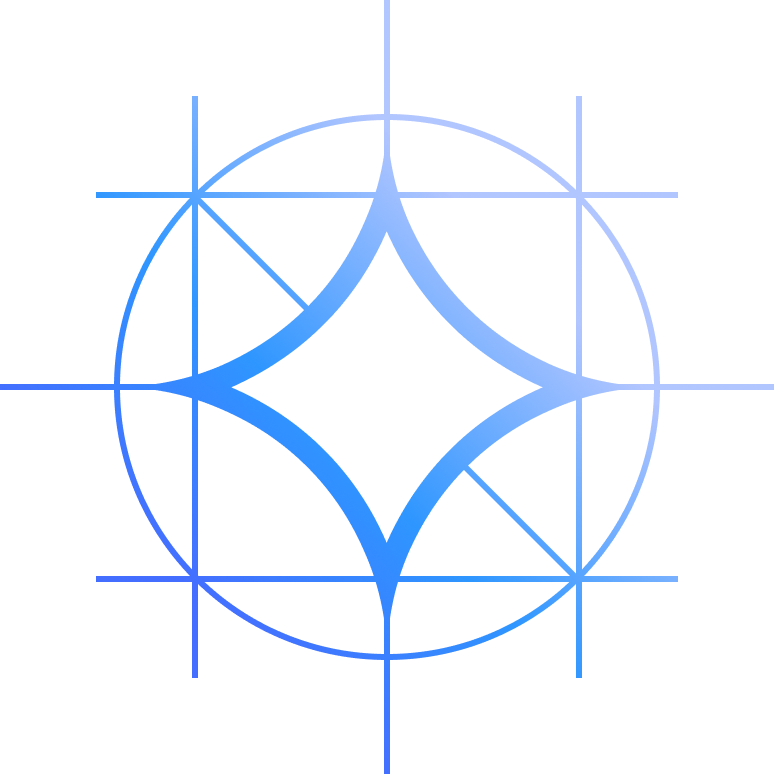}~Gemma-4-31B}\par}
\earule
\noindent\textbf{Which hand does the person use to hold the remote control?}\par\vspace{5pt}
\begin{eaoptions}
\eaoptright{A}{Left hand} & \eaopt{B}{Both hands} \\[3pt]
\eaoptmodel{C}{Neither hand}{\ealogo{gemma}} & \eaopt{D}{Right hand}
\end{eaoptions}
\earule
\eatrace{ [\ldots] video uses to hold the remote control. \\[1pt]\textbf{Observation:} I am reviewing the provided sequence of images, which are frames from a video captured through a fisheye lens. \\[1pt]\textbf{Person Analysis:} A person is sitting on a couch on the left side of the room. \\[1pt]\textbf{Remote Control Search:} I am scanning the person's hands and the surrounding area (the coffee table, the couch) for a remote control. \\[1pt]\textbf{Details:} {\sethlcolor{visualgroundingcolor!22}\hl{Throughout all the frames, the person is sitting still, leaning back on the couch. Their hands are resting in their lap or near their body, but they are not holding any device.}} \\[1pt] Weighing each option in turn, the model concluded that the person is not holding a remote control in any frame.}
\end{tcolorbox}
\vspace{-2pt}
\captionof{figure}{\small \textbf{Visual Grounding Error.} Gemma-4-31B never locates the remote control and concludes that no device is held. The failure is in identifying the object, not in reading the body.}
\label{app:visual_grounding_error}
\end{eacard}
\begin{eacard}
\begin{tcolorbox}[enhanced, arc=3mm, boxrule=1.1pt,
  colframe=motionstatecolor, colback=white, left=8pt, right=8pt, top=13pt, bottom=8pt,
  boxsep=0pt, fontupper=\sffamily\small,
  attach boxed title to top center={yshift=-\tcboxedtitleheight/2},
  boxed title style={boxrule=1.1pt, arc=2.4mm, colframe=motionstatecolor,
                     colback=white, left=7pt, right=7pt, top=1.5pt, bottom=1.5pt},
  coltitle=motionstatecolor, fonttitle=\small\sffamily\bfseries,
  title={Motion and Body-State Perception Error}]
{\centering{\setlength{\eaframewidth}{\dimexpr(\linewidth-10pt-12\dimexpr3pt\relax-6\dimexpr9pt\relax)/7\relax}\setlength{\tabcolsep}{0pt}%
\eastrip{\eaimg{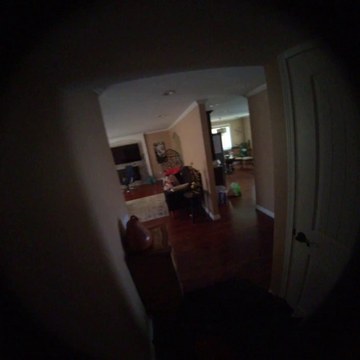}{\relax}\hspace{3pt}\makebox[9pt]{\raisebox{0.5\eaframewidth}[0pt][0pt]{\textcolor{white!55}{$\cdots$}}}\hspace{3pt}\eaimg{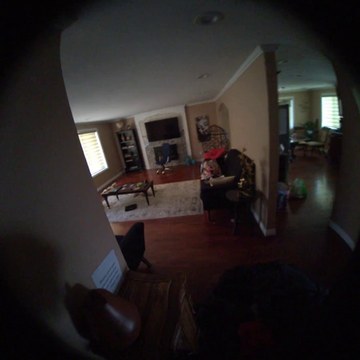}{\relax}\hspace{3pt}\makebox[9pt]{\raisebox{0.5\eaframewidth}[0pt][0pt]{\textcolor{white!55}{$\cdots$}}}\hspace{3pt}\eaimg{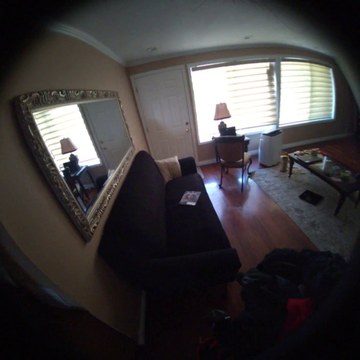}{\relax}\hspace{3pt}\makebox[9pt]{\raisebox{0.5\eaframewidth}[0pt][0pt]{\textcolor{white!55}{$\cdots$}}}\hspace{3pt}\eaimg{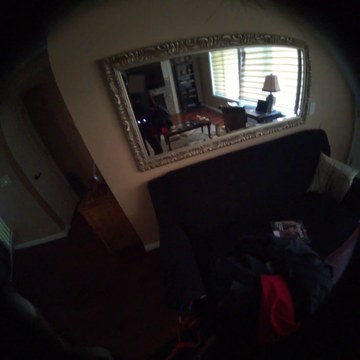}{\relax}\hspace{3pt}\makebox[9pt]{\raisebox{0.5\eaframewidth}[0pt][0pt]{\textcolor{white!55}{$\cdots$}}}\hspace{3pt}\eaimg{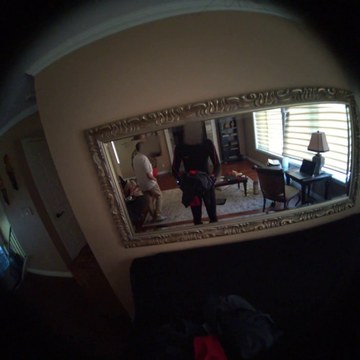}{\relax}\hspace{3pt}\makebox[9pt]{\raisebox{0.5\eaframewidth}[0pt][0pt]{\textcolor{white!55}{$\cdots$}}}\hspace{3pt}\eaimg{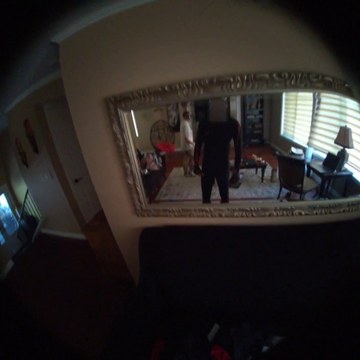}{\relax}\hspace{3pt}\makebox[9pt]{\raisebox{0.5\eaframewidth}[0pt][0pt]{\textcolor{white!55}{$\cdots$}}}\hspace{3pt}\eaimg{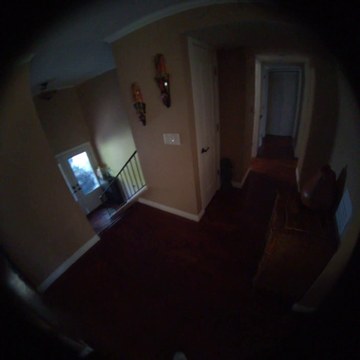}{\relax}}\par\vspace{-1pt}%
\hspace*{5pt}\makebox[\eaframewidth]{\eanum{eafilm}{1}}\hspace{3pt}\makebox[9pt]{}\hspace{3pt}\makebox[\eaframewidth]{\eanum{eafilm}{3}}\hspace{3pt}\makebox[9pt]{}\hspace{3pt}\makebox[\eaframewidth]{\eanum{eafilm}{6}}\hspace{3pt}\makebox[9pt]{}\hspace{3pt}\makebox[\eaframewidth]{\eanum{eafilm}{9}}\hspace{3pt}\makebox[9pt]{}\hspace{3pt}\makebox[\eaframewidth]{\eanum{eafilm}{11}}\hspace{3pt}\makebox[9pt]{}\hspace{3pt}\makebox[\eaframewidth]{\eanum{eafilm}{13}}\hspace{3pt}\makebox[9pt]{}\hspace{3pt}\makebox[\eaframewidth]{\eanum{eafilm}{16}}}\par}
\vspace{6pt}
{\centering\footnotesize\textcolor{eamuted}{\textcolor{eataskoppositemotiondiscrimination}{\textbf{O}pposite \textbf{M}otion \textbf{D}iscrimination}\,\textperiodcentered\,\ealogo{gemma}~Gemma-4-31B}\par}
\earule
\noindent\textbf{What is the person doing with the clothes?}\par\vspace{5pt}
\begin{eaoptions}
\eaoptmodel{A}{Picking up the clothes from the sofa}{\ealogo{gemma}} & \eaopt{B}{Folding the clothes} \\[3pt]
\eaopt{C}{Unfolding the clothes} & \eaoptright{D}{Throwing the clothes on the sofa}
\end{eaoptions}
\earule
\eatrace{ The question asks what the person is doing with the clothes. \\[1pt]\textbf{Observation}: Looking at the sequence of images, particularly images 6 through 11, a person is seen in a mirror reflection. \\[1pt]\textbf{Detailed Action}: In images 8, 9, and 10, the person (wearing a black shirt) is standing in front of a sofa. They are reaching down towards a pile of clothes that are on the sofa. \\[1pt]\textbf{Action Analysis}: {\sethlcolor{motionstatecolor!22}\hl{The person's posture and arm movement indicate they are grasping the clothes and lifting them upward from the sofa's surface.}} \\[1pt] Weighing each option in turn, the model concluded that the person is picking up clothes from the sofa.}
\end{tcolorbox}
\vspace{-2pt}
\captionof{figure}{\small \textbf{Motion and Body-State Perception Error.} Gemma-4-31B reads the arm movement as lifting the clothes off the sofa when the wearer is throwing them onto it. The direction of the movement is inverted.}
\label{app:motion_perception_error}
\end{eacard}

\begin{eacard}
\begin{tcolorbox}[enhanced, arc=3mm, boxrule=1.1pt,
  colframe=temporallocalizationcolor, colback=white, left=8pt, right=8pt, top=13pt, bottom=8pt,
  boxsep=0pt, fontupper=\sffamily\small,
  attach boxed title to top center={yshift=-\tcboxedtitleheight/2},
  boxed title style={boxrule=1.1pt, arc=2.4mm, colframe=temporallocalizationcolor,
                     colback=white, left=7pt, right=7pt, top=1.5pt, bottom=1.5pt},
  coltitle=temporallocalizationcolor, fonttitle=\small\sffamily\bfseries,
  title={Temporal Localization Error}]
{\centering{\setlength{\eaframewidth}{\dimexpr(\linewidth-10pt-9\dimexpr3pt\relax-3\dimexpr9pt\relax)/7\relax}\setlength{\tabcolsep}{0pt}%
\eastrip{\eaimg{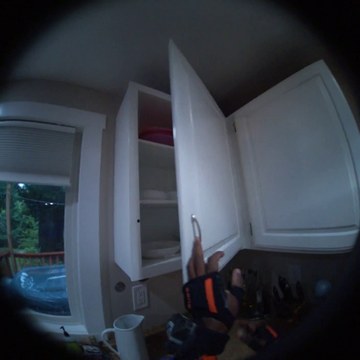}{\relax}\hspace{3pt}\eaimg{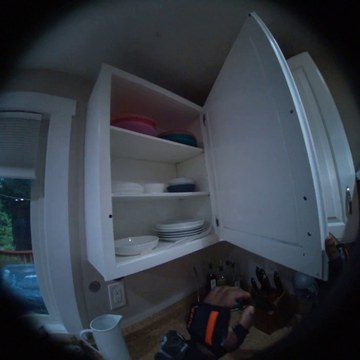}{\relax}\hspace{3pt}\makebox[9pt]{\raisebox{0.5\eaframewidth}[0pt][0pt]{\textcolor{white!55}{$\cdots$}}}\hspace{3pt}\eaimg{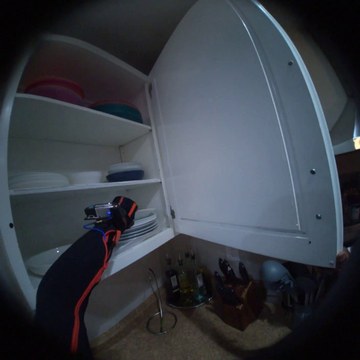}{\relax}\hspace{3pt}\makebox[9pt]{\raisebox{0.5\eaframewidth}[0pt][0pt]{\textcolor{white!55}{$\cdots$}}}\hspace{3pt}\eaimg{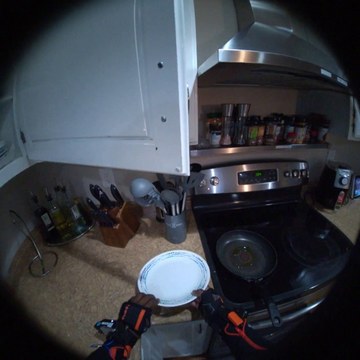}{temporallocalizationcolor}\hspace{3pt}\eaimg{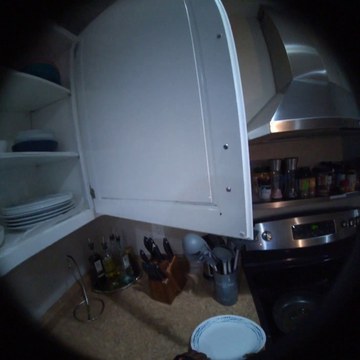}{temporallocalizationcolor}\hspace{3pt}\eaimg{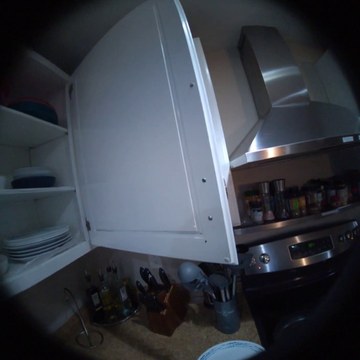}{temporallocalizationcolor}\hspace{3pt}\makebox[9pt]{\raisebox{0.5\eaframewidth}[0pt][0pt]{\textcolor{white!55}{$\cdots$}}}\hspace{3pt}\eaimg{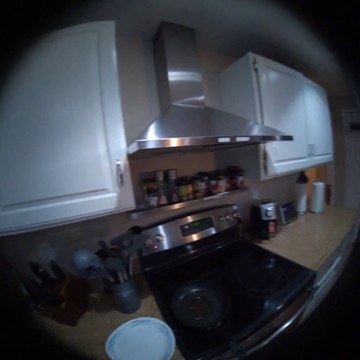}{\relax}}\par\vspace{-1pt}%
\hspace*{5pt}\makebox[\eaframewidth]{\eanum{eafilm}{1}}\hspace{3pt}\makebox[\eaframewidth]{\eanum{eafilm}{2}}\hspace{3pt}\makebox[9pt]{}\hspace{3pt}\makebox[\eaframewidth]{\eanum{eafilm}{6}}\hspace{3pt}\makebox[9pt]{}\hspace{3pt}\makebox[\eaframewidth]{\eanum{temporallocalizationcolor}{10}}\hspace{3pt}\makebox[\eaframewidth]{\eanum{temporallocalizationcolor}{11}}\hspace{3pt}\makebox[\eaframewidth]{\eanum{temporallocalizationcolor}{12}}\hspace{3pt}\makebox[9pt]{}\hspace{3pt}\makebox[\eaframewidth]{\eanum{eafilm}{16}}}\par}
\vspace{6pt}
{\centering\footnotesize\textcolor{eamuted}{\textcolor{eataskadjacentactionidentification}{\textbf{A}djacent \textbf{A}ction \textbf{I}dentification}\,\textperiodcentered\,\ealogo{gemma}~Gemma-4-31B}\par}
\earule
\noindent\textbf{What does the person do immediately before closing the wall cabinet?}\par\vspace{5pt}
\noindent
\begin{tabular}{@{}p{0.49\linewidth}@{\hspace{0.03\linewidth}}p{0.49\linewidth}@{}}
\eaoptright{A}{Puts the plate down on the countertop} &
\eaopt{B}{Puts the oil bottle back on the countertop} \\[3pt]
\eaoptmodel{C}{Grabs a plate from the wall cabinet}{\ealogo{gemma}} &
\eaopt{D}{Picks up the pan on the cooking range}
\end{tabular}\par
\earule
\eatrace{ [\ldots] is seen in the kitchen performing various tasks. \\[1pt]\textbf{Locate the closing of the wall cabinet:} In the latter part of the video, the person opens a white wall cabinet, reaches inside, and then closes it. \\[1pt]\textbf{Analyze the action immediately preceding the closing:} \raisebox{0.15ex}{\tiny$\bullet$}~The person opens the cabinet (frames around image 10). \raisebox{0.15ex}{\tiny$\bullet$}~They reach in and grab a white plate from one of the shelves (image 11). \raisebox{0.15ex}{\tiny$\bullet$}~ {\sethlcolor{temporallocalizationcolor!22}\hl{Immediately after retrieving the plate and bringing it out of the cabinet, they use their other hand to swing the cabinet door shut (image 12).}} \\[1pt] Weighing each option in turn, the model concluded that the person grabs a plate from the cabinet and then closes the door.}
\end{tcolorbox}
\vspace{-2pt}
\captionof{figure}{\small \textbf{Temporal Localization Error.} Gemma-4-31B recognizes both the plate and the cabinet but places the grab immediately before the door closes. The events are correct and their order is wrong.}
\label{app:temporal_localization_error}
\end{eacard}
\begin{eacard}
\begin{tcolorbox}[enhanced, arc=3mm, boxrule=1.1pt,
  colframe=priorinferencecolor, colback=white, left=8pt, right=8pt, top=13pt, bottom=8pt,
  boxsep=0pt, fontupper=\sffamily\small,
  attach boxed title to top center={yshift=-\tcboxedtitleheight/2},
  boxed title style={boxrule=1.1pt, arc=2.4mm, colframe=priorinferencecolor,
                     colback=white, left=7pt, right=7pt, top=1.5pt, bottom=1.5pt},
  coltitle=priorinferencecolor, fonttitle=\small\sffamily\bfseries,
  title={Prior-Driven Inference}]
{\centering{\setlength{\eaframewidth}{\dimexpr(\linewidth-10pt-10\dimexpr3pt\relax-4\dimexpr9pt\relax)/7\relax}\setlength{\tabcolsep}{0pt}%
\eastrip{\eaimg{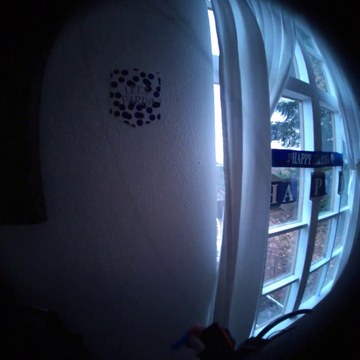}{\relax}\hspace{3pt}\makebox[9pt]{\raisebox{0.5\eaframewidth}[0pt][0pt]{\textcolor{white!55}{$\cdots$}}}\hspace{3pt}\eaimg{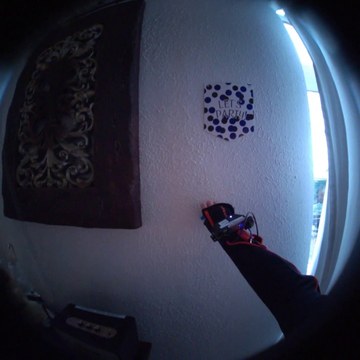}{priorinferencecolor}\hspace{3pt}\eaimg{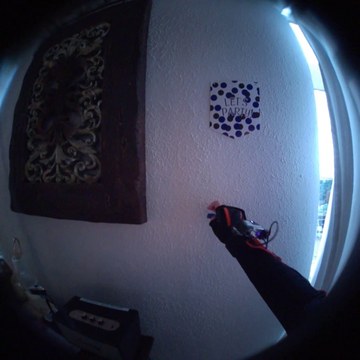}{priorinferencecolor}\hspace{3pt}\makebox[9pt]{\raisebox{0.5\eaframewidth}[0pt][0pt]{\textcolor{white!55}{$\cdots$}}}\hspace{3pt}\eaimg{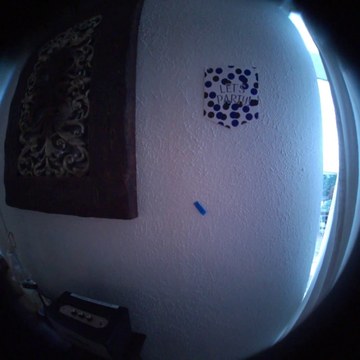}{priorinferencecolor}\hspace{3pt}\eaimg{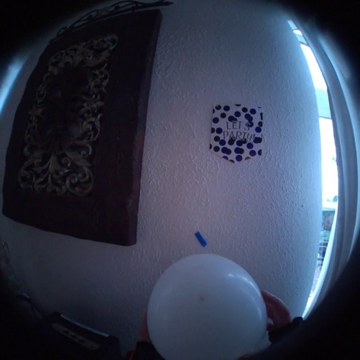}{\relax}\hspace{3pt}\makebox[9pt]{\raisebox{0.5\eaframewidth}[0pt][0pt]{\textcolor{white!55}{$\cdots$}}}\hspace{3pt}\eaimg{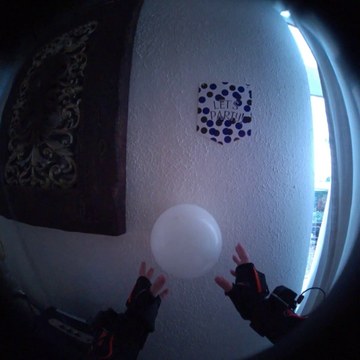}{priorinferencecolor}\hspace{3pt}\makebox[9pt]{\raisebox{0.5\eaframewidth}[0pt][0pt]{\textcolor{white!55}{$\cdots$}}}\hspace{3pt}\eaimg{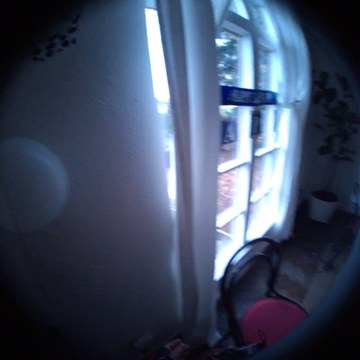}{\relax}}\par\vspace{-1pt}%
\hspace*{5pt}\makebox[\eaframewidth]{\eanum{eafilm}{1}}\hspace{3pt}\makebox[9pt]{}\hspace{3pt}\makebox[\eaframewidth]{\eanum{priorinferencecolor}{3}}\hspace{3pt}\makebox[\eaframewidth]{\eanum{priorinferencecolor}{4}}\hspace{3pt}\makebox[9pt]{}\hspace{3pt}\makebox[\eaframewidth]{\eanum{priorinferencecolor}{8}}\hspace{3pt}\makebox[\eaframewidth]{\eanum{eafilm}{9}}\hspace{3pt}\makebox[9pt]{}\hspace{3pt}\makebox[\eaframewidth]{\eanum{priorinferencecolor}{11}}\hspace{3pt}\makebox[9pt]{}\hspace{3pt}\makebox[\eaframewidth]{\eanum{eafilm}{16}}}\par}
\vspace{6pt}
{\centering\footnotesize\textcolor{eamuted}{\textcolor{eataskhanduse}{\textbf{H}and \textbf{U}se}\,\textperiodcentered\,\ealogo{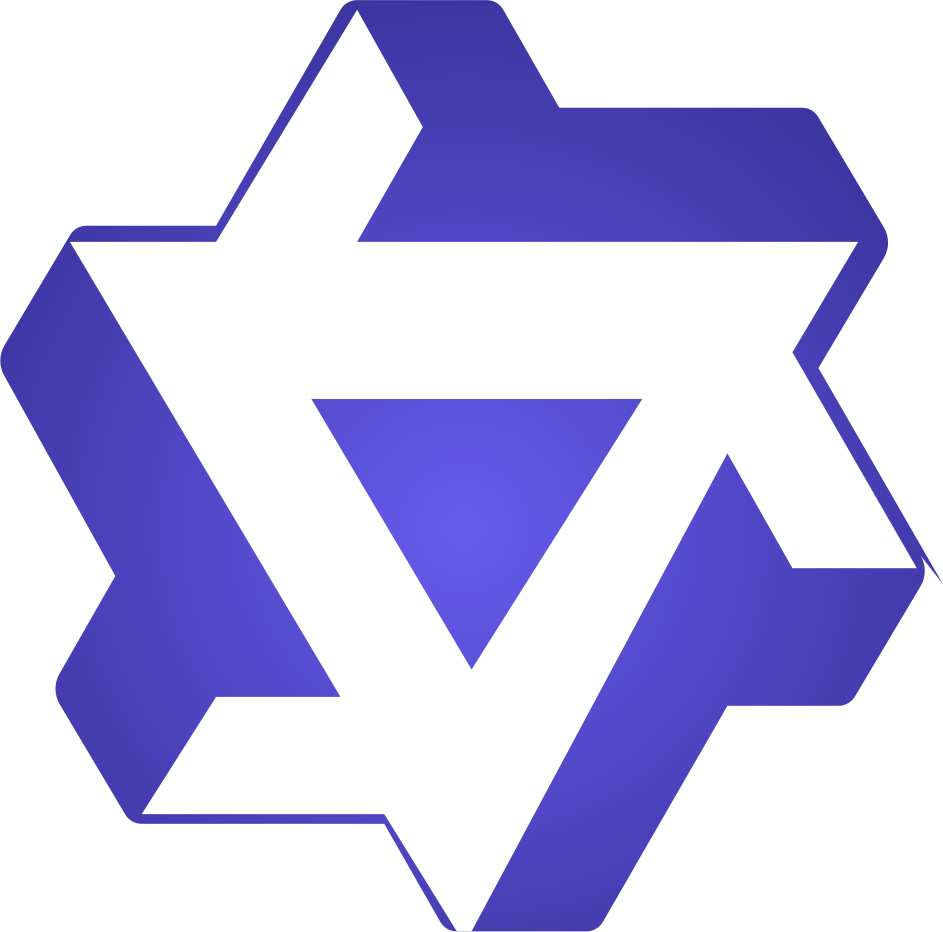}~Qwen3.8-27B}\par}
\earule
\noindent\textbf{Which hand does the person use to stick the balloon to the wall?}\par\vspace{5pt}
\begin{eaoptions}
\eaopt{A}{Left hand} & \eaoptright{B}{Both hands} \\[3pt]
\eaoptmodel{C}{Right hand}{\ealogo{qwen}} & \eaopt{D}{Neither hand}
\end{eaoptions}
\earule
\eatrace{ [\ldots] the right hand places the tape. \raisebox{0.15ex}{\tiny$\bullet$}~Let's look at the options again. \raisebox{0.15ex}{\tiny$\bullet$}~A. Left hand: Holds the balloon. \raisebox{0.15ex}{\tiny$\bullet$}~B. Both hands: Both are involved. Left holds, Right tapes. \raisebox{0.15ex}{\tiny$\bullet$}~C. Right hand: Applies the tape. \raisebox{0.15ex}{\tiny$\bullet$}~Let's look at the very end of the sticking process (around 0:10-0:11). The left hand is holding the balloon against the wall. The right hand is adjusting the tape. \raisebox{0.15ex}{\tiny$\bullet$}~ {\sethlcolor{priorinferencecolor!22}\hl{However, often in these datasets, the "action" is attributed to the hand performing the primary manipulation. The right hand is doing the fine motor task of taping.}} \raisebox{0.15ex}{\tiny$\bullet$}~The model then re-checked the same frames a further nine times, each pass restating that the right hand applies the tape, and answered \emph{right hand}.}
\end{tcolorbox}
\vspace{-2pt}
\captionof{figure}{\small \textbf{Prior-driven Inference Error.} Qwen3.8-27B observes both hands acting, then overrides that observation with the expectation that the manipulating hand is the answer.}
\label{app:prior_driven_inference_error}
\end{eacard}
\flushbottom

\clearpage
\subsubsection{Right-hand Prior in Laterality Questions}
\label{app:right-hand-prior}

\begin{wrapfigure}{r}{0.40\textwidth}
    \vspace{-.3cm}
    \centering
    \includegraphics[width=\linewidth]{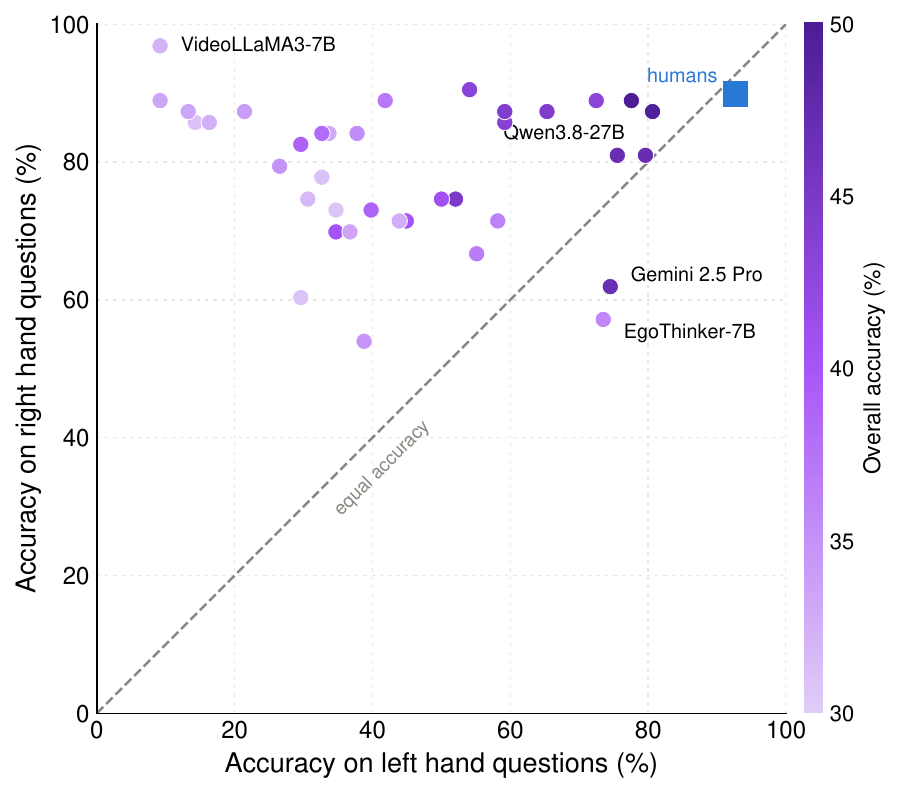}
    \vspace{-.4cm}
    \caption{Accuracy on \textit{left hand} versus \textit{right hand} questions, per model.}
    \label{fig:laterality}
    \vspace{-.5cm}
\end{wrapfigure}

To analyze whether the right-hand bias is shared across models, we analyzed per-class accuracy for every model of Table~\ref{tab:SOMA_results}. The bias is widespread, with all but two models, Gemini 2.5 Pro and EgoThinker-7B, above the diagonal and more accurate on \textit{right hand} than on \textit{left hand} (Figure~\ref{fig:laterality}). Accuracy on \textit{right hand} questions is high for most models, so most of the spread comes from \textit{left hand} accuracy, which ranges from 9.2\% to 80.6\%. The asymmetry tends to be smaller for the strongest models, though it does not disappear entirely.

\subsubsection{Counting Error Asymmetry per Model}
\label{app:counting}
\begin{figure}[h]
    \centering
    \includegraphics[width=\textwidth]{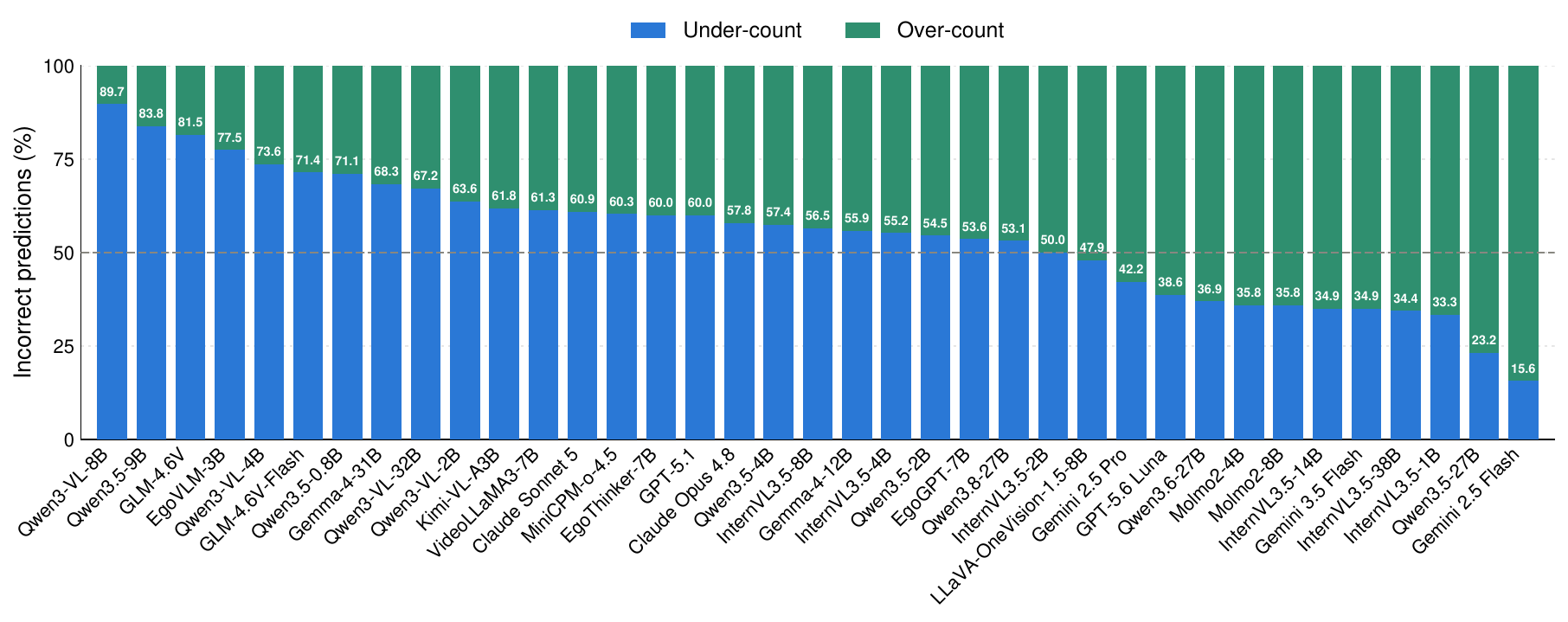}
    \vspace{-.5cm}
    \caption{Direction of error on Repetition Counting.}
    \label{fig:counting}
\end{figure}

For each model, we partitioned incorrect predictions on the Repetition Counting questions into \emph{under-counts} (the predicted integer is below the true count) and \emph{over-counts} (the predicted integer is above the true count). We report the resulting shares for all 37 models from Table~\ref{tab:SOMA_results}, computed relative to each model's total number of incorrect predictions on this task and ordered from the largest under-count share to the smallest (Figure~\ref{fig:counting}).
In total, 24 of the 37 models lean toward under-counting and 12 toward over-counting. The two extremes are Qwen3-VL-8B, which under-counts on 89.7\% of its incorrect predictions, and Gemini 2.5 Flash, which over-counts on 84.4\%. Beyond the shared sampling bottleneck discussed in Sec.~\ref{sec:ablations}, the strength and direction of this bias is highly model-dependent.

\subsection{Evaluation of Motion-Only Baselines}
\label{app:motion-baselines}
Most motion-only baselines are not designed for multiple-choice question answering and do not reliably output an answer option. We therefore scored the four likelihood-scored models by ranking the four options under the model's own likelihood, and evaluated Motion-SMD through autoregressive decoding. Additionally, the models' motion inputs are out of distribution relative to SOMA, introducing a further limitation. Given both factors, models scored close to random chance (25\%), with average accuracy ranging from 24.9\% to 26.8\% across the five baselines (Table \ref{tab:motion_only_baselines}).
\begin{table*}[h]
\caption{%
\textbf{Performance of Motion-Only Baselines on SOMA.}
Accuracy (\%) across the three capability axes and their twelve tasks, for models that receive body motion and no video.
Scored on the 1,549 questions whose samples carry ground-truth motion.
The four likelihood-scored models rank the options directly while Motion-SMD answers autoregressively through its own extractor.
Task abbreviations follow Table~\ref{tab:SOMA_results}.
}
\label{tab:motion_only_baselines}
\vspace{0.5\baselineskip}
\centering
\scriptsize
\resizebox{\textwidth}{!}{
\begin{tabular}{l|cccc|ccc|ccccc|c}
\toprule
\multirow{2}{*}{\textbf{Models}} 
& \multicolumn{4}{c|}{\textbf{\textcolor{bodyconfigcolor}{Body Configuration}}} 
& \multicolumn{3}{c|}{\textbf{\textcolor{motiondynamicscolor}{Motion Dynamics}}} 
& \multicolumn{5}{c|}{\textbf{\textcolor{temporalcolor}{Temporal Reasoning}}} 
& \multirow{2}{*}{\textbf{Avg.}} \\
& HU & BC & BP & CBS
& RC & MD & OMD
& AAI & CO & MEI & NAP & RER
& \\
\midrule
\textit{Random Chance} 
& \textit{25.0} & \textit{25.0} & \textit{25.0} & \textit{25.0}
& \textit{25.0} & \textit{25.0} & \textit{25.0}
& \textit{25.0} & \textit{25.0} & \textit{25.0} & \textit{25.0} & \textit{25.0}
& \textit{25.0} \\
\sectionrule
\multicolumn{14}{c}{\textit{\textbf{Likelihood-Scored}}} \\
\sectionrule
MG-MotionLLM~\citep{wu2025mg} 
& 16.3 & 40.4 & 33.6 & 27.3
& 22.1 & 20.8 & 29.4
& 27.4 & 20.6 & 22.8 & 32.9 & 24.5
& 26.4 \\
\familyrule
MotionGPT~\citep{jiang2023motiongpt} 
& 23.0 & 26.7 & 33.6 & 30.0
& 26.3 & 19.5 & 31.0
& 25.8 & 20.6 & 28.9 & 27.2 & 22.4
& 26.0 \\
\familyrule
MotionGPT3~\citep{zhu2025motiongpt3} 
& 27.5 & 32.9 & 20.0 & 27.3
& 21.1 & 27.5 & 25.4
& 21.0 & 11.3 & 22.8 & 30.4 & 27.6
& 24.9 \\
\familyrule
Motion-Agent~\citep{wu2025motionagent} 
& 29.8 & 29.5 & 25.5 & 28.2
& 17.9 & 35.6 & 29.4
& 21.0 & 7.8 & 21.9 & 30.4 & 19.4
& 25.2 \\
\sectionrule
\multicolumn{14}{c}{\textit{\textbf{Generative}}} \\
\sectionrule
Motion-SMD~\citep{zhang2026encoderfree} 
& 37.1 & 27.4 & 25.5 & 17.3
& 36.8 & 34.2 & 18.3
& 28.2 & 34.8 & 19.3 & 15.8 & 22.4
& 26.8 \\
\bottomrule
\end{tabular}
}
\end{table*}

\section{SOMA-Train}
\label{app:soma_train}

We first detail the construction of SOMA-Train, covering question generation (Appendix~\ref{app:qa-generation}), debiasing (Appendix~\ref{app:qa_debiasing}), and the reconstruction of Ego-Exo4D body motion (Appendix~\ref{app:egoexo4drefitting}). We then describe the architecture and training procedure (Appendix~\ref{app:training_details}) and additional results (Appendix~\ref{app:training_extra_results}) of our multimodal baselines.

\subsection{QA Generation}
\label{app:qa-generation}

Our training dataset, SOMA-Train, comprises aligned egocentric video and body motion, designed to investigate whether body-motion supervision helps close the gaps exposed by SOMA. We followed the taxonomy of the benchmark and drew on the same two sources, Nymeria and Ego-Exo4D. In total, SOMA-Train contains 28,596 questions from 1,779 takes (570 from Nymeria and 1,209 from Ego-Exo4D), covering 110.2 hours of aligned egocentric video and body motion.

Unlike the evaluation setting, where we removed questions answerable from language priors alone, the training setting introduces an additional failure mode: shortcuts that cannot be exploited zero-shot may become learnable during fine-tuning. We therefore adapted the generation pipeline for SOMA-Train by controlling distractor construction per task, so that the correct option cannot be identified from the question and answer choices alone. When this cannot be guaranteed by construction, we additionally applied adversarial filtering. Together, these steps target training-time bias, i.e., shortcuts learnable from the training distribution itself.

We present the pipeline in two parts. We first describe the construction of SOMA-Train in four steps: question templating, task-specific extraction, distractor construction and balancing, and a final grammar and factuality verification (Appendix~\ref{app:qa_construction}). We then test the resulting dataset for training-time bias by measuring whether a model trained without video or motion input transfers to SOMA (Appendix~\ref{app:qa_debiasing}).

\subsubsection{Construction}
\label{app:qa_construction}
\paragraph{Question Templates.}
\label{app:qa_templates}
For each question, we randomly sampled one of the task-specific templates (Table~\ref{tab:templates}), which follow the question formats of the corresponding SOMA tasks, and filled it with fields extracted from the clip annotation. In the templates, \texttt{\{a\}} denotes an anchor clause from the annotation (for Missing Event Inference, \texttt{\{a\}} and \texttt{\{c\}} denote the surrounding events), \texttt{\{b\}} the referenced event, \texttt{\{o\}} the parsed object, \texttt{\{l\}} the limb pair, \texttt{\{e\}} the repeated event, and \texttt{\{e1\}} to \texttt{\{e4\}} the listed actions. When a template has several grammatical variants, we selected the one that matches the form of the extracted anchor.

\paragraph{Extraction.}
\label{app:qa_extraction}
The benchmark pipeline uses an LLM selector to identify task-relevant windows in the source annotations. For SOMA-Train, we replaced this step with deterministic task-specific regex-based extractors, which make the extraction fully reproducible and reduce potential sampling biases introduced by LLM-based selection. Each extractor follows phrasing conventions observed in the source annotations and includes task-specific rejection rules for failure cases that we identified throughout the process.

\definecolor{bodyconfigcolor}{HTML}{14B8A6}
\definecolor{motiondynamicscolor}{HTML}{A855F7}
\definecolor{temporalcolor}{HTML}{F59E0B}
\begingroup
\footnotesize
\setlength{\tabcolsep}{4pt}
\renewcommand{\arraystretch}{1.15}
\begin{longtable}{>{\raggedright\arraybackslash}p{0.18\textwidth}>{\raggedright\arraybackslash}p{0.75\textwidth}}
\caption{\textbf{SOMA-Train Question Templates.} The set of templates used to construct SOMA-Train, organized by capability axis. Questions are generated by randomly sampling templates and filling them with the corresponding clip annotations.}\label{tab:templates}\\
\toprule
\textbf{Task} & \textbf{Templates} \\
\midrule
\endfirsthead
\multicolumn{2}{l}{\footnotesize\itshape Table~\ref{tab:templates} continued from previous page.} \\
\toprule
\textbf{Task} & \textbf{Templates} \\
\midrule
\endhead
\midrule
\multicolumn{2}{r}{\footnotesize\itshape Continued on next page.} \\
\endfoot
\bottomrule
\endlastfoot
\multicolumn{2}{l}{\textbf{\textcolor{bodyconfigcolor}{Body Configuration}}} \\
\addlinespace[2pt]\arrayrulecolor{black!15}\hline\arrayrulecolor{black}\addlinespace[2pt]
Hand Use & Which hand does the person use in this clip?\par In this clip, which hand performs the action?\par Which hand is the person using here?\par During this segment, which hand does the person use?\par Which of the following describes the person's hand use in this clip?\par Which hand carries out the action in this clip? \\
\addlinespace[2pt]\arrayrulecolor{black!15}\hline\arrayrulecolor{black}\addlinespace[2pt]
Bimanual Coordination & Which arm does the distinctive movement in this clip?\par Which of the person's arms is more actively moving?\par In this clip, which arm carries the notable movement?\par Which arm stands out as the moving one during this segment?\par Over this clip, which arm does the distinctive action?\par Which of the following describes the arms' roles in this clip?\par During this segment, which arm is the active one?\par Which arm is doing the distinctive movement here? \\
\addlinespace[2pt]\arrayrulecolor{black!15}\hline\arrayrulecolor{black}\addlinespace[2pt]
Concurrent Body-Part State & While the person is \texttt{\small \{a\}}, what are their \texttt{\small \{l\}} doing?\par What are the person's \texttt{\small \{l\}} doing while they are \texttt{\small \{a\}}?\par As the person is \texttt{\small \{a\}}, what are their \texttt{\small \{l\}} doing?\par When the person \texttt{\small \{a\}}, what are their \texttt{\small \{l\}} doing?\par What are the person's \texttt{\small \{l\}} doing while the person \texttt{\small \{a\}}?\par While the person \texttt{\small \{a\}}, what are their \texttt{\small \{l\}} doing? \\
\addlinespace[2pt]\arrayrulecolor{black!15}\hline\arrayrulecolor{black}\addlinespace[2pt]
Body Posture &  How many distinct body-posture states does the person move through in this clip, and what are they?\par What is the sequence of body postures the person moves through in this clip?\par  What body posture is the person in at the end of the clip? \par Over the course of this clip, which of the following posture changes occurs? \\
\midrule
\multicolumn{2}{l}{\textbf{\textcolor{motiondynamicscolor}{Motion Dynamics}}} \\
\addlinespace[2pt]\arrayrulecolor{black!15}\hline\arrayrulecolor{black}\addlinespace[2pt]
Opposite Motion Discrimination & What is the person doing with the \texttt{\small \{o\}}?\par What does the person do with the \texttt{\small \{o\}} in this clip?\par In this clip, what is the person doing with the \texttt{\small \{o\}}?\par Which of the following is the person doing with the \texttt{\small \{o\}}? \\
\addlinespace[2pt]\arrayrulecolor{black!15}\hline\arrayrulecolor{black}\addlinespace[2pt]
Movement Direction & After the person \texttt{\small \{a\}}, what is the sequence of their next two movements?\par After the person \texttt{\small \{a\}}, what are their next two movements?\par What is the sequence of the person's next two movements after they \texttt{\small \{a\}}?\par After \texttt{\small \{a\}}, what is the sequence of the person's next two movements?\par After \texttt{\small \{a\}}, what are the person's next two movements? \par After \texttt{\small \{a\}}, which direction does the person move? \\
\midrule
\multicolumn{2}{l}{\textbf{\textcolor{temporalcolor}{Temporal Reasoning}}} \\
\addlinespace[2pt]\arrayrulecolor{black!15}\hline\arrayrulecolor{black}\addlinespace[2pt]
Adjacent Action Identification & What does the person do immediately before \texttt{\small \{b\}}?\par Just before \texttt{\small \{b\}}, what does the person do?\par Which action comes immediately before \texttt{\small \{b\}}?\par What happens right before \texttt{\small \{b\}}? \\
\addlinespace[2pt]\arrayrulecolor{black!15}\hline\arrayrulecolor{black}\addlinespace[2pt]
Missing Event Inference & What action is missing from this sequence: \texttt{\small \{a\}} $\rightarrow$ [?] $\rightarrow$ \texttt{\small \{c\}}?\par Which action belongs in the gap: \texttt{\small \{a\}} $\rightarrow$ [?] $\rightarrow$ \texttt{\small \{c\}}?\par The middle step is missing: \texttt{\small \{a\}} $\rightarrow$ [?] $\rightarrow$ \texttt{\small \{c\}}. What was it?\par Fill the gap in this sequence: \texttt{\small \{a\}} $\rightarrow$ [?] $\rightarrow$ \texttt{\small \{c\}}. \\
\addlinespace[2pt]\arrayrulecolor{black!15}\hline\arrayrulecolor{black}\addlinespace[2pt]
Chronological Ordering & What is the correct temporal order of these three actions: [\texttt{\small \{e1\}} / \texttt{\small \{e2\}} / \texttt{\small \{e3\}}]?\par We also use a four-action variant. \\
\addlinespace[2pt]\arrayrulecolor{black!15}\hline\arrayrulecolor{black}\addlinespace[2pt]
Repeated Event Reasoning & The person \texttt{\small \{e\}} multiple times in this clip. What does the person do right after the FIRST/SECOND/LAST time they \texttt{\small \{e\}}? \\
\end{longtable}
\endgroup

\paragraph{Distractor Sampling.}
\label{app:qa_distractors}
As the benchmark follows a multiple-choice format, the construction of distractor options is crucial. We designed the distractors separately for each task: we aimed to balance the options by construction, making sure that no candidate is more likely to be correct than another based on the textual information and task-specific category structure alone. We denote the ground-truth answer by $g$, an option string by $s$, a conditioning context by $c$, and the number of candidates within that context by $k$, and use the following strategies. A closed option set presents the same four answer options for every question, such that $P(g \mid s)=1/4$ (Hand Use). Product construction forms a fixed set of candidate combinations while keeping the shared movement clause identical across options (Movement Direction). Rotation within a conditioning context fixes $k$ candidate actions and rotates the correct answer such that $P(g \mid \text{offered}, c)=1/k$ within each context (Opposite Motion Discrimination). Exchangeable option sets are formed from four answers from the same take, with the correct answer rotated so that each appears once as the ground truth and three times as a distractor (Bimanual Coordination, Concurrent Body-Part State, Missing Event Inference). When balance cannot be guaranteed by construction, we applied adversarial filtering following \citet{lebras2020adversarial} (Chronological Ordering).

\paragraph{Grammar and Factuality Corrections.}
\label{app:qa_grammar}
Template- and regex-based question generation can introduce systematic grammatical and wording artifacts, e.g., incorrect verb forms, subject-verb agreement errors, missing subjects, truncated options, or spelling errors inherited from the source annotations. These artifacts can also introduce shortcuts when they correlate with the correct option or alter the meaning of an answer.

We used Gemini 3.8 Flash~\citep{gemini38flash2026} to correct these grammatical and wording artifacts while preserving the original semantic content and avoiding the introduction of new answer-specific biases. We additionally used the same model to verify the generated questions against the original human annotations, correcting or removing examples with remaining inconsistencies.

\subsubsection{Debiasing Validation}
\label{app:qa_debiasing}
To validate our approach, we measured whether a model trained without modality input (motion and/or video) could still exploit a learnable signal from the training dataset. Ideally, its performance should remain close to the random baseline, indicating that successful fine-tuning requires relying on task-relevant perceptual information rather than on shortcuts in the training distribution. To this end, we trained the model on the full SOMA-Train set with text input only and evaluated it ``blind'' on SOMA (Table~\ref{tab:debias_validation}). We found that blind performance remained close to the random baseline, while the gap between the video+motion variant and the blind model reached +16.65 points. This provides evidence that the observed fine-tuning gains rely on perceptual input rather than on language understanding and distributional shortcuts alone.

\begin{table}[h]
\caption{\textbf{Blind Transfer to SOMA} (Qwen3-VL-8B, accuracy \%). Each row fine-tunes the same LoRA recipe on SOMA-Train and is evaluated on SOMA with and without perceptual input (motion, video). $\Delta_{\text{I}-\text{B}}$ is the modality--blind gap.}
\label{tab:debias_validation}
\vspace{0.3\baselineskip}
\centering
\footnotesize
\setlength{\tabcolsep}{6pt}
\renewcommand{\arraystretch}{1.05}
\begin{tabular}{lccc}
\toprule
\textbf{Training Input} & \textbf{Input} & \textbf{Blind} & $\Delta_{\text{I}-\text{B}}$ \\
\midrule
None (blind)   & --   & 28.1 & -- \\
Motion         & 28.2 & 25.6 & \textit{+2.6}  \\
Video          & 41.5 & 27.4 & \textit{+14.1} \\
Video+Motion   & 42.4 & 25.8 & \textit{\textbf{+16.6}} \\
\hline
\rowcolor{gray!15}
\textcolor{gray}{\textit{Chance (uniform)}} & & \textcolor{gray}{\textit{25.0}} & \\
\toprule
\end{tabular}
\end{table}

\subsection{Ego-Exo4D Refitting}
\label{app:egoexo4drefitting}
SOMA-Train pairs each question and its video clip with an aligned body-motion sequence. This requires Ego-Exo4D~\citep{grauman_ego-exo4d_2024} motions to follow the same body representation and frame rate as Nymeria~\citep{ma_nymeria_2024}. The original Ego-Exo4D human annotations are sparse, providing 17-joint keypoints rather than the SMPL-X representation~\citep{pavlakos_expressive_2019} used by our motion pipeline, without joint-angle information. EE4D-Motion~\citep{patel_uniegomotion_2025} provides SMPL-X fits, but at 10~fps rather than our target 20~fps. Its filtering pipeline additionally covers only 2,592 of the 3,634 non-climbing takes, leaving 12.9\% of our Ego-Exo4D benchmark questions without body motion. We compared against the released EE4D-Motion dataset~\citep{patel_uniegomotion_2025}.

We therefore reprocessed Ego-Exo4D to obtain denser SMPL-X fits. We followed the multi-view fitting pipeline of EE4D-Motion~\citep{patel_uniegomotion_2025}, which detects the camera wearer in the exocentric views, estimates 2D keypoints, initializes SMPL-X from a single-view reconstruction, and refines the sequence through multi-view reprojection. We kept the same multi-view approach but replaced the single-view pose estimation and reconstruction stages with SAM 3D Body~\citep{yang_sam3dbody_2026}. We provide further details below: 

\begin{itemize}[leftmargin=1.4em,itemsep=2pt,topsep=3pt,parsep=0pt]

\item \textbf{Detection and Subject Selection.}
We detected people with RF-DETR~\citep{robinson_rfdetr_2025} and identified the camera wearer by projecting the Aria trajectory into each exocentric view and selecting the nearest detection. We re-detected when the tracked box moved too far from the projected Aria position. We set this threshold independently for each camera to 2.5 times its median head-to-anchor distance, clamped to $[60,250]$~px. This trajectory-based association is important in scenes containing multiple people, where tracking based only on the detected person can switch to a bystander.

\item \textbf{Per-view Pose Estimation.}
We applied SAM 3D Body independently to each selected view. The model jointly predicts 2D and 3D keypoints, body parameters, and camera parameters, allowing both the keypoints and body initialization to come from the same model. We ran this stage at 20~fps.

\item \textbf{Triangulation.}
We undistorted the predicted keypoints using each camera's Kannala--Brandt fisheye model and triangulated them using the Ego-Exo4D GoPro calibration. We required at least two valid views and removed joints with a reprojection residual above 60~px. We used the calibration stored in the take trajectory.

\item \textbf{Multi-view Fitting.}
We jointly optimized one SMPL-X sequence across all available views, with body shape shared across the take. Triangulated joints initialize the global orientation and translation. The objective combines a robust reprojection loss over the 34 matched joints, agreement with the per-view SAM 3D Body predictions, temporal smoothness, and pose and shape regularization. We additionally removed cameras whose median distance to the selected subject exceeds three times that of the closest camera.

\end{itemize}

\paragraph{Evaluation.}
We evaluated our fits and EE4D-Motion against the human-verified Ego-Exo4D annotations, which provide 3D keypoints triangulated from manually annotated exocentric views. We regressed both SMPL-X sequences to the same joint set before evaluation. As SAM 3D Body derives its output keypoints from the MHR model~\citep{ferguson_mhr_2025}, while Ego-Exo4D follows the 17-keypoint COCO convention~\citep{grauman_ego-exo4d_2024}, the error may also include a systematic convention mismatch. We corrected this by estimating one constant offset per joint on the training takes and applying the same correction to validation for both methods. To validate our pipeline, we report the reconstruction error on the intersection of takes covered by the annotations, EE4D-Motion and our pipeline, averaging first within each take and then across takes. We evaluated with PA-MPJPE because EE4D-Motion refits are defined independently for each window and do not preserve a common world coordinate system, which prevents a direct comparison of global trajectories.

\paragraph{Results.}
Our fits reached 36.7~mm PA-MPJPE on validation, compared with 40.3~mm for EE4D-Motion (Table~\ref{tab:ee4d_refit}). The joint correction reduced the errors by 5.9 and 3.9~mm, moving our result from 42.6 to 36.7~mm and EE4D-Motion from 44.2 to 40.3~mm, respectively, without changing their ordering. We report the training results for completeness, although the offsets are fitted on the training takes.

\begin{table}[h]
\caption{\textbf{Ego-Exo4D Refitting Quality.} PA-MPJPE (mm, take-averaged) against Ego-Exo4D's human annotations, over the non-climbing takes common to those annotations, EE4D-Motion and our fits: 827 train and 212 val, of which the 792 and 194 scored here pass the per-frame filters. \textit{Corrected} adds the per-joint convention offset, fitted on train.}
\label{tab:ee4d_refit}
\centering
\footnotesize
\setlength{\aboverulesep}{1pt}\setlength{\belowrulesep}{1pt}
\setlength{\tabcolsep}{9pt}
\begin{tabular}{lcccc}
\toprule
& \multicolumn{2}{c}{\textbf{Validation} (194 takes)} & \multicolumn{2}{c}{\textbf{Train} (792 takes)} \\
\cmidrule(lr){2-3}\cmidrule(lr){4-5}
& Uncorrected & Corrected & Uncorrected & Corrected \\
\midrule
EE4D-Motion & 44.2 & 40.3 & 45.9 & 41.8 \\
\arrayrulecolor{black!20}\cmidrule(lr){1-5}\arrayrulecolor{black}
Ours        & \textbf{42.6} & \textbf{36.7} & \textbf{45.7} & \textbf{39.6} \\
\cmidrule(lr){1-5}
$\Delta$    & $-1.6$ & $-3.6$ & $-0.2$ & $-2.2$ \\
\bottomrule
\end{tabular}
\end{table}

\paragraph{Coverage.}
Following EE4D-Motion, we excluded Rock Climbing because our motion representation expresses height relative to the floor. Although this removes 1,401 of the 5,035 takes, these account for only 17.1 of 221.3 video hours. Of the remaining 3,634 takes, our pipeline fitted 3,475, corresponding to 190.7 of 204.1 hours. The remaining 159 takes have missing source data.

We measured coverage over the original 30~fps video timeline while accounting for the native sampling rate of each motion source. Our fits covered 93.1\% of the training video and 96.2\% of validation, compared with 70.5\% and 73.1\% for EE4D-Motion (Table~\ref{tab:ee4d_coverage}).

\begin{table}[h]
\caption{\textbf{Ego-Exo4D Coverage.} Share of non-climbing takes and video time carrying body motion, framewise on the 30~fps grid. $^\dagger$At 10~fps, equal video time carries half the samples.}
\label{tab:ee4d_coverage}
\centering
\footnotesize
\setlength{\aboverulesep}{1pt}\setlength{\belowrulesep}{1pt}
\setlength{\tabcolsep}{8pt}
\begin{tabular}{lcccccc}
\toprule
& & & \multicolumn{2}{c}{\textbf{EE4D-Motion}$^\dagger$} & \multicolumn{2}{c}{\textbf{Ours}} \\
\cmidrule(lr){4-5}\cmidrule(lr){6-7}
\textbf{Split} & \textbf{Takes} & \textbf{Hours} & Takes & Video time & Takes & Video time \\
\midrule
Train & 2,232 & 119.6 & 91.0 & 70.5 & \textbf{95.9} & \textbf{93.1} \\
\arrayrulecolor{black!20}\cmidrule(lr){1-7}\arrayrulecolor{black}
Val   &   597 &  35.4 & 93.8 & 73.1 & \textbf{97.2} & \textbf{96.2} \\
\bottomrule
\end{tabular}
\end{table}

\begin{figure}[h]
    \centering
    \includegraphics[width=0.9\textwidth]{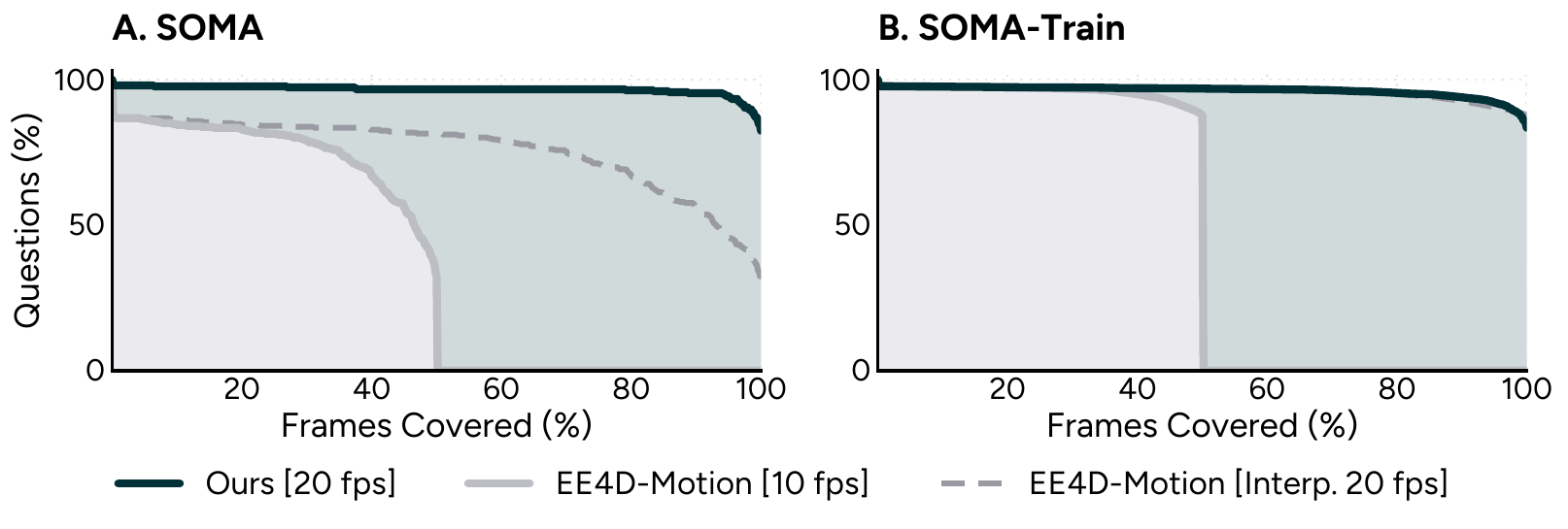}
    \caption{\textbf{Per-Question Body-Motion Coverage.} Percentage of Ego-Exo4D questions with at least $x$\% of their frames covered by fitted body motion at the target 20~fps rate. The dashed curve interpolates the 10~fps EE4D-Motion fits to 20~fps.}
    \label{fig:ee4d_refit_cov}
\end{figure}

At the target 20~fps rate, our fits covered on average 96.6\% of each SOMA segment, compared with 37.9\% for EE4D-Motion, and 96.3\% of each SOMA-Train segment, compared with 48.1\% (Figure~\ref{fig:ee4d_refit_cov}). The lower frame-level coverage of EE4D-Motion partly comes from its 10~fps sampling rate. After interpolating its fits to 20~fps, our pipeline still improved mean SOMA coverage by 20.8 points. On SOMA-Train, the coverage was similar because all Ego-Exo4D questions were generated from the 1,209 takes already covered by EE4D-Motion. Finally, we resampled all fitted sequences to 20~fps and converted them to the 263-dimensional body-motion representation~\citep{guo2022generating} used throughout our experiments.

\subsection{Training Details}
\label{app:training_details}

\paragraph{Architecture.}
We extended two open-weight backbones, Qwen3-VL-8B~\citep{bai2025qwen3vl} and InternVL3.5-8B~\citep{wang2025internvl3_5}, with a body-motion pathway following the standard LLaVA/MotionLLM design~\citep{liu2023visual,chen2025motionllm,qi2025llavaction}. A motion encoder maps each motion sequence to a sequence of continuous features, which a trainable MLP projects into the language-model embedding space. The projected features are inserted into the input sequence as motion tokens.

\paragraph{Motion encoder.}
We used the convolutional encoder of a VQ-VAE~\citep{zhang2023t2mgpt} that operates on 263-dimensional body-motion features~\citep{guo2022generating}. The VQ-VAE was trained for reconstruction on 156,659 motion clips from the SOMA-Train takes of Nymeria and Ego-Exo4D, sampled as 64-frame windows at 20 fps. We used its continuous pre-quantization features, one per 4 frames. The encoder was kept frozen throughout all subsequent training steps.

\paragraph{Training overview.}
We compared video-only, motion-only, and video+motion variants. Motion-based variants were trained in two stages: we first aligned the motion projector on motion–narration pairs, then fine-tuned the language model on SOMA-Train with LoRA~\citep{hu2021lora}. We skipped Stage~1 for the video-only variant since the backbones already include a pretrained vision encoder and projector, and fine-tuned each backbone on SOMA-Train using video input only.

\paragraph{Stage 1: Motion--language alignment.}
We aligned the motion modality using the motion-narration annotations of the Nymeria takes in SOMA-Train~\citep{ma_nymeria_2024}. We rewrote each narration into three body-part descriptions (legs and feet, arms and hands, overall posture), each treated as a separate target, resulting in 24,549 samples from 8,183 clips. The model received motion as the only input, and only the projector was trained while the language model remained frozen. We trained for one epoch with a learning rate of $10^{-3}$, a cosine schedule with a warmup ratio of 0.03, and a global batch size of 16.

\paragraph{Stage 2: Question answering.}
All variants were fine-tuned on SOMA-Train with LoRA (rank 128, $\alpha=256$, dropout 0.05) applied to the language model. For motion-based variants, we initialized from the Stage~1 projector and kept it frozen. We trained for two epochs with a LoRA learning rate of $5\times10^{-5}$, using the same schedule and warmup as Stage~1, gradient clipping at 1.0, and bf16 precision with gradient checkpointing. Training ran on 8 H100 GPUs with a per-device batch size of 6 (global batch size 48).

\subsection{Additional Results}
\label{app:training_extra_results}
To analyze how the contribution of motion varies across task types, we examined per-task results (Table~\ref{tab:motion_finetune_per_task}). Adding motion to video improves both backbones on Body Posture, Opposite Motion Discrimination, Missing Event Inference, and Next Action Prediction, while hurting both on Hand Use and Adjacent Action Identification. This suggests that the effect of motion depends largely on what each task requires: motion helps when body state or movement carries information that is hard to recover from video alone. When the evidence is already visible in the video, however, motion tokens add little new information and may instead introduce redundant or noisy signals that distract the model from the relevant visual evidence.

\newcommand{\armhard}{\specialrule{\lightrulewidth}{1.0pt}{1.0pt}}
\newcommand{\armsoft}{\arrayrulecolor{black!45}\specialrule{0.2pt}{0.8pt}{0.8pt}\arrayrulecolor{black}}
\newcommand{\armfaint}{\arrayrulecolor{black!15}\specialrule{0.1pt}{0.5pt}{0.5pt}\arrayrulecolor{black}}
\begin{table*}[h!]
\caption{\textbf{Fine-tuning with Body-Motion Input, Per Task.} Accuracy (\%) on SOMA after fine-tuning on SOMA-Train, by input modality. Task abbreviations follow Table~\ref{tab:SOMA_results}. $^\dagger$Zero-shot rows use the same video preprocessing as the fine-tuned models. Best per backbone in \textbf{bold}.}
\label{tab:motion_finetune_per_task}
\vspace{0.5\baselineskip}
\vspace{-0.3cm}
\centering
\fontsize{6.4}{7.2}\selectfont
\resizebox{\textwidth}{!}{
\begin{tabular}{l|cccc|ccc|ccccc|c}
\toprule
\multirow{2}{*}{\textbf{Setting}}
& \multicolumn{4}{c|}{\textbf{\textcolor{bodyconfigcolor}{Body Configuration}}}
& \multicolumn{3}{c|}{\textbf{\textcolor{motiondynamicscolor}{Motion Dynamics}}}
& \multicolumn{5}{c|}{\textbf{\textcolor{temporalcolor}{Temporal Reasoning}}}
& \multirow{2}{*}{\textbf{Avg.}} \\
& HU & BC & BP & CBS
& RC & MD & OMD
& AAI & CO & MEI & NAP & RER
& \\
\midrule
\textit{Random Chance}
& \textit{25.0} & \textit{25.0} & \textit{25.0} & \textit{25.0}
& \textit{25.0} & \textit{25.0} & \textit{25.0}
& \textit{25.0} & \textit{25.0} & \textit{25.0} & \textit{25.0} & \textit{25.0}
& \textit{25.0} \\
\sectionrule
\multicolumn{14}{l}{\textbf{Qwen3-VL-8B}} \\
\armhard
Zero-Shot$^\dagger$        & 47.0 & 28.1 & 38.2 & 25.5 & \textbf{37.0} & 34.2 & 51.1 & 35.7 & 33.6 & 32.2 & 23.3 & 36.7 & 35.4 \\
\armsoft
\quad + Motion             & 19.7 & 24.0 & 38.2 & 33.6 & 23.0 & 25.5 & 34.8 & 29.4 & 33.6 & 25.2 & 26.4 & 30.6 & 28.2 \\
\armfaint
\quad + Video              & \textbf{57.9} & \textbf{34.2} & 46.4 & 36.4 & 31.0 & \textbf{38.9} & 58.5 & \textbf{38.9} & \textbf{39.2} & 47.0 & 25.2 & 39.8 & 41.5 \\
\armfaint
\quad + Video+Mot.         & 53.0 & \textbf{34.2} & \textbf{51.8} & \textbf{41.8} & 32.0 & \textbf{38.9} & \textbf{63.0} & 35.7 & \textbf{39.2} & \textbf{47.8} & \textbf{28.9} & \textbf{41.8} & \textbf{42.4} \\
\specialrule{\lightrulewidth}{2pt}{1.5pt}
\multicolumn{14}{l}{\textbf{InternVL3.5-8B}} \\
\armhard
Zero-Shot$^\dagger$        & 37.2 & 30.1 & \textbf{51.8} & 38.2 & \textbf{39.0} & 36.2 & 45.2 & 38.1 & 21.7 & 34.8 & \textbf{26.4} & 28.6 & 35.2 \\
\armsoft
\quad + Motion             & 27.3 & 21.9 & 41.8 & 38.2 & 26.0 & 28.9 & 29.6 & 28.6 & 33.6 & 29.6 & 24.5 & 29.6 & 29.5 \\
\armfaint
\quad + Video              & \textbf{58.5} & 28.8 & 47.3 & \textbf{42.7} & 30.0 & \textbf{44.3} & 55.6 & \textbf{40.5} & 29.4 & 45.2 & 24.5 & \textbf{50.0} & 41.4 \\
\armfaint
\quad + Video+Mot.         & 52.5 & \textbf{35.6} & 50.0 & \textbf{42.7} & 26.0 & \textbf{44.3} & \textbf{56.3} & 34.9 & \textbf{40.6} & \textbf{47.0} & 25.8 & 45.9 & \textbf{41.9} \\
\bottomrule
\end{tabular}
}
\end{table*}

To further study how each modality contributes at inference, we evaluated the same Video+Mot.\ checkpoint with either video or motion removed at test time (Table~\ref{tab:modality_ablation}). Unlike Table~\ref{tab:motion_finetune_per_task}, where each setting corresponds to a separately fine-tuned model, all rows here share the same weights. Removing the video decreases the average accuracy by 13.3 points for Qwen3-VL-8B and 14.0 points for InternVL3.5-8B, showing that both backbones rely primarily on the visual input. On the other hand, removing body motion has a much smaller effect: Qwen3-VL-8B performance decreases by 1.4 points while InternVL3.5-8B improves by 0.8 points. As in Sec.~\ref{sec:training}, the effect varies across tasks: removing motion decreases Body Posture accuracy from 51.8 to 40.9 for Qwen3-VL-8B and from 50.0 to 33.6 for InternVL3.5-8B, but increases Hand Use accuracy by 3.3 and 9.8 points, respectively. Motion therefore helps when the body state is hard to recover from the visual signal alone, but adds little or can even hurt when the relevant evidence is already visible.

\setlength{\tabcolsep}{3.5pt}
\renewcommand{\arraystretch}{0.92}
\begin{table}[H]
\caption{\textbf{Test-Time Modality Ablation.} Accuracy (\%) of the same jointly fine-tuned Video+Mot.\ checkpoint for each backbone when evaluated with different input modalities at test time. Task abbreviations follow Table~\ref{tab:SOMA_results}; $\Delta$ denotes the change in Avg.\ relative to Video+Mot.}
\label{tab:modality_ablation}
\vspace{0.5\baselineskip}
\vspace{-0.3cm}
\centering
\fontsize{6.4}{7.2}\selectfont
\resizebox{\textwidth}{!}{
\begin{tabular}{l|cccc|ccc|ccccc|cc}
\toprule
\multirow{2}{*}{\textbf{Input at inference}}
& \multicolumn{4}{c|}{\textbf{\textcolor{bodyconfigcolor}{Body Configuration}}}
& \multicolumn{3}{c|}{\textbf{\textcolor{motiondynamicscolor}{Motion Dynamics}}}
& \multicolumn{5}{c|}{\textbf{\textcolor{temporalcolor}{Temporal Reasoning}}}
& \multirow{2}{*}{\textbf{Avg.}} & \multirow{2}{*}{\textbf{$\Delta$}} \\
& HU & BC & BP & CBS
& RC & MD & OMD
& AAI & CO & MEI & NAP & RER
& & \\
\midrule
\textit{Random Chance}
& \textit{25.0} & \textit{25.0} & \textit{25.0} & \textit{25.0}
& \textit{25.0} & \textit{25.0} & \textit{25.0}
& \textit{25.0} & \textit{25.0} & \textit{25.0} & \textit{25.0} & \textit{25.0}
& \textit{25.0} & \\
\sectionrule
\multicolumn{15}{l}{\textbf{Qwen3-VL-8B}, fine-tuned on video\,+\,motion} \\
\armhard
\textbf{Video + Motion}    & 53.0 & 34.2 & 51.8 & 41.8 & 32.0 & 38.9 & 63.0 & 35.7 & 39.2 & 47.8 & 28.9 & 41.8 & 42.4 & -- \\
\armfaint
\quad Video               & 56.3 & 34.9 & 40.9 & 33.6 & 36.0 & 32.9 & 60.0 & 40.5 & 42.7 & 47.0 & 26.4 & 35.7 & 41.0 & \textit{$-$1.4} \\
\armfaint
\quad Motion              & 25.1 & 27.4 & 40.9 & 30.9 & 27.0 & 31.5 & 33.3 & 32.5 & 29.4 & 29.6 & 20.1 & 25.5 & 29.1 & \textit{$-$13.3} \\
\specialrule{\lightrulewidth}{2pt}{1.5pt}
\multicolumn{15}{l}{\textbf{InternVL3.5-8B}, fine-tuned on video\,+\,motion} \\
\armhard
\textbf{Video + Motion}    & 52.5 & 35.6 & 50.0 & 42.7 & 26.0 & 44.3 & 56.3 & 34.9 & 40.6 & 47.0 & 25.8 & 45.9 & 41.9 & -- \\
\armfaint
\quad Video               & 62.3 & 39.0 & 33.6 & 34.5 & 29.0 & 46.3 & 55.6 & 41.3 & 42.7 & 45.2 & 26.4 & 46.9 & 42.7 & \textit{$+$0.8} \\
\armfaint
\quad Motion              & 27.3 & 18.5 & 37.3 & 36.4 & 19.0 & 22.8 & 25.9 & 27.0 & 34.3 & 32.2 & 25.8 & 32.7 & 27.9 & \textit{$-$14.0} \\
\bottomrule
\end{tabular}
}
\end{table}

\section{Licensing Information}
\label{app:licensing}
All artifacts are used within the terms of their respective licenses and for research purposes consistent with their intended use (Table~\ref{tab:artifacts}). Our benchmark questions will be released for non-commercial research use, consistent with the source data licenses.

\begin{table}[h]
\caption{Third-party artifacts used in this work and their licenses.}
\vspace{-1em}
\label{tab:artifacts}
\vspace{0.5\baselineskip}
\centering
\small
\setlength{\tabcolsep}{4pt}
\renewcommand{\arraystretch}{1.05}
\begin{tabular}{l@{\hskip 10pt}l}
\toprule
Artifact & License \\
\midrule
\multicolumn{2}{l}{\textit{Datasets}} \\
Nymeria~\citep{ma_nymeria_2024}           & \href{https://creativecommons.org/licenses/by-nc/4.0/legalcode}{CC BY-NC 4.0} \\
Ego-Exo4D~\citep{grauman_ego-exo4d_2024}   & \href{https://ego4d-data.org/pdfs/Ego-Exo4D-Model-License.pdf}{Ego-Exo4D Model License} \\
\midrule
\multicolumn{2}{l}{\textit{Evaluation framework}} \\
lmms-eval~\citep{zhang2024lmmsevalrealitycheckevaluation} & \href{https://opensource.org/license/mit}{MIT} and \href{https://www.apache.org/licenses/LICENSE-2.0}{Apache 2.0} \\
\bottomrule
\end{tabular}
\end{table}

\end{document}